\PassOptionsToPackage{table,xcdraw}{xcolor}
\documentclass[pdflatex,sn-mathphys-num]{sn-jnl}
\usepackage[left,mathlines]{lineno}
\usepackage{etoolbox} 
\renewenvironment{internallinenumbers}{}{}
\usepackage[utf8]{inputenc}
\usepackage[T1]{fontenc}

\usepackage{amsmath}
\usepackage{amssymb}
\usepackage{amsfonts}
\usepackage{bm}
\usepackage{mathrsfs}
\usepackage{nicefrac}
\usepackage{dutchcal}

\usepackage{graphicx}
\usepackage{svg}
\usepackage{xcolor}
\usepackage{wrapfig}
\usepackage{rotating}
\usepackage{placeins}
\usepackage{booktabs}
\usepackage{multirow}
\usepackage{adjustbox}
\usepackage{threeparttable}
\usepackage{diagbox}
\usepackage{makecell}
\usepackage{tabularray}
\usepackage{booktabs} 

\usepackage{booktabs}
\usepackage{array}
\usepackage{ifthen}
\usepackage[normalem]{ulem} 
\usepackage{xcolor}
\usepackage{amssymb}

\newboolean{showedits}
\setboolean{showedits}{true} 
\ifthenelse{\boolean{showedits}}
{
\newcommand{\del}[1]{\textcolor{red}{\sout{#1}}} 
}{
\newcommand{\del}[1]{} 

}

\newboolean{showcomments}
\setboolean{showcomments}{true} 
\newcommand{\id}[1]{$-$Id: scgPaper.tex 32478 2010-04-29 09:11:32Z oscar $-$}

\ifthenelse{\boolean{showcomments}}
{\newcommand{\nbc}[3]{
{\colorbox{#3}{\bfseries\sffamily\scriptsize\textcolor{white}{#1}}}
{\textcolor{#3}{\sf\small$\blacktriangleright$\textit{#2}$\blacktriangleleft$}}}
}
{\newcommand{\nbc}[3]{}
\renewcommand{\del}[1]{} 
}

\definecolor{ibcolor}{rgb}{0.9,0.5,0}
\definecolor{dsrcolor}{rgb}{0.5,0.6,0}
\definecolor{cfcolor}{rgb}{0,0.5,0.9}
\definecolor{lwcolor}{rgb}{0.2,0.8,0.4}
\definecolor{eycolor}{rgb}{0.7,0.6,1.0}
\definecolor{oldcolor}{rgb}{0.2,0.2,0.2}
\definecolor{tdcolor}{rgb}{0.0,0.5,0.7}

\definecolor{headergray}{gray}{0.93}
\definecolor{codegray}{gray}{0.96}
\usepackage{placeins}

\usepackage{algorithm}
\usepackage{algpseudocode}
\usepackage{lmodern}
\usepackage{xspace}
\usepackage{enumitem}
\usepackage{pifont}
\usepackage{bbding}
\usepackage{balance}
\usepackage{listings}
\usepackage{tocloft}
\usepackage{framed}
\usepackage{microtype}
\usepackage{rotating}
\usepackage[most]{tcolorbox}
\usepackage{listings}
\usepackage{url}
\usepackage{hyperref}
\usepackage{tabularx}
\usepackage{array}
\definecolor{priorcol}{RGB}{237,247,253}
\definecolor{relationcol}{RGB}{238,247,242}
\definecolor{strategycol}{RGB}{252,244,234}
\definecolor{priorhead}{RGB}{208,231,244}
\definecolor{relationhead}{RGB}{218,238,226}
\definecolor{strategyhead}{RGB}{247,224,196}

\newcolumntype{Y}{>{\raggedright\arraybackslash}X}

\title{Creation begins with understanding: LLMs as strategy designers for privacy-preserving tabular data synthesis}

\author{%
\parbox{0.96\textwidth}{%
\centering
\mbox{Jinmeng Li\textsuperscript{1}},
\mbox{Quan Zhang\textsuperscript{2}},
\mbox{Hangting Ye\textsuperscript{1}},
\mbox{He Zhao\textsuperscript{3}},\\[-0.1em]
\mbox{Firas Laakom\textsuperscript{4}},
\mbox{Dandan Guo\textsuperscript{1,4,*}} and
\mbox{J\"urgen Schmidhuber\textsuperscript{4,5}}\\[0.9em]
{\small
\textsuperscript{1}School of Artificial Intelligence,
Jilin University, Changchun, China\\
\textsuperscript{2}Broad College of Business, Michigan State University,
East Lansing, USA\\
\textsuperscript{3}Commonwealth Scientific and Industrial
Research Organisation (CSIRO), Australia\\
\textsuperscript{4}Center of Excellence for Generative AI,
King Abdullah University of Science and Technology (KAUST),
Thuwal, Saudi Arabia\\
\textsuperscript{5}The Swiss AI Lab, IDSIA-USI/SUPSI, Lugano, Switzerland\\
\textsuperscript{*}Corresponding author. E-mail:
\href{mailto:guodandan@jlu.edu.cn}{guodandan@jlu.edu.cn}
}
}%
}

\definecolor{promptbg}{HTML}{F7F8FC}
\definecolor{promptframe}{HTML}{8AA4C8}
\definecolor{prompttitle}{HTML}{2D4F73}
\definecolor{promptnumber}{HTML}{8A8F98}

\definecolor{priorhead}{HTML}{D9EAF7}
\definecolor{priorcol}{HTML}{F2F8FC}

\definecolor{relationhead}{HTML}{DCEFE3}
\definecolor{relationcol}{HTML}{F3FAF5}

\definecolor{strategyhead}{HTML}{F9E6D2}
\definecolor{strategycol}{HTML}{FDF7F0}

\tcbuselibrary{listings,breakable,skins}

\newtcblisting{promptbox}[1]{
    enhanced,
    breakable,
    listing only,
    colback=promptbg,
    colframe=promptframe,
    coltitle=prompttitle,
    fonttitle=\bfseries\small,
    title={#1},
    arc=1.2mm,
    boxrule=0.6pt,
    left=1mm,
    right=1mm,
    top=1mm,
    bottom=1mm,
    before skip=6pt,
    after skip=8pt,
    listing options={
        basicstyle=\ttfamily\scriptsize,
        breaklines=true,
        breakatwhitespace=false,
        columns=fullflexible,
        keepspaces=true,
        showstringspaces=false,
        numbers=left,
        numberstyle=\tiny\color{promptnumber},
        numbersep=8pt,
        xleftmargin=1.5em,
        framexleftmargin=0.8em,
        tabsize=2
    }
}

\lstdefinestyle{textlisting}{
    basicstyle=\ttfamily\scriptsize,
    backgroundcolor=\color{promptbg},
    frame=single,
    rulecolor=\color{promptframe},
    framerule=0.6pt,
    framesep=6pt,
    breaklines=true,
    breakatwhitespace=false,
    columns=fullflexible,
    keepspaces=true,
    showstringspaces=false,
    numbers=left,
    numberstyle=\tiny\color{promptnumber},
    numbersep=8pt,
    xleftmargin=1.2em,
    framexleftmargin=1.0em,
    tabsize=2,
    captionpos=t
}

\definecolor{nmiBlue}{RGB}{45,85,130}
\definecolor{nmiLightBlue}{RGB}{238,244,250}
\definecolor{nmiBorder}{RGB}{150,170,190}
\definecolor{nmiGray}{RGB}{248,249,250}
\definecolor{codeBlue}{RGB}{30,90,160}
\definecolor{codeGreen}{RGB}{55,125,95}
\definecolor{codeRed}{RGB}{180,70,70}
\definecolor{codeGray}{RGB}{90,95,100}
\definecolor{codePurple}{RGB}{115,75,145}

\lstdefinestyle{synthesisPython}{
    language=Python,
    basicstyle=\ttfamily\scriptsize,
    keywordstyle=\color{codeBlue}\bfseries,
    stringstyle=\color{codePurple},
    commentstyle=\color{codeGray},
    showstringspaces=false,
    columns=fullflexible,
    keepspaces=true,
    breaklines=true,
    frame=none,
    backgroundcolor=\color{nmiGray},
    tabsize=4,
    escapeinside={(*@}{@*)}
}

\newtcolorbox{nmicodebox}[1]{
    enhanced,
    colback=nmiGray,
    colframe=nmiBorder,
    boxrule=0.65pt,
    arc=3pt,
    left=5pt,
    right=5pt,
    top=4pt,
    bottom=4pt,
    title=#1,
    coltitle=nmiBlue,
    fonttitle=\bfseries\small,
    colbacktitle=nmiLightBlue,
    boxed title style={
        colback=nmiLightBlue,
        colframe=nmiBorder,
        boxrule=0.55pt,
        arc=3pt
    },
    attach boxed title to top left={xshift=4pt,yshift=-1.0pt},
    top=10pt
}

\DeclareRobustCommand{\paneltarget}[2]{%
  \hypertarget{#1:#2}{\textbf{#2,}}%
}

\DeclareRobustCommand{\panelref}[2]{%
  \hyperlink{#1:#2}{~\ref*{#1}#2}%
}

\AtBeginEnvironment{figure}{\internallinenumbers}
\AtBeginEnvironment{figure*}{\internallinenumbers}
\AtBeginEnvironment{wrapfigure}{\internallinenumbers}

\AtBeginEnvironment{table}{\internallinenumbers}
\AtBeginEnvironment{table*}{\internallinenumbers}

\begin{document}

\maketitle

\section*{Abstract}

Sharing tabular data in high-stakes domains is constrained by privacy regulations. Synthetic data offer a promising alternative, but deep generative models are costly to train and difficult to audit, while LLM-based methods often serialize records as text, obscuring tabular structure and exposing sensitive data. We introduce Tabular Synthesis Strategy Designer (TabSSD), which uses an LLM to design synthesis procedures rather than directly generate records. TabSSD provides the LLM with tree-derived summaries of variable dependence rather than raw records, which produces Python programs for local execution and evaluation. Across twelve datasets, TabSSD strikes a favourable balance among statistical fidelity, predictive utility, and empirical privacy risk, achieving the best average rank across six metrics among ten methods. Moreover, it substantially reduces local computation and token consumption relative to the compared methods. By enabling human-guided refinement and eliminating user-side model tuning, TabSSD lowers the expertise and infrastructure barriers to transparent tabular data synthesis.

\section*{Introduction}

Tabular data, which combine heterogeneous (e.g., numerical and categorical) variables, underpin applications in various domains such as healthcare and physical systems~\cite{dsa2023prediction,Gamella2025}. Collaboration and data sharing in these settings are increasingly constrained by stringent data-protection regulations, including the European Union's General Data Protection Regulation~\cite{gdpr2016}. Tabular data synthesis offers a promising solution for data use without releasing sensitive records. High-quality synthetic data must balance statistical fidelity, downstream utility, and privacy: they should preserve the distributions and variable relationships of the original data so that models trained on them remain effective and generalizable, while minimizing memorization and re-identification risks~\cite{yan2022multifaceted,vanbreugel2024synthetic}. In safety-critical applications, the synthesis process should also be transparent and auditable, enabling domain experts to identify potential failure modes and determine whether the resulting data are suitable for downstream use.

\begin{figure}[!thbp]
\internallinenumbers
\centering
\includegraphics[width=\linewidth]{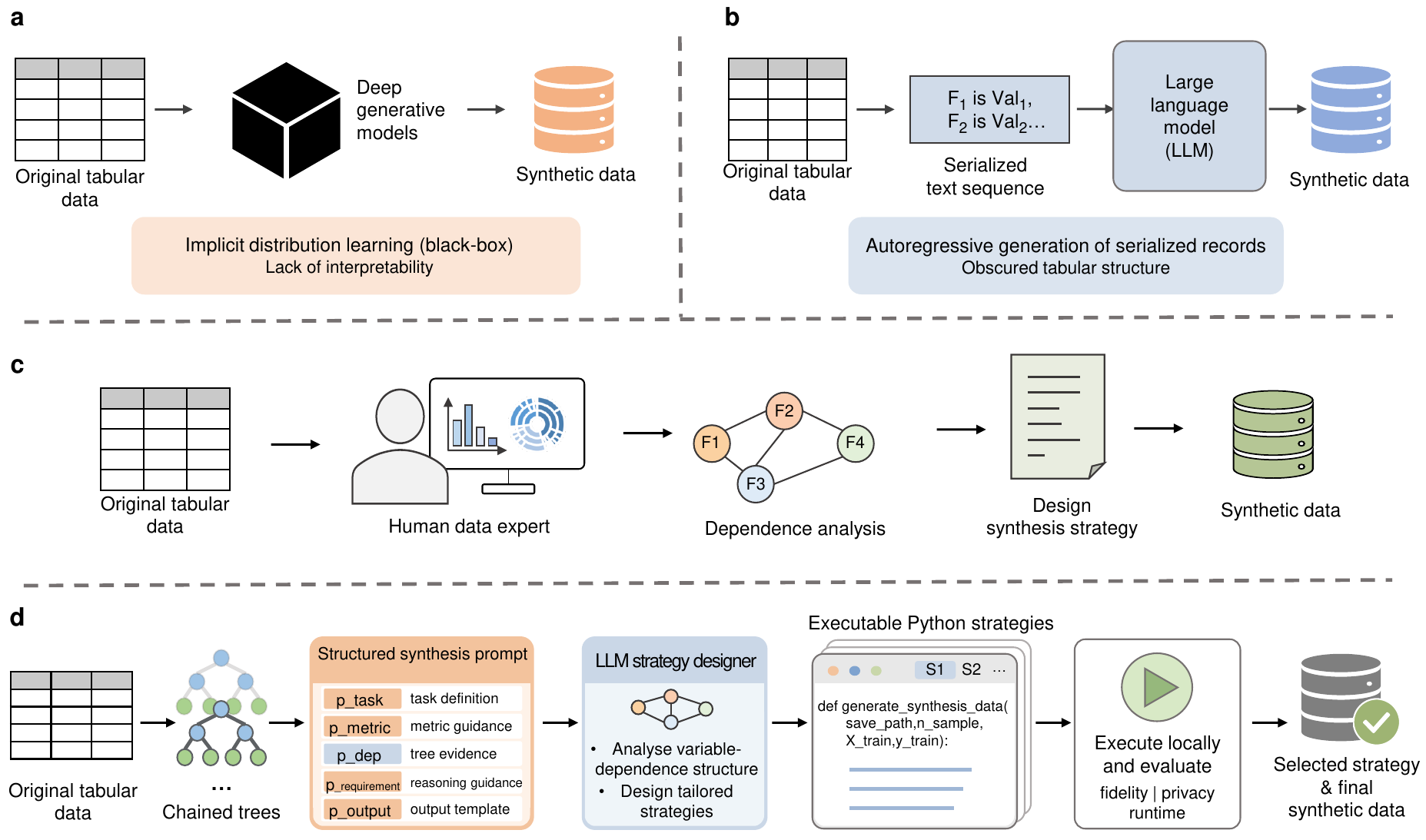}
\caption[Comparison of tabular data synthesis paradigms and overview of TabSSD.]{
\textbf{Comparison of tabular data synthesis paradigms and overview of the proposed TabSSD framework.}
\paneltarget{fig:framework_comparison}{a} Conventional deep generative models encode data distributions implicitly within trained parameters, resulting in opaque generation mechanisms that are difficult to inspect or audit.
\paneltarget{fig:framework_comparison}{b} Existing LLM-based methods serialize tabular records into one-dimensional text sequences, obscuring variable dependence, increasing computational overhead, and exposing raw records in prompts.
\paneltarget{fig:framework_comparison}{c} Human experts instead analyse variable dependence and domain constraints to design tailored synthesis strategies.
\paneltarget{fig:framework_comparison}{d} TabSSD similarly positions the LLM as a strategy designer. Chained trees extract the variable-dependence structure from the original data. This information is distilled and combined with the task definition, evaluation criteria, reasoning instructions, and output specification,  forming a prompt. Without accessing raw records, the LLM analyses the variable-dependence structure and produces multiple executable Python strategies. These candidates are executed and evaluated locally for statistical fidelity, empirical privacy risk, and runtime. Finally, the selected strategy is used to generate synthetic data.
}
\label{fig:framework_comparison}
\end{figure}

Existing synthesis approaches, however, face limitations arising from their underlying modelling paradigms. Classical statistical methods~\cite{reynolds2009gaussian,chawla2002smote,nowok2016synthpop} provide interpretable and principled mechanisms for tabular synthesis, but the reliance on distributional assumptions can limit their effectiveness for high-dimensional data with strong variable dependence~\cite{ohagan2016clustering,blagus2013smote}. Deep generative models, including generative adversarial networks~\cite{schmidhuber1990making,schmidhuber1991possibility,goodfellow2014generative} and diffusion models~\cite{jarzynski1997equilibrium,sohl2015deep,ho2020denoising}, offer greater modelling flexibility but encode data distributions implicitly within neural network parameters (Fig.\panelref{fig:framework_comparison}{a}). Their generation mechanisms are therefore difficult to inspect or audit. In high-stakes domains, this opacity limits experts' ability to understand the generation logic, diagnose failure modes, and correct undesirable behaviours~\cite{rudin2019stop,fang2025understanding}. Moreover, adapting these models to new data typically requires computationally intensive retraining.

Large language models (LLMs) have recently been adapted for tabular tasks~\cite{hegselmann2023tabllm,ye2025llm}, 
but using them directly as record-level generators for tabular synthesis creates a modality mismatch:
LLMs are optimized for sequential text, whereas tabular records are heterogeneous, non-sequential, and order-invariant~\cite{wu2025tabular,fang2024large}. Existing approaches commonly address this mismatch by serializing rows into text sequences (Fig.\panelref{fig:framework_comparison}{b}). Such representations obscure variable dependence and impose an artificial order that does not reflect the intrinsic structure of tabular data~\cite{wu2025tabular}. Methods based on serialized records generally rely on either task-specific fine-tuning~\cite{borisovlanguage,yang2024p} or in-context learning~\cite{seedat2023curated,kim2024epic}. Fine-tuning is data-intensive and computationally expensive, and in-context learning must infer the joint distribution from a limited set of examples and incurs increasing context-window and token costs as more examples are provided. Record serialization can also expose raw data during LLM adaptation or prompting, introducing additional privacy risks~\cite{ward2025tables,carlini2021extracting}. These limitations motivate a different paradigm in which the LLM reasons over summarized data structure rather than generating records directly from raw examples.

We propose the Tabular Synthesis Strategy Designer (TabSSD), which shifts the role of the LLM from a record-level generator to a high-level strategy designer. TabSSD draws on conditional modelling in statistical practice (Fig.\panelref{fig:framework_comparison}{c}), where a complex joint distribution is decomposed into a sequence of conditional models. To represent tabular structure without exposing raw records, TabSSD uses chained classification and regression trees (chained trees, hereafter) as a language-compatible bridge. These trees capture variable dependence through decision rules~\cite{breiman1984classification,grinsztajn2022tree}, compressing variable-dependence structure into compact prompts. Guided by these prompts, the LLM identifies redundant predictive dependence, determines sampling orders, selects appropriate conditional samplers, and expresses the resulting synthesis strategy as executable Python code (Fig.\panelref{fig:framework_comparison}{d}). Because the synthesis logic is represented as inspectable, executable code, domain experts can audit its rules, diagnose empirical anomalies, and make targeted revisions without computationally expensive retraining.
By automating dependence extraction, strategy generation, local validation, and candidate selection, TabSSD lowers the expertise and computational requirements of tabular synthesis. Through a standardized interface, it produces locally executable and inspectable programs without requiring users to design deep generative models, fine-tune LLMs, perform extensive hyperparameter tuning, or rely on local GPU acceleration.

We evaluate TabSSD on twelve heterogeneous datasets spanning a range of dimensionalities and sample sizes. 
Compared with representative deep generative models and LLM-based methods, TabSSD achieves a favourable balance among statistical fidelity, downstream utility, and privacy while substantially reducing computational and token costs. A case study on a severely imbalanced clinical dataset further demonstrates its practical use for minority-outcome augmentation and assesses whether the generated data preserve clinically meaningful phenotype distributions beyond conventional statistical metrics. Overall, these results show that strategy-level LLM reasoning provides a transparent and efficient approach to privacy-preserving tabular data synthesis and may extend to other heterogeneous data modalities.

\section*{Results}

\subsection*{Overview of TabSSD}
As shown in Fig.\panelref{fig:framework_comparison}{d}, TabSSD is an end-to-end framework that converts tabular data into executable synthesis strategies. It first extracts predictive dependence among variables using chained trees and encodes this information in a compact prompt. The LLM then analyses the dependence structure and designs multiple candidate strategies, each implemented as a Python program. These programs run locally and undergo evaluation for statistical fidelity, empirical privacy risk, and computational efficiency. The selected program synthesizes the data and provides an inspectable generation procedure. This design shifts the role of the LLM from record-level generation to strategy design. Because the LLM receives summaries of the variable-dependence structure rather than training records, the original data remain in the local environment throughout synthesis and evaluation. TabSSD runs on CPU only and is summarized as Algorithm~\ref{alg:framework}. 

\subsection*{Datasets, baselines, evaluation, and settings}

We assess TabSSD on 12 real-world datasets and in a clinical data-augmentation experiment. Supplementary Table~\ref{tab:dataset_stat} provides the data descriptions. We compare TabSSD with methods that span several synthesis paradigms: tabular deep generative models (CTGAN~\cite{xu2019modeling} and TVAE~\cite{xu2019modeling}), diffusion-based models (TabDDPM~\cite{kotelnikov2023tabddpm}, TabDiff~\cite{shi2025tabdiff}, and TABSYN~\cite{tabsyn}), a fine-tuned LLM using record serialization (GReaT~\cite{borisovlanguage}), tree- and energy-based models (ARF~\cite{watson2023adversarial} and NRGBoost~\cite{bravo2025nrgboost}), and the interpolation method SMOTE~\cite{chawla2002smote}. Each method uses the same train–test splits and generates as many synthetic samples as the corresponding training set contains. When required, validation uses only the training data. In the main experiments, TabSSD queries Gemini-2.5-Pro \(10\) times per dataset to generate candidate strategies. 
All local stages of TabSSD, including tree fitting, candidate execution, strategy evaluation, and final synthesis, run on CPU. 
Among the baselines, the tree-based methods ARF and NRGBoost run on CPU, and others use GPU acceleration because their running on CPU is prohibitively slow.
We follow the scheme of TABSYN~\cite{tabsyn} to report results averaged over 20 randomly sampled synthetic data.
Supplement~\ref{appendix:implementation_details} provides the implementation details. Across all datasets, we use the same prompt template, validation procedure, and selection criteria. No dataset-specific synthesis has been designed, tuned, or selected manually.

We evaluate the synthetic data from four perspectives: statistical fidelity, downstream predictive utility, empirical privacy risk, and computational efficiency. We additionally examine inspectability and editability through code-level case studies. Statistical fidelity is measured by marginal distribution error, pairwise correlation error, and \(\alpha\)-precision, quantifying how well the synthetic data preserve the empirical properties of the real training data. Under the train-on-synthetic, test-on-real protocol, AdaBoost~\cite{freund1997decision}, Random Forest~\cite{breiman2001random}, and XGBoost~\cite{chen2016xgboost} are used as downstream models to measure predictive utility, and we report their mean performance. Distance to closest record (DCR) measures empirical memorization risk, and \(\delta\)-presence measures empirical re-identification risk. Computational efficiency is reflected by the time required for the entire data synthesis process. The code-level case studies examine whether the resulting procedures are inspectable and editable. Supplement~\ref{appendix:Metrics} defines the evaluation metrics.

\begin{figure}[!htbp]
\internallinenumbers
\centering
\includegraphics[width=\linewidth]{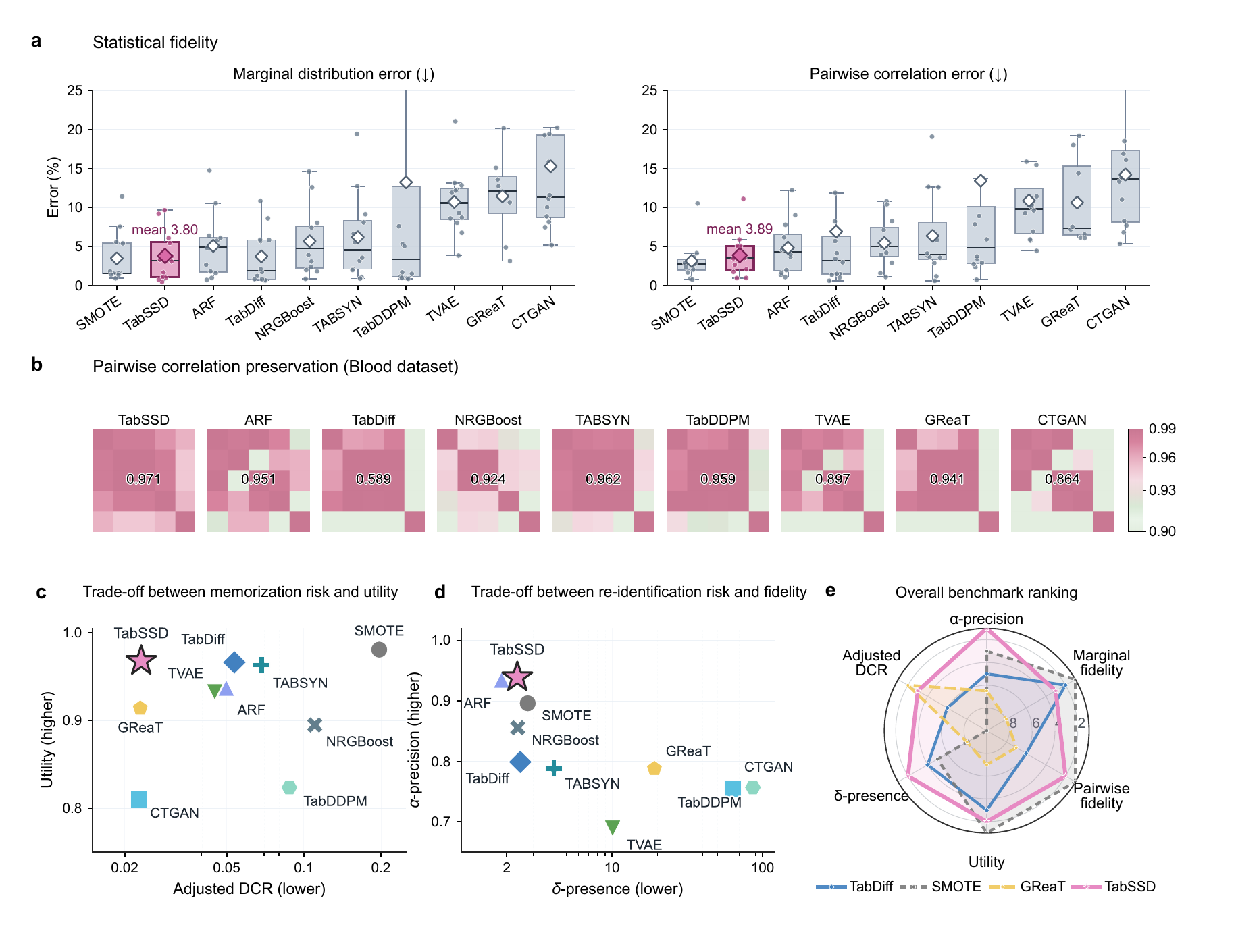}
\caption[Statistical fidelity, predictive utility, and empirical privacy risk of TabSSD and representative tabular data synthesis methods.]{
\textbf{Statistical fidelity, predictive utility, and empirical privacy risk of representative tabular data synthesis methods.}
\paneltarget{fig:fidelity_tradeoff}{a}
Statistical fidelity on 12 datasets, measured by marginal distribution error and pairwise correlation error. Lower values indicate higher fidelity. Boxes span the interquartile range, centre lines show the median, whiskers extend to 1.5 times this range, diamonds show the arithmetic mean, and dots denote performance on individual datasets. 
Because GReaT serializes tabular records as a text sequence, inputs from four high-dimensional datasets (\textit{SpeedDating}, \textit{Jasmine}, \textit{DNA}, and \textit{DARWIN}) exceed the LLM's context window; the GReaT results cover the remaining eight datasets. TabSSD is comparable to the best-performing SMOTE and better than the generative baselines.
\paneltarget{fig:fidelity_tradeoff}{b}
Pairwise correlation preservation on the \textit{Blood} dataset. Each cell reports the preservation score for a variable pair, and the value in the centre reports the mean across all pairs. Higher values indicate better performance. 
\paneltarget{fig:fidelity_tradeoff}{c}
Trade-off between downstream utility and DCR. A lower adjusted DCR indicates less proximity to training records and lower potential memorization risk. 
\paneltarget{fig:fidelity_tradeoff}{d}
Trade-off between \(\delta\)-presence and \(\alpha\)-precision. Lower \(\delta\)-presence indicates lower empirical re-identification risk, and higher \(\alpha\)-precision indicates higher sample-level fidelity. 
Panels c and d report the performance averaged across datasets with available results. Methods on the top left are preferred, and TabSSD is on the Pareto front in both comparisons.
\paneltarget{fig:fidelity_tradeoff}{e}
Ranking profiles of TabSSD, TabDiff, SMOTE, and GReaT across marginal fidelity, pairwise fidelity, utility, adjusted DCR, \(\alpha\)-precision, and \(\delta\)-presence. For each metric, all ten evaluated methods are ranked in the preferred direction, with rank 1 denoting the best performance. The ranks are mapped such that better performance appears closer to the outer ring. Across the six metrics, TabSSD ranks third, second, second, third, first, and second, respectively, yielding an average rank of 2.17. Among all methods compared, TabSSD exhibits the strongest and most balanced overall ranking profile.
}
\label{fig:fidelity_tradeoff}
\end{figure}

\subsection*{Synthetic data fidelity, utility, and privacy risks}

We compare TabSSD with the baseline synthesis methods on 12 real-world datasets and examine statistical fidelity, downstream predictive utility, and empirical privacy risk. Fig.~\ref{fig:fidelity_tradeoff} summarizes the trade-offs: TabSSD combines low statistical errors and competitive predictive utility with a favourable empirical privacy profile. Supplementary Tables~\ref{tab:marginal_density_error}--\ref{tab:delta_presence} report the complete numerical results.

\textit{Statistical fidelity and variable dependence.} SMOTE attains high statistical fidelity and predictive utility through linear interpolation between two data records. This mechanism preserves local geometry but produces the least favourable DCR result, indicating higher memorization risk and limiting its applicability for privacy-preserving data sharing~\citep{ganev2026smote}. We therefore focus the performance analysis and comparison on the deep generative and LLM-based methods. Among these methods, TabSSD has the second-lowest mean marginal distribution error (3.80\%), close to TabDiff (3.74\%; Fig.\panelref{fig:fidelity_tradeoff}{a}). TabSSD also has the lowest mean pairwise correlation error (3.89\%), a relative reduction of 19.79\% from ARF (4.85\%), the second-best method compared, and 43.95\% from TabDiff (6.94\%). On the \textit{Blood} dataset, TabSSD achieves a mean pairwise correlation preservation score of 0.971, better recovering variable correlations than the other generative methods (Fig.\panelref{fig:fidelity_tradeoff}{b}).

\textit{Fidelity--utility--privacy trade-offs.} High statistical fidelity and predictive utility can be achieved by sacrificing privacy protection. Fig.\panelref{fig:fidelity_tradeoff}{c} and Fig.\panelref{fig:fidelity_tradeoff}{d} show that TabSSD is on the Pareto front of the utility--memorization and fidelity--re-identification comparisons, respectively. TabSSD combines competitive predictive utility and leading \(\alpha\)-precision with memorization and re-identification risks that are comparable to or lower than those of the strongest baselines. Among other well-performing methods, GReaT and CTGAN have a favourable adjusted DCR but lower predictive utility and \(\alpha\)-precision, whereas ARF and TabDiff have favourable \(\delta\)-presence but a lower utility and/or a less favourable adjusted DCR than TabSSD. 
Fig.\panelref{fig:fidelity_tradeoff}{e} summarizes the six metrics in a rank-based radar plot. Each metric is ranked across the evaluated methods, with better ranks plotted closer to the outer boundary. TabSSD achieves the best average rank of 2.17 across the six metrics and is near the outer boundary, indicating its balanced performance.

More importantly, TabSSD mitigates a source of exposure not captured by these empirical metrics: raw records are not transmitted to the LLM during strategy generation. The LLM receives abstract partition rules rather than raw records, while tree extraction and synthesis execution remain within the data holder's environment. 
 Ablation studies support the contributions of tree-based dependence extraction, compact dependence encoding, and explicit prompt guidance (see Supplementary Tables~\ref{tab:feature_dependence_extraction_ablation}--\ref{tab:prompt_guidance_ablation}). In addition, the framework remains applicable across the evaluated LLM backbones, although synthesis performance varies across models and datasets (see Supplementary Table~\ref{tab:llm_backbone_sensitivity}).

\begin{figure}[!htbp]
\internallinenumbers
\centering
\includegraphics[width=\linewidth]{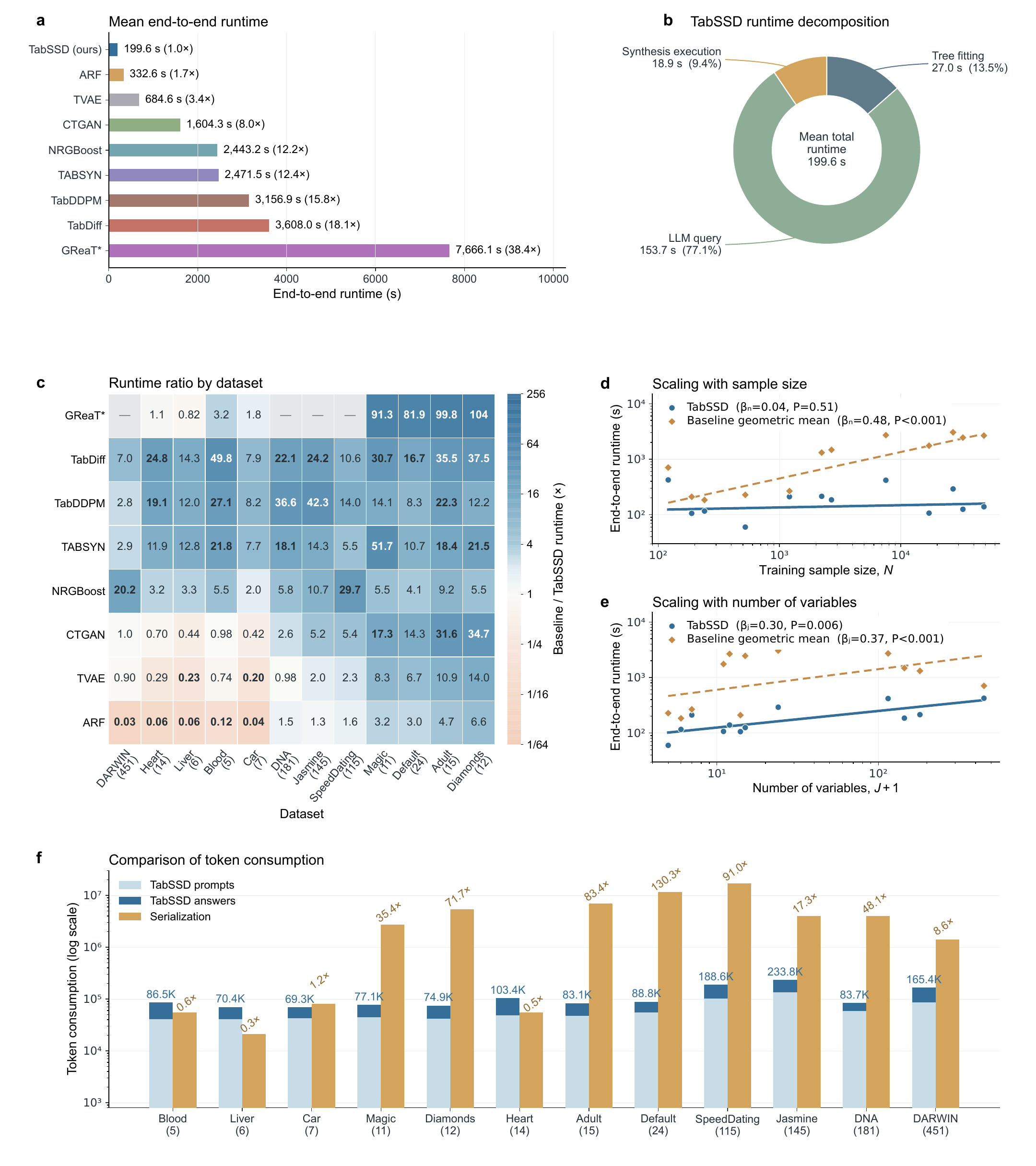}
\caption[Scalability and efficiency.]{
\textbf{Scalability and efficiency.}
Deep generative baselines and the serialization-based method GReaT use GPU acceleration. 
The runtime of TabSSD reported in \textbf{a} to \textbf{e} is calculated as the sum of the tree-fitting time and the parallel execution time of ten LLM queries and local syntheses.
\paneltarget{fig:scalability}{a} End-to-end runtime of data synthesis averaged across datasets. Parenthetical values indicate runtime relative to TabSSD.
\paneltarget{fig:scalability}{b} Decomposition of TabSSD's mean end-to-end runtime into tree fitting, LLM querying, and synthesis execution.
\paneltarget{fig:scalability}{c} Ratios of baseline runtime to TabSSD runtime on the 12 datasets. Values greater than 1 indicate that TabSSD is faster; dashes indicate unavailable results of GReaT due to the LLM's context window constraint. Datasets are ordered by training-set size in ascending order. Parenthetical values below dataset names indicate the number of variables \(J+1\).
\paneltarget{fig:scalability}{d} End-to-end runtime as a function of training-data size \(N\). Each point represents one dataset, with TabSSD compared against the geometric mean of the compared baselines.
\paneltarget{fig:scalability}{e} End-to-end runtime as a function of the number of variables \(J+1\). In \textbf{d} and \textbf{e}, lines show trends estimated using a log-linear model,
\(\log T=\alpha+\beta_N\log N+\beta_J\log(J+1)\).
The coefficients \(\beta_N\) and \(\beta_J\) quantify the associations of runtime with training-set size and variable dimension, respectively.
\paneltarget{fig:scalability}{f} Token consumption  on a logarithmic scale. Stacked blue bars show the ten-query token consumption of TabSSD prompts and answers. Orange bars show the tokens required for one-time training data serialization and new data generation. Values above the TabSSD bars report total prompt-and-answer token consumption, and values above the serialization bars report token consumption relative to TabSSD. Parenthetical values below dataset names in \textbf{c} and \textbf{f} indicate the number of variables \(J+1\).
In \textbf{a} and \textbf{c}, asterisks indicate that GReaT results are based on the eight datasets that fit within the LLM context window.
}
\label{fig:scalability}
\end{figure}

\subsection*{Scalability and token efficiency}

\textit{Computing efficiency and scalability.} We assess computational scalability across 12 datasets that vary in training data size \(N\) and variable dimension \(J+1\) ($J$ features and $1$ label) and report method efficiency in Fig.~\ref{fig:scalability}. 
TabSSD requires a \(199.6\)-second end-to-end runtime averaged across datasets. ARF, the fastest generative baseline compared, requires \(332.6\) seconds, or 1.7 times the TabSSD runtime (Fig.\panelref{fig:scalability}{a}). The other baselines require 3.4 to 38.4 times the TabSSD runtime. Notably, the reported runtimes of the deep generative baselines are obtained with GPU acceleration; on CPUs, these methods are prohibitively slow. By contrast, TabSSD's local execution uses CPU only.
The runtime decomposition in Fig.\panelref{fig:scalability}{b} attributes 77.1\% of TabSSD runtime to LLM querying and 22.9\% to tree fitting and synthesis execution. 
Fig.\panelref{fig:scalability}{c} presents dataset-level runtime ratios, calculated as baseline method runtime divided by TabSSD runtime. Values above 1 indicate that TabSSD is faster. 
Overall, TabSSD is competitive in terms of computing speed, and its largest runtime advantages occur on high-dimensional data.
Following established empirical computational-complexity analyses~\cite{goldsmith2007measuring}, we fit a log-linear model, \(\log T=\alpha+\beta_N\log N+\beta_J\log (J+1)\), to TabSSD runtime and the geometric mean runtime of the baselines. The baseline runtime has a significant positive association with training data size (\(\beta_N=0.48\)), whereas TabSSD exhibits an insignificant, near-zero association (\(\beta_N=0.04\); Fig.\panelref{fig:scalability}{d}). Within the evaluated datasets, TabSSD runtime shows no detectable association with training-data size (\(\beta_N=0.04\), \(P=0.51\)). This finding is consistent with the use of compact tree summaries instead of serializing records. Variable dimensionality has similar associations with TabSSD and baseline runtime (\(\beta_J=0.30\) and \(0.37\), respectively), but TabSSD retains a lower absolute runtime (Fig.\panelref{fig:scalability}{e}).
Supplementary Table~\ref{tab:run_time} reports the results for each dataset.

\textit{Token efficiency and context management.} We compare TabSSD with record serialization, a common approach used by LLM-based tabular synthesis methods. On low-dimensional datasets, TabSSD and serialization have comparable token consumption. As the variable dimension increases, however, serialization becomes increasingly costly and requires up to 130.3 times as many tokens as TabSSD (Fig.\panelref{fig:scalability}{f}).
TabSSD’s token efficiency stems from its use of compact chained-tree rules rather than raw record and variable values. For high-dimensional data, a greedy set-cover algorithm selects a subset of trees that covers all the variables (Algorithm~\ref{alg:framework}). Thus, increasing the data dimension adds only the rules needed to include previously uncovered variables, rather than replicating the entire dataset in the prompt. For example, a coverage of \textit{DARWIN}'s $451$ variables requires rules from only $54$ trees. By separating dependence extraction from prompt construction, TabSSD limits token growth and avoids the context-window failures that are often observed with record serialization.

\begin{figure}[!htbp]
\internallinenumbers
\centering
\includegraphics[width=\linewidth]{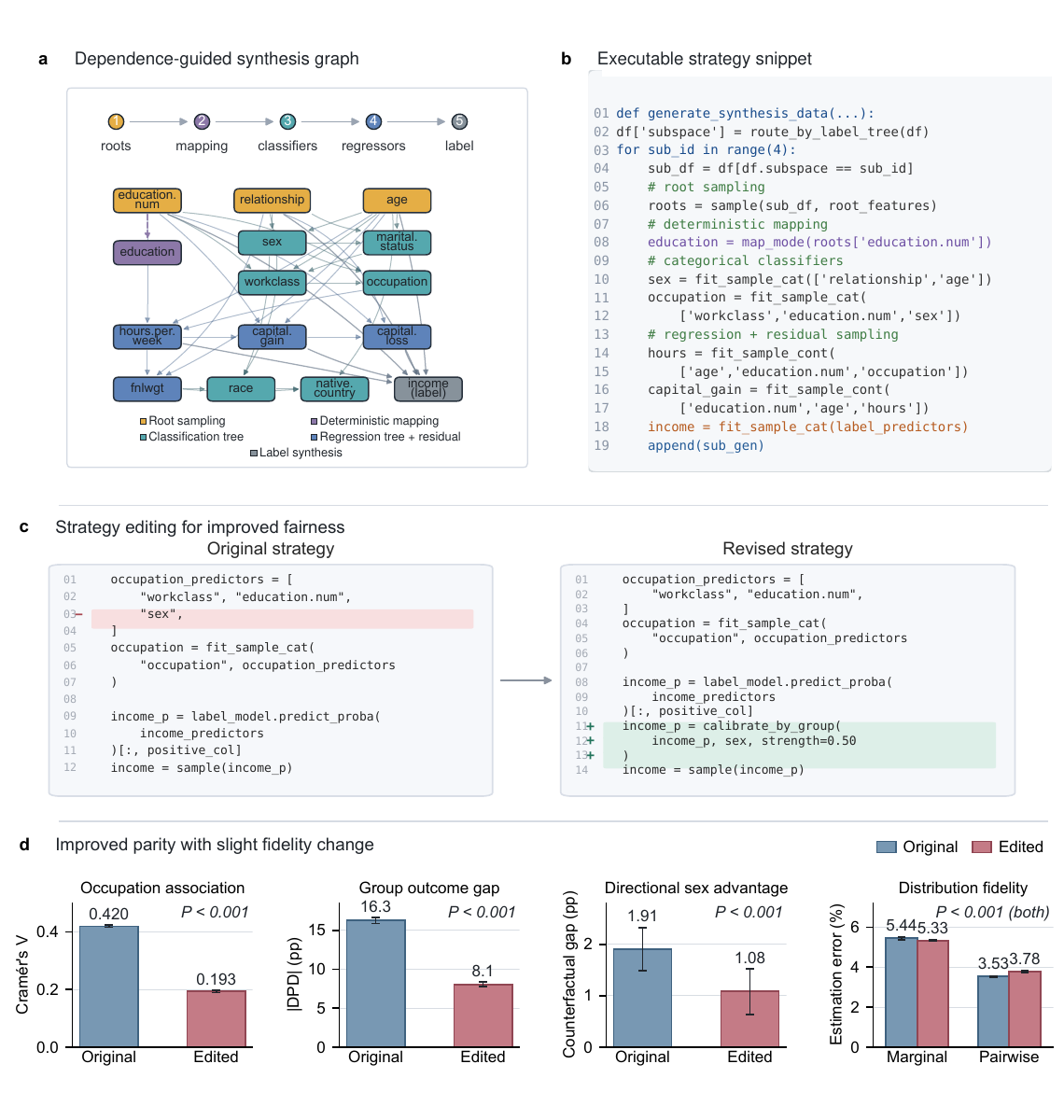}
\caption{
\textbf{Distilling variable dependence into executable synthesis strategies with human-in-the-loop fairness editing.}
\paneltarget{fig:adult_vis}{a} Variable-dependence graph distilled by TabSSD for the \textit{Adult} dataset. Nodes denote variables, directed edges encode the predictive conditional dependence used by the synthesis program, and node colours indicate the models assigned to variables. 
\paneltarget{fig:adult_vis}{b} An excerpt from the executable synthesis strategy generated by TabSSD. Each record is routed to a tree-induced subspace, within which variables are synthesized with the models and predictors specified by the distilled dependence graph.
\paneltarget{fig:adult_vis}{c} Human-in-the-loop editing of the executable strategy.  \texttt{Sex} is removed from the predictors of \texttt{Occupation} and sex-aware calibration is used in the generated probability of high income. 
The edited strategy executes locally and requires no additional LLM query.
\paneltarget{fig}{d} Fairness and fidelity comparison between the original and edited strategies. Fairness is quantified using Cramér's~\(V\) for the association between \texttt{Sex} and \texttt{Occupation}, the absolute demographic-parity difference in the generated high-income outcome, and the directional counterfactual sex gap estimated by a random-forest income auditor. Fidelity is quantified using marginal-distribution and pairwise-correlation errors. Lower values indicate better parity or higher fidelity. Bars and error bars show the mean and s.d., respectively.  \(P\) values are reported for two-sided paired \(t\)-tests. 
}
\label{fig:adult_vis}
\end{figure}

\subsection*{Transparent data synthesis and human-in-the-loop auditing}

\textit{From variable dependence to transparent code synthesis.} TabSSD produces inspectable and executable Python programs. We illustrate this procedure using \textit{Adult}, a census dataset commonly used to predict whether annual income exceeds 50,000 U.S. dollars from demographic, educational, employment, and financial variables~\cite{adult_2}. To reduce redundancy across the chained trees (Supplementary Fig.~\ref{fig:cart}), the LLM distils the extracted variable dependence into a directed graph. As shown in Fig.\panelref{fig:adult_vis}{a}, the graph designates \texttt{education.num}, \texttt{relationship}, and \texttt{age} as root variables, maps \texttt{education} deterministically from \texttt{education.num}, and assigns an appropriate estimator to each remaining variable---for example, a regression tree for \texttt{capital.gain}. TabSSD renders the resulting strategy as executable Python code, which is shown in Fig.\panelref{fig:adult_vis}{b} with details provided in Supplement~\ref{appendix:adult_case}. This enables domain experts to inspect, audit, and modify the synthesis logic. 
Notably, the LLM does not receive the original variable names; the variable names restored in the figure are solely for illustration.

\textit{Human-in-the-loop interventions.} Generative models can inherit biases in their training data, like sex-related disparities in \textit{Adult}. TabSSD's generated scripts support direct code-level interventions without tree retraining or prompt editing. For example, if a user wants to reduce gender bias in their synthetic data to be shared, they can remove \texttt{sex} from the predictors for \texttt{occupation} and add group-wise probability calibration for \texttt{income} (Fig.\panelref{fig:adult_vis}{c}) without regenerating the data synthesis strategy. We assess the original and edited strategies using Cramér's \(V\) for the association between \texttt{sex} and \texttt{occupation}, the absolute demographic-parity difference (\(\lvert\mathrm{DPD}\rvert\)) between female and male high-income rates, and the counterfactual sex gap obtained by changing \texttt{sex} while holding all other attributes fixed (Fig.\panelref{fig:adult_vis}{d}). Lower values indicate better parity. 
The edit reduces the mean sex--occupation association from \(0.420\) to \(0.193\), the mean absolute DPD from \(16.3\) to \(8.1\) percentage points, and the mean counterfactual sex gap from \(1.91\) to \(1.08\) percentage points, while largely preserving overall statistical fidelity. This experiment illustrates code-level editability rather than prescribing a general fairness intervention.

\textit{Code diagnosis and domain-knowledge injection.} The executable representation further enables targeted diagnosis and the direct incorporation of domain knowledge. Users can correct empirically identified errors at specific lines of Python code (Supplementary Table~\ref{tab:failure_modes}), impose logical constraints such as mutually exclusive medical diagnoses, and insert custom \texttt{if-else} statements to construct targeted stress-test scenarios. By combining variable dependence learned from the data with explicit expert rules, TabSSD makes the synthesis process transparent, auditable, and readily editable.

\begin{figure}[t]
\internallinenumbers
\centering
\includegraphics[width=\linewidth]{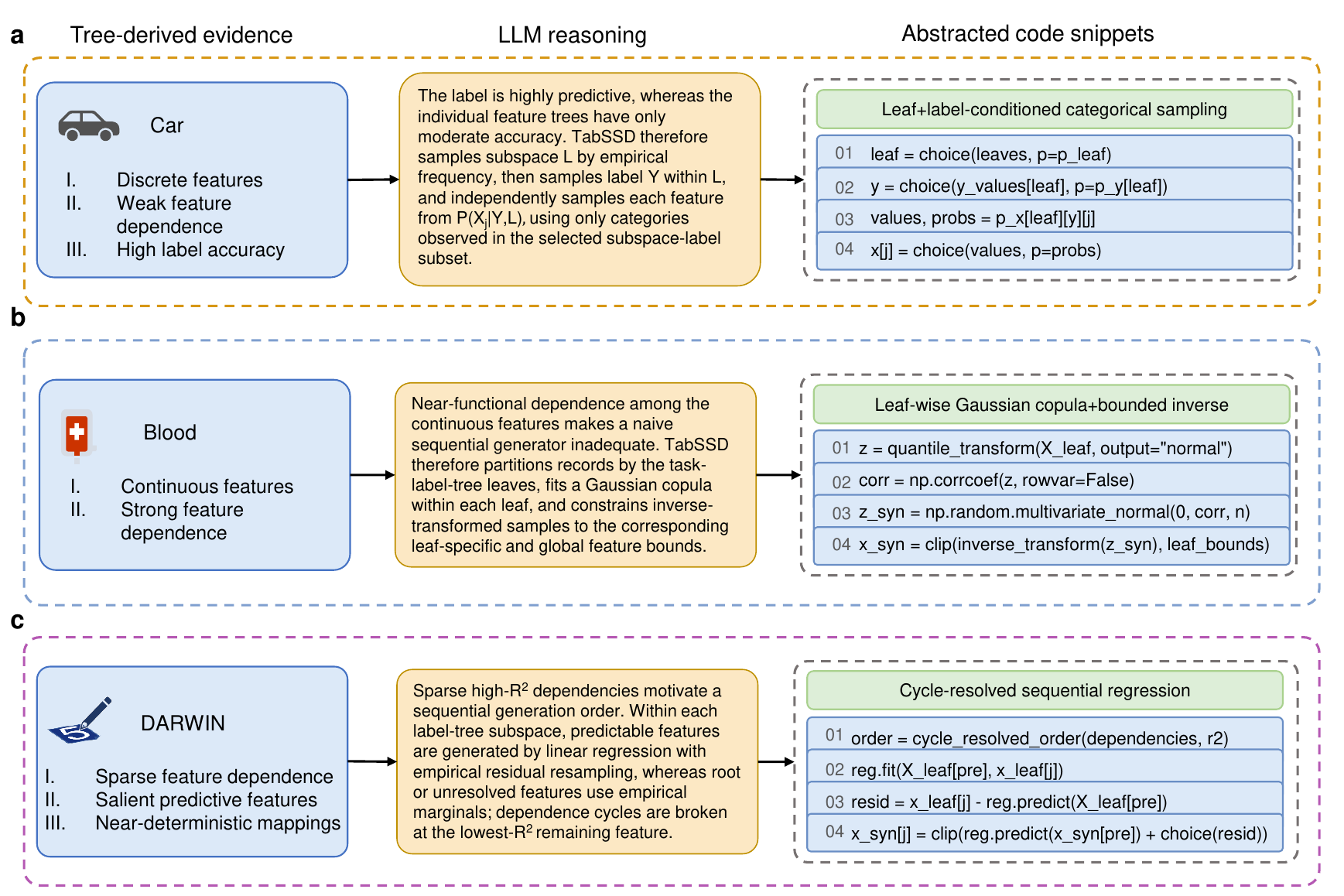}
\caption[Dataset-adaptive synthesis strategies for representative tabular datasets.]{
\textbf{Dataset-adaptive synthesis strategies guided by variable types and dependence from chained trees.}
\textbf{a}, For \textit{Car}, weak inter-feature dependence coupled with strong label predictability leads to categorical sampling conditioned on both the label-tree leaf and class label.
\textbf{b}, For \textit{Blood}, strong dependence among continuous variables motivates a leaf-wise Gaussian copula followed by a bounded inverse transformation.
\textbf{c}, For \textit{DARWIN}, sparse dependence combined with near-deterministic relationships motivates cycle-resolved sequential regression with empirical residual resampling.
}
\label{fig:adapt}
\end{figure}

\subsection*{Tailored data synthesis strategies}

We examine three datasets with different structures: \textit{Car}, which contains only categorical features; \textit{Blood}, which contains only continuous features; and \textit{DARWIN}, a high-dimensional, small-sample dataset. The same prompting and validation pipeline is used to produce synthesis strategies. Fig.~\ref{fig:adapt} shows that TabSSD adapts its synthesis procedure to each dataset rather than applying a fixed template. Specifically, TabSSD combines empirical sampling, node-conditional categorical sampling, Gaussian copulas, and cycle-resolved sequential regression. 

For \textit{Car}, the chained trees show weak feature dependence but high label predictability. TabSSD first samples a label within a selected leaf according to its empirical distribution and then samples each feature conditional on the leaf and label. For \textit{Blood}, the variables exhibit near-deterministic functional dependence. TabSSD fits a Gaussian copula within each label-tree leaf, transforms the samples back to the original scale, and constrains them to the corresponding variable ranges. For \textit{DARWIN}, variable dependence is sparse. TabSSD uses predictive variable dependence and the associated \(R^2\) values to establish a sequential generation order, breaking a dependence cycle at the feature with the smallest \(R^2\). Within each label-tree leaf, ordinary least squares regression with empirical residual resampling generates predictable features. Leaf-level or global empirical distributions supply root and unmodelled features, and the leaf-level empirical distribution supplies labels. Supplement~\ref{appendix:tailored} provides the corresponding Python implementations.

\subsection*{Clinically robust augmentation under class imbalance}
\label{appendix:augmentation}
\begin{figure}[t]
\internallinenumbers
\centering
\includegraphics[width=\linewidth]{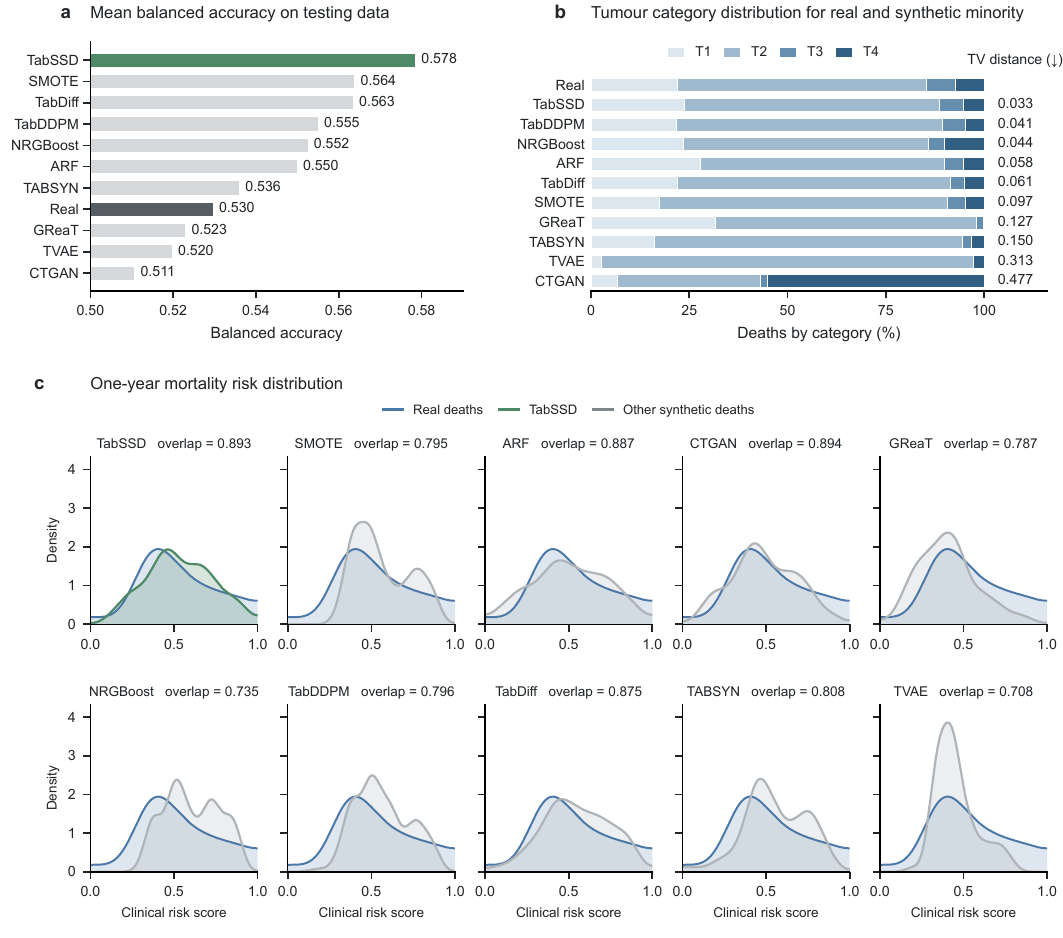}
\caption[Thoracic-surgery case study of clinical phenotype preservation and predictive utility.]{
\textbf{Clinical phenotype preservation and mortality-prediction utility under \(2\times\) augmentation in the \textit{Thoracic Surgery} case study.}
\paneltarget{fig:augmentation}{a} Mean balanced accuracy on test data. AdaBoost, random forest, and XGBoost are trained on augmented training data, and the results are averaged across the three classifiers.
\paneltarget{fig:augmentation}{b} Tumour category distributions among 41 real deaths in training data and 458 synthetic deaths for each synthesis method. Lower total-variation (TV) distance indicates better recovery of the real tumour phenotype distribution among the minority.
\paneltarget{fig:augmentation}{c} Cross-fitted one-year mortality-risk distributions for the real and synthetic data. Five class-weighted logistic-regression models are cross-fitted on the real training data; real records receive out-of-fold scores, and synthetic records receive scores averaged across the five models. Higher density overlap indicates better recovery of the minority risk profile.
}
\label{fig:augmentation}
\end{figure}

We examine TabSSD for data augmentation to address imbalanced classification in a clinically motivated setting. Specifically, we investigate whether TabSSD-generated data can improve minority identification under class imbalance while preserving the clinical phenotype distribution. We conduct this evaluation using the \textit{Thoracic Surgery} dataset~\cite{thoracic_surgery_data_277} that comprises 470 patients, including 70 deaths within one year, and 16 predictors. Death within one year serves as the binary outcome, with deceased patients forming the minority class. We use 270 samples, including 41 deaths, for training, and the other 200, including 29 deaths, for testing. 
Each data synthesis method generates 458 synthetic samples per class, corresponding to twice the number of the majority in the training data, and data synthesis augments the training data. We show representative results in Fig.~\ref{fig:augmentation}. 
Fig.\panelref{fig:augmentation}{a} shows that TabSSD achieves the highest mean balanced accuracy (0.578) among the evaluated methods, using AdaBoost, random forest, and XGBoost as classifiers. 
This result indicates that TabSSD is effective for data augmentation under class imbalance, improving minority-class detection.
We defer detailed results to Supplementary Table~\ref{tab:thoracic_predictive_utility}.
We next assess synthetic data's preservation of the tumour category distribution. A tumour category (T1, T2, T3, or T4) represents the local extent of invasion and relates directly to disease severity. As shown in Fig.\panelref{fig:augmentation}{b}, TabSSD most closely reproduces the observed tumour-category distribution with a total-variation distance of 0.033, achieving the best distribution recovery among the compared methods. It retains the predominance of T2 while preserving the less frequent T3 and T4.

Finally, we evaluate whether the synthetic data generated by TabSSD and the baseline methods preserve the predictive characteristics of the minority class. Specifically, we examine how closely the synthetic records corresponding to patients who die within one year recover the distribution of one-year mortality risk observed in real data. We estimate these risks using class-weighted logistic regression trained on the real data. To obtain out-of-sample predictions for the real records, we perform five-fold cross-fitting; four folds are used for model training and one fold for prediction. A synthetic record's prediction is averaged over the five models. We then compare the risk-score distributions of the real and synthetic minority-class records using the overlap coefficient, which measures the shared area between their density curves; values closer to 1 indicate better recovery of the real data. Fig.\panelref{fig:augmentation}{c} shows that TabSSD achieves an overlap coefficient of 0.893, very close to the best-performing CTGAN (0.894), demonstrating that TabSSD well preserves the data distribution and delivers strong predictive utility.

\section*{Discussion}
In this work, we introduce TabSSD, which repositions LLMs from record-level data generators to synthesis strategy designers. TabSSD separates the local extraction of variable-dependence structure from strategy design and synthesis execution: chained trees summarize variable dependence in the training data, the LLM translates the summary into executable synthesis programs, and candidate programs are executed and validated within the data holder's environment. Across 12 benchmark datasets, this strategy-level formulation strikes a favourable balance among statistical fidelity, downstream predictive utility, empirical privacy risk, and computational efficiency. The case studies of fairness editing and clinical data augmentation further illustrate how exposing the synthesis logic as executable code enables targeted intervention and application-specific validation. Unlike extant methods that encode the generator in learned parameters or read and generate records token by token, TabSSD produces an inspectable computational procedure while avoiding the transmission of raw data to the LLM during strategy generation.

Our findings have practical and broader implications. TabSSD lowers the expertise and infrastructure barriers to tabular synthesis by automating dependence extraction, strategy design, local validation, and candidate selection. Through a standardized interface, it produces executable programs without requiring users to design deep generative models, fine-tune LLMs, perform extensive hyperparameter tuning, or undertake local GPU-accelerated neural network training. More broadly, TabSSD suggests that, for structured data, LLMs may be more valuable as designers of synthesis procedures than as direct record generators. Tree-derived summaries bridge private, order-invariant tables and general-purpose LLMs, while local statistical tools handle data access, model fitting, sampling, and evaluation. Representing the synthesis mechanism as executable code also enables failures to be traced, domain constraints to be incorporated, and strategies to be revised without retraining. This modular separation of data structure, LLM reasoning, and local verification provides a basis for human-governed generative AI.

Several limitations delineate the scope of the conclusions, although some reflect broader challenges common to current LLM-based systems rather than TabSSD alone. TabSSD depends on both the adequacy of tree-derived summaries and the reliability of the backbone LLM. Chained trees may miss weak or higher-order variable dependence, and stochastic LLM outputs may yield invalid, inefficient, or low-fidelity synthesis routines. We mitigate these risks through multi-candidate generation, local execution, and code inspection (Supplementary Table~\ref{tab:failure_modes}). Future work could further incorporate execution feedback, iterative program repair, and verifier-guided search over alternative strategies~\citep{madaan2023self,shinn2023reflexion,yao2023tree,yao2022react,lightman2024let}. Scalability to ultra-high-dimensional tables can be limited by LLM context windows. Step-specific retrieval-augmented context construction~\citep{lewis2020retrieval} and graph-based decomposition~\citep{velickovic2018graph} could restrict reasoning to the dependence relevant to each synthesis decision. Moreover, future work can further reduce information leakage by integrating formal privacy mechanisms into dependence extraction and strategy validation. Human-centred studies are also needed to assess whether practitioners can reliably understand, audit, and modify generated strategies in realistic settings. Taken together, our findings suggest that LLMs may be particularly valuable as designers of transparent, locally executable procedures rather than as unconstrained generators of sensitive structured data.

\clearpage

\section*{Methods}
\label{sec:methods}

\subsection*{Problem setup}
\label{sec:problem_setup}



Let \(D_{\mathrm{train}}=\{(\mathbf{X}_i,Y_i)\}_{i=1}^{N}\) denote an
original tabular dataset with \(N\) samples, where
\(\mathbf{X}_i=(X_{i1},\ldots,X_{iJ})\) and \(Y_i\) denote the feature
vector and target value of the \(i\)-th sample, respectively. The dataset
comprises \(J+1\) variables: \(J\) feature variables
\(X_1,\ldots,X_J\) and one target variable, or label, \(Y\). Here, \(X_{ij}\)
denotes the value of feature variable \(X_j\) for the \(i\)-th sample.
The goal of tabular data synthesis is to construct a generator \({G}\) that synthesizes table \(\hat{{D}}\) from the original data ${D_{train}}$, $\hat{{D}} \leftarrow {G}({D_{train}}).$
The synthetic data should preserve statistical properties of the original dataset, such as marginal distributions and variable dependence, without compromising privacy.
For supervised tasks, we distinguish between the features $\mathbf{X}_i$ and the target $Y_i$, such as a continuous target for regression tasks or a categorical target for classification tasks. If the data holder does not specify a task for the tabular data, $Y_i$ is omitted, and the workflow of TabSSD remains unchanged.

TabSSD separates the learning of variable-dependence structure from synthesis execution. The framework first extracts variable-dependence structure from \(D_{\mathrm{train}}\) using chained trees. It then encodes this information into a compact prompt \(p\), which is provided to an LLM to generate $K$ candidate synthesis strategies \(\{S_k\}_{k=1}^{K}\). Each candidate strategy is executed locally to produce a candidate synthetic table \(\hat{D}_k\). The final strategy is selected based on fidelity, empirical privacy risk, runtime criteria, and potentially other criteria. 
This design changes the role of the LLM from a record-level generator to a synthesis strategy designer. The LLM receives summaries of the variable-dependence structure rather than raw records, and the original data are used only within the local environment for chained tree training, strategy execution, and candidate evaluation.
The workflow is 
summarized in Algorithm~\ref{alg:framework}.

\subsection*{Variable dependence extraction, encoding, and prompt design}
\label{sec:dependency_mining}

Tabular synthesis requires preserving variables' marginal distributions and
dependence to maintain statistical fidelity and downstream utility.
Because it is inherently difficult for an LLM to reason about variable dependence given tabular data, we use chained trees to capture
variable relationships without imposing parametric
assumptions.
Concretely, we construct chained trees $T_{\mathrm{chain}}
=
\{\mathrm{Tree}_{X_1},\ldots,\mathrm{Tree}_{X_J},
\mathrm{Tree}_Y\}$ and train a feature tree \(\mathrm{Tree}_{X_j}\) to predict each feature \(X_j\)
from all other features, denoted by  \(\mathbf{X}_{\setminus j}\), and the label tree \(\mathrm{Tree}_Y\) to predict \(Y\) from
\(\mathbf{X}\):
\begin{equation}
\mathrm{Tree}_{X_j}:
\mathbf{X}_{\setminus j}\rightarrow X_j,
\qquad
\mathrm{Tree}_Y:
\mathbf{X}\rightarrow Y.
\end{equation}

For each \(\mathrm{Tree}_{X_j}\), we denote by \({A}(X_j)\) the set of split variables. The variables in \({A}(X_j)\)  are ordered according to the minimum tree depth where they are first split, as variables splitting closer to the root define broader partitions of the data and are more influential in predicting $X_j$.
We denote the goodness of fit of \(\mathrm{Tree}_{X_j}\) 
by \(\phi(\mathrm{Tree}_{X_j})\), which contains the coefficient of determination (\(R^2\)) and root-mean-square error (RMSE) for continuous
variables and prediction accuracy for categorical variables:
\begin{equation}
\phi(\mathrm{Tree}_{X_j})=
\begin{cases}
\big(R^2(\mathrm{Tree}_{X_j}), \mathrm{RMSE}(\mathrm{Tree}_{X_j})\big), & \text{if } X_j \text{ is continuous},\\[4pt]
\mathrm{Acc}(\mathrm{Tree}_{X_j}), & \text{if } X_j \text{ is categorical}.
\end{cases}
\end{equation}


Directly incorporating all decision rules into the LLM prompt would introduce substantial overhead and rapidly inflate the prompt length, particularly for high-dimensional datasets. We encode each $\mathrm{Tree}_{X_j}$'s decision rule into compact dependence signals, including variable types and ranges,  split variables \({A}(X_j)\), and goodness-of-fit scores \(\phi(\mathrm{Tree}_{X_j})\). For \(\mathrm{Tree}_{Y}\),
we retain its split rules and the distribution of $Y$ in leaf nodes. 
For high-dimensional datasets, even the compact dependence encoding can exceed the LLM's context-window limit. To enhance scalability, we apply a greedy set-cover procedure. At each iteration, we select the $\mathrm{Tree}_{X_j}$ whose variable set ${A}(X_j)\cup\{X_j\}$ covers the largest number of previously uncovered features, and repeat this process until all features are covered. We always retain $\mathrm{Tree}_{Y}$ when $Y$ is defined.
Given the variable-dependence structure extracted by the chained trees, we construct a structured prompt that guides the LLM to reason about variable relationships and generate an executable synthesis strategy. The prompt \(p\) is composed of five  components:
\begin{equation}
p =
p_{\mathrm{task}}
\oplus
p_{\mathrm{metric}}
\oplus
p_{\mathrm{requirement}}
\oplus
p_{\mathrm{dep}}
\oplus
p_{\mathrm{output}},
\label{eq:prompt_composition}  
\end{equation}    
where \(\oplus\) denotes textual concatenation.  
\(p_{\mathrm{task}}\) is the task definition that informs the LLM that the goal is to generate an executable tabular data synthesis strategy rather than directly populating data records. 
\(p_{\mathrm{metric}}\) is the guidance on 
evaluation metrics. It tells the LLM that the generated synthetic data will be evaluated according to statistical fidelity, downstream machine-learning utility, and empirical privacy-related risk. This component does not directly optimize these metrics but encourages the LLM to reason about trade-offs among the metrics when designing the synthesis strategy.
\(p_{\mathrm{requirement}}\) instructs the LLM to analyse \(p_{\mathrm{dep}}\) to infer variable dependence and distributional characteristics. It also encourages dividing heterogeneous data into more homogeneous subspaces when appropriate, thereby simplifying the modelling of each subspace.
\(p_{\mathrm{dep}}\) contains information about the variable-dependence structure extracted by the chained trees.  
\(p_{\mathrm{output}}\) provides both an output template and a worked in-context example. The template requires each candidate synthesis strategy to be expressed as an executable Python program with the following standardized interface:
\[
\texttt{generate\_synthesis\_data(save\_path, n\_sample, X\_train, y\_train)}.
\]
During local execution, \texttt{X\_train} and \texttt{y\_train} provide the training data, \texttt{n\_sample} specifies the number of records, and \texttt{save\_path} specifies the output location. The function saves \texttt{n\_sample} records matching the training schema. 
An example prompt is provided in Supplement~\ref{appendix:adult_original_prompt}.

\subsection*{Synthesis strategy validation and selection}

\label{sec:llm_strategy_designer}

Given the prompt $p$, we query the LLM multiple times to generate a set of candidate synthesis strategies.
Because LLM outputs are stochastic and the extracted dependence information can support multiple valid synthesis approaches, these candidates often vary in their variable-generation order, subspace partitioning, and choice of statistical or machine learning models. Consequently, the candidates exhibit disparate statistical fidelity, empirical utility, privacy risk, and execution time. Generating multiple candidates increases strategy diversity, allowing subsequent validation and selection to filter out suboptimal or non-executable pipelines. The impact of the number of LLM queries on overall synthesis performance is detailed in Supplementary Fig.~\ref{fig:query_n}.

Each candidate strategy is executed locally to generate a synthetic table; any strategy that fails to execute is discarded. We then select a synthesis strategy through a filtering step prioritizing privacy and computational efficiency, followed by an evaluation step based on statistical fidelity. During filtering, we discard strategies whose execution time exceeds a predefined limit or whose resulting synthetic table exhibits a $\delta$-presence privacy risk above an acceptable threshold. The remaining
strategies constitute the feasible candidate set
\(\Omega_{\mathrm{feas}}\). We then compute the marginal distribution error and pairwise correlation error for the remaining synthetic tables, ultimately selecting the strategy that minimizes the combined errors:
\begin{equation}
S^*
=
\underset{S\in{\Omega}_{\mathrm{feas}}}{\arg\min}\;
E\bigl(\hat{D}(S),D_{\mathrm{train}}\bigr),
\end{equation}
where \({E}\) is the average of the marginal distribution error and pairwise correlation error. Finally, 
\({S}^{*}\) or its synthetic table can be published or used for downstream tasks. 

\begin{algorithm}[t]
\caption{LLM-guided strategy-level tabular data synthesis}
\label{alg:framework}
\begin{algorithmic}[1]

\Statex \textbf{Notation:}
\(D_{\mathrm{train}}=\{(\mathbf{X}_i,Y_i)\}_{i=1}^{N}\), where
\(\mathbf{X}_i=(X_{i1},\ldots,X_{iJ})\).
\(\mathrm{Tree}_{X_j}\) predicts \(X_j\) from
\(\mathbf{X}_{\setminus j}\), whereas
\(\mathrm{Tree}_{Y}\) predicts \(Y\) from \(\mathbf{X}\).

\Statex \(A(X_j)\) denotes the split-variable set of
\(\mathrm{Tree}_{X_j}\);
\(\phi(\mathrm{Tree}_{X_j})\) denotes its goodness-of-fit statistics;
\(\operatorname{Exec}(S;D,n)\) executes strategy \(S\) on \(D\)
to generate \(n\) samples; and
\(E(\hat{D},D)\) denotes the statistical error.
If no target is specified, \(Y\) and \(\mathrm{Tree}_{Y}\) are omitted.

\Require Training data \(D_{\mathrm{train}}\);
number of synthetic samples \(n_{\mathrm{sample}}\);
number of LLM queries \(K\);
\(\delta\)-presence risk threshold \(\tau_\delta=10\);
execution-time threshold \(\tau_T=100\,\mathrm{s}\)

\State Define the synthesis task, evaluation objectives, and output requirements

\State Construct
\(
\mathcal{T}_{\mathrm{chain}}
\leftarrow
\{\mathrm{Tree}_{X_1},\ldots,\mathrm{Tree}_{X_J},
\mathrm{Tree}_{Y}\}
\)
from \(D_{\mathrm{train}}\)

\State Extract
\(
\{A(X_j),\phi(\mathrm{Tree}_{X_j})\}_{j=1}^{J}
\)
and, when present, the split structure and leaf-level target
distributions of \(\mathrm{Tree}_{Y}\)

\State
\(
\mathcal{T}_{\mathrm{enc}}
\leftarrow
\mathcal{T}_{\mathrm{chain}}
\)

\If{\(D_{\mathrm{train}}\) is high-dimensional}

    \State
    \(
    \mathcal{T}_{\mathrm{sel}}\leftarrow\emptyset,
    \quad
    \mathcal{U}\leftarrow\emptyset
    \)

    \While{\(\mathcal{U}\neq\{X_1,\ldots,X_J\}\)}

        \State Select the tree covering the most previously
        uncovered variables:
        \[
        j^{*}\leftarrow
        \underset{
        j:\mathrm{Tree}_{X_j}\notin\mathcal{T}_{\mathrm{sel}}
        }{\arg\max}
        \left|
        \bigl(A(X_j)\cup\{X_j\}\bigr)\setminus\mathcal{U}
        \right|
        \]

        \State
        \(
        \mathcal{T}_{\mathrm{sel}}
        \leftarrow
        \mathcal{T}_{\mathrm{sel}}
        \cup
        \{\mathrm{Tree}_{X_{j^{*}}}\}
        \)

        \State
        \(
        \mathcal{U}
        \leftarrow
        \mathcal{U}
        \cup A(X_{j^{*}})
        \cup\{X_{j^{*}}\}
        \)

    \EndWhile

    \State
    \(
    \mathcal{T}_{\mathrm{enc}}
    \leftarrow
    \mathcal{T}_{\mathrm{sel}}
    \cup\{\mathrm{Tree}_{Y}\}
    \)

\EndIf

\State Encode the information retained in
\(\mathcal{T}_{\mathrm{enc}}\) as \(p_{\mathrm{dep}}\)

\State Construct the structured prompt
\[
p \leftarrow
p_{\mathrm{task}}
\oplus p_{\mathrm{metric}}
\oplus p_{\mathrm{requirement}}
\oplus p_{\mathrm{dep}}
\oplus p_{\mathrm{output}}
\]

\State initialize the feasible strategy set
\(\Omega_{\mathrm{feas}}\leftarrow\emptyset\)

\Repeat
    \For{\(k=1,\ldots,K\)}
        \State Query the LLM with \(p\) to generate
        a candidate strategy \(S_k\)

        \State Attempt
        \(\hat D(S_k)\leftarrow
        \operatorname{Exec}(S_k;D_{\mathrm{train}},n_{\mathrm{sample}})\);
        skip non-executable or invalid candidates

         \State Evaluate the statistical-fidelity error
        \(E\bigl(\hat D(S_k),D_{\mathrm{train}}\bigr)\),
        \(\delta\)-presence risk
        \(R_\delta\bigl(\hat D(S_k),D_{\mathrm{train}}\bigr)\),
        and execution time \(T(S_k)\)

        \If{\(R_\delta\bigl(\hat D(S_k),D_{\mathrm{train}}\bigr)
        \leq\tau_\delta\) \textbf{and} \(T(S_k)\leq\tau_T\)}
            \State \(\Omega_{\mathrm{feas}}\leftarrow
            \Omega_{\mathrm{feas}}\cup\{S_k\}\)
        \EndIf
    \EndFor
\Until{\(\Omega_{\mathrm{feas}}\neq\emptyset\)}
Raw records are not transmitted

\State Select
\[
S^*
\leftarrow
\underset{S\in\Omega_{\mathrm{feas}}}{\arg\min}\;
E\bigl(\hat D(S),D_{\mathrm{train}}\bigr)
\]

\State
\[
\hat D^*
\leftarrow
\hat D(S^*)
\]

\State \Return \(S^*\) and \(\hat D^*\)
\end{algorithmic}
\end{algorithm}
\FloatBarrier




\section*{Author contributions}
J.L., H.Y., and D.G. conceived the study. J.L., H.Y., D.G., and Q.Z. contributed substantially to the design of various aspects of the study. J.L. developed the code and performed the experiments. J.L., H.Y., D.G., and Q.Z. analysed and interpreted the experimental results. All authors contributed to the critical revision of the manuscript and reviewed and approved the final version.

\section*{Competing interests}
The authors declare no competing interests.

\clearpage

\bibliography{references}

@inproceedings{
    ganev2026smote,
    title={{SMOTE} and Mirrors: Exposing Privacy Leakage from Synthetic Minority Oversampling},
    author={Georgi Ganev and MohammadReza Nazari and Rees Davison and Amirhassan Fallah Dizche and XINMIN Wu and Ralph Abbey and Jorge G Silva and Emiliano De Cristofaro},
    booktitle={The Fourteenth International Conference on Learning Representations},
    year={2026},
    url={https://openreview.net/forum?id=ZQSZMpsQKj}
}

@article{dsa2023prediction,
  author  = {D'Sa, Karishma and Evans, James R. and Virdi, Gurvir S. and Vecchi, Giulia and Adam, Alexander and Bertolli, Ottavia and Fleming, James and Chang, Hojong and Leighton, Craig and Horrocks, Mathew H. and Athauda, Dilan and Choi, Minee L. and Gandhi, Sonia},
  title   = {Prediction of mechanistic subtypes of {Parkinson's} using patient-derived stem cell models},
  journal = {Nature Machine Intelligence},
  year    = {2023},
  month   = aug,
  volume  = {5},
  number  = {8},
  pages   = {933--946},
  issn    = {2522-5839},
  doi     = {10.1038/s42256-023-00702-9},
  url     = {https://doi.org/10.1038/s42256-023-00702-9}
}

@article{Gamella2025,
  author  = {Gamella, Juan L. and Peters, Jonas and B{\"u}hlmann, Peter},
  title   = {Causal chambers as a real-world physical testbed for {AI} methodology},
  journal = {Nature Machine Intelligence},
  year    = {2025},
  volume  = {7},
  number  = {1},
  pages   = {107--118},
  month   = jan,
  doi     = {10.1038/s42256-024-00964-x},
  url     = {https://doi.org/10.1038/s42256-024-00964-x},
  issn    = {2522-5839}
}

@article{gdpr2016,
  author  = {{European Parliament} and {Council}},
  title   = {General data protection regulation},
  journal = {Off. J. Eur. Union},
  volume  = {L 119},
  pages   = {1--88},
  year    = {2016}
}

@inproceedings{xu2019modeling,
 author = {Xu, Lei and Skoularidou, Maria and Cuesta-Infante, Alfredo and Veeramachaneni, Kalyan},
 booktitle = {Advances in Neural Information Processing Systems},
 title = {Modeling Tabular data using {Conditional GAN}},
 volume = {32},
 year = {2019}
}

@techreport{schmidhuber1990making,
  author      = {Schmidhuber, J{\"u}rgen},
  title       = {Making the World Differentiable: On Using Fully Recurrent Self-Supervised Neural Networks for Dynamic Reinforcement Learning and Planning in Non-Stationary Environments},
  institution = {Institut f{\"u}r Informatik, Technische Universit{\"a}t M{\"u}nchen},
  number      = {FKI-126-90},
  year        = {1990},
  type        = {Technical Report}
}

@inproceedings{schmidhuber1991possibility,
  title={A possibility for implementing curiosity and boredom in model-building neural controllers},
  author={Schmidhuber, J{\"u}rgen},
  booktitle={Proc. of the international conference on simulation of adaptive behavior: From animals to animats},
  pages={222--227},
  year={1991}
}

@article{jarzynski1997equilibrium,
  title = {Equilibrium free-energy differences from nonequilibrium measurements: A master-equation approach},
  author = {Jarzynski, C.},
  journal = {Phys. Rev. E},
  volume = {56},
  issue = {5},
  pages = {5018--5035},
  numpages = {0},
  year = {1997},
  month = {Nov},
  publisher = {American Physical Society},
  doi = {10.1103/PhysRevE.56.5018},
  url = {https://link.aps.org/doi/10.1103/PhysRevE.56.5018}
}

@inproceedings{
    borisovlanguage,
    title={Language Models are Realistic Tabular Data Generators},
    author={Vadim Borisov and Kathrin Sessler and Tobias Leemann and Martin Pawelczyk and Gjergji Kasneci},
    booktitle={The Eleventh International Conference on Learning Representations },
    year={2023}
}

@incollection{reynolds2009gaussian,
author="Reynolds, Douglas",
editor="Li, Stan Z.
and Jain, Anil",
title="Gaussian Mixture Models",
bookTitle="Encyclopedia of Biometrics",
year="2009",
publisher="Springer US",
address="Boston, MA",
pages="659--663",
isbn="978-0-387-73003-5",
doi="10.1007/978-0-387-73003-5_196"
}

@article{nowok2016synthpop,
  title={synthpop: Bespoke creation of synthetic data in {R}},
  author={Nowok, Beata and Raab, Gillian M and Dibben, Chris},
  journal={Journal of Statistical Software},
  volume={74},
  pages={1--26},
  year={2016}
}

@inproceedings{
    shi2025tabdiff,
    title={{TabDiff}: a Mixed-type Diffusion Model for Tabular Data Generation},
    author={Juntong Shi and Minkai Xu and Harper Hua and Hengrui Zhang and Stefano Ermon and Jure Leskovec},
    booktitle={The Thirteenth International Conference on Learning Representations},
    year={2025},
    url={https://openreview.net/forum?id=swvURjrt8z}
}

@inproceedings{goodfellow2014generative,
  title     = {Generative Adversarial Nets},
  author    = {Goodfellow, Ian J. and Pouget-Abadie, Jean and Mirza, Mehdi and Xu, Bing and Warde-Farley, David and Ozair, Sherjil and Courville, Aaron and Bengio, Yoshua},
  booktitle = {Advances in Neural Information Processing Systems},
  volume    = {27},
  pages     = {2672--2680},
  year      = {2014}
}

@article{yan2022multifaceted,
  title   = {A multifaceted benchmarking of synthetic electronic health record generation models},
  author  = {Yan, Chao and Yan, Yao and Wan, Zhiyu and Zhang, Ziqi and Omberg, Larsson and Guinney, Justin and Mooney, Sean D. and Malin, Bradley A.},
  journal = {Nature Communications},
  volume  = {13},
  number  = {1},
  pages   = {7609},
  year    = {2022},
  doi     = {10.1038/s41467-022-35295-1}
}

@article{vanbreugel2024synthetic,
  title   = {Synthetic data in biomedicine via generative artificial intelligence},
  author  = {{van Breugel}, Boris and Liu, Tennison and Oglic, Dino and {van der Schaar}, Mihaela},
  journal = {Nature Reviews Bioengineering},
  volume  = {2},
  number  = {12},
  pages   = {991--1004},
  year    = {2024},
  doi     = {10.1038/s44222-024-00245-7}
}

@inproceedings{ho2020denoising,
  title     = {Denoising Diffusion Probabilistic Models},
  author    = {Ho, Jonathan and Jain, Ajay and Abbeel, Pieter},
  booktitle = {Advances in Neural Information Processing Systems},
  volume    = {33},
  pages     = {6840--6851},
  year      = {2020}
}

@inproceedings{sohl2015deep,
  title     = {Deep Unsupervised Learning Using Nonequilibrium Thermodynamics},
  author    = {Sohl-Dickstein, Jascha and Weiss, Eric and Maheswaranathan, Niru and Ganguli, Surya},
  booktitle = {Proceedings of the 32nd International Conference on Machine Learning},
  series    = {Proceedings of Machine Learning Research},
  volume    = {37},
  pages     = {2256--2265},
  year      = {2015}
}

@inproceedings{kotelnikov2023tabddpm,
  title={{TabDDPM}: Modelling tabular data with diffusion models},
  author={Kotelnikov, Akim and Baranchuk, Dmitry and Rubachev, Ivan and Babenko, Artem},
  booktitle={International Conference on Machine Learning},
  pages={17564--17579},
  year={2023},
  organization={PMLR}
}

@article{chawla2002smote,
  title={{SMOTE}: synthetic minority over-sampling technique},
  author={Chawla, Nitesh V and Bowyer, Kevin W and Hall, Lawrence O and Kegelmeyer, W Philip},
  journal={Journal of Artificial Intelligence Research},
  volume={16},
  pages={321--357},
  year={2002}
}

@article{blagus2013smote,
  title   = {{SMOTE} for High-Dimensional Class-Imbalanced Data},
  author  = {Blagus, Rok and Lusa, Lara},
  journal = {BMC Bioinformatics},
  volume  = {14},
  number  = {1},
  pages   = {106},
  year    = {2013},
  doi     = {10.1186/1471-2105-14-106}
}

@article{ohagan2016clustering,
  title   = {Clustering with the Multivariate Normal Inverse Gaussian Distribution},
  author  = {O'Hagan, Adrian and Murphy, Thomas Brendan and Gormley, Isobel Claire and McNicholas, Paul D. and Karlis, Dimitris},
  journal = {Computational Statistics \& Data Analysis},
  volume  = {93},
  pages   = {18--30},
  year    = {2016},
  doi     = {10.1016/j.csda.2014.09.006}
}

@inproceedings{tabsyn,
  title={Mixed-Type Tabular Data Synthesis with Score-based Diffusion in Latent Space},
  author={Zhang, Hengrui and Zhang, Jiani and Srinivasan, Balasubramaniam and Shen, Zhengyuan and Qin, Xiao and Faloutsos, Christos and Rangwala, Huzefa and Karypis, George},
  booktitle={The Twelfth International Conference on Learning Representations},
  year={2024}
}

@inproceedings{watson2023adversarial,
  title={Adversarial random forests for density estimation and generative modeling},
  author={Watson, David S and Blesch, Kristin and Kapar, Jan and Wright, Marvin N},
  booktitle={International Conference on Artificial Intelligence and Statistics},
  pages={5357--5375},
  year={2023},
  organization={PMLR}
}

@inproceedings{yang2024p,
  title={P-TA: Using proximal policy optimization to enhance tabular data augmentation via large language models},
  author={Yang, Shuo and Yuan, Chenchen and Rong, Yao and Steinbauer, Felix and Kasneci, Gjergji},
  booktitle={Findings of the Association for Computational Linguistics: ACL 2024},
  pages={248--264},
  year={2024}
}

@article{zeng2026glm,
  title={Glm-5: from vibe coding to agentic engineering},
  author={Zeng, Aohan and Lv, Xin and Hou, Zhenyu and Du, Zhengxiao and Zheng, Qinkai and Chen, Bin and Yin, Da and Ge, Chendi and Huang, Chenghua and Xie, Chengxing and others},
  journal={arXiv preprint arXiv:2602.15763},
  year={2026}
}

@misc{yang2025qwen3,
    title = {Qwen3-Max: Just Scale it},
    author = {{Qwen Team}},
    month = {September},
    year = {2025}
}

@article{comanici2025gemini,
  title={Gemini 2.5: Pushing the frontier with advanced reasoning, multimodality, long context, and next generation agentic capabilities},
  author={Comanici, Gheorghe and Bieber, Eric and Schaekermann, Mike and Pasupat, Ice and Sachdeva, Noveen and Dhillon, Inderjit and Blistein, Marcel and Ram, Ori and Zhang, Dan and Rosen, Evan and others},
  journal={arXiv preprint arXiv:2507.06261},
  year={2025}
}

@inproceedings{grinsztajn2022tree,
  title={Why do tree-based models still outperform deep learning on typical tabular data?},
  author={Grinsztajn, Leo and Oyallon, Edouard and Varoquaux, Ga{\"e}l},
  booktitle={Advances in Neural Information Processing Systems},
  volume={35},
  pages={507--520},
  year={2022}
}

@article{freund1997decision,
  title={A decision-theoretic generalization of on-line learning and an application to boosting},
  author={Freund, Yoav and Schapire, Robert E},
  journal={Journal of Computer and System Sciences},
  volume={55},
  number={1},
  pages={119--139},
  year={1997},
  publisher={Elsevier}
}

@article{breiman2001random,
  title={Random forests},
  author={Breiman, Leo},
  journal={Machine Learning},
  volume={45},
  number={1},
  pages={5--32},
  year={2001},
  publisher={Springer}
}

@inproceedings{chen2016xgboost,
  title={{XGBoost}: A scalable tree boosting system},
  author={Chen, Tianqi and Guestrin, Carlos},
  booktitle = {Proceedings of the 22nd ACM SIGKDD International Conference on Knowledge Discovery and Data Mining},
  pages={785--794},
  year={2016}
}

@inproceedings{bravo2025nrgboost,
    title={{NRGB}oost: Energy-Based Generative Boosted Trees},
    author={Jo{\~a}o Bravo},
    booktitle={The Thirteenth International Conference on Learning Representations},
    year={2025},
    url={https://openreview.net/forum?id=wQHyjIZ1SH}
}

@article{
    fang2024large,
    title={Large Language Models ({LLM}s) on Tabular Data: Prediction, Generation, and Understanding - A Survey},
    author={Xi Fang and Weijie Xu and Fiona Anting Tan and Ziqing Hu and Jiani Zhang and Yanjun Qi and Srinivasan H. Sengamedu and Christos Faloutsos},
    journal={Transactions on Machine Learning Research},
    issn={2835-8856},
    year={2024},
    url={https://openreview.net/forum?id=IZnrCGF9WI},
}

@article{wu2025tabular,
  title={Tabular data understanding with {LLM}: A survey of recent advances and challenges},
  author={Wu, Xiaofeng and Ritter, Alan and Xu, Wei},
  journal={arXiv preprint arXiv:2508.00217},
  year={2025}
}

@inproceedings{ye2025llm,
  author    = {Ye, Hangting and Li, Jinmeng and Zhao, He
               and Guo, Dandan and Chang, Yi},
  title     = {{LLM} Meeting Decision Trees on Tabular Data},
  booktitle = {Advances in Neural Information Processing Systems},
  volume    = {38},
  pages     = {118095--118131},
  year      = {2025},
  doi       = {10.52202/085713-3938}
}

@inproceedings{hegselmann2023tabllm,
  title={{TabLLM}: Few-shot classification of tabular data with large language models},
  author={Hegselmann, Stefan and Buendia, Alejandro and Lang, Hunter and Agrawal, Monica and Jiang, Xiaoyi and Sontag, David},
  booktitle={International Conference on Artificial Intelligence and Statistics},
  pages={5549--5581},
  year={2023},
  organization={PMLR}
}

@inproceedings{kim2024epic,
  title={EPIC: Effective Prompting for Imbalanced-Class Data Synthesis in Tabular Data Classification via Large Language Models},
  author={Kim, Jinhee and Kim, Taesung and Choo, Jaegul},
  booktitle={Advances in Neural Information Processing Systems},
  volume={37},
  pages={31504--31542},
  year={2024}
}

@inproceedings{seedat2023curated,
  title = 	 {Curated {LLM}: Synergy of {LLM}s and Data Curation for tabular augmentation in low-data regimes},
  author =       {Seedat, Nabeel and Huynh, Nicolas and van Breugel, Boris and van der Schaar, Mihaela},
  booktitle = 	 {Proceedings of the 41st International Conference on Machine Learning},
  pages = 	 {44060--44092},
  year = 	 {2024},

  volume = 	 {235},
  series = 	 {Proceedings of Machine Learning Research},

  url = 	 {https://proceedings.mlr.press/v235/seedat24a.html}

}

@inproceedings{qian2023synthcity,
  title     = {{Synthcity}: A Benchmark Framework for Diverse Use Cases of
               Tabular Synthetic Data},
  author    = {Qian, Zhaozhi and Davis, Rob and {van der Schaar}, Mihaela},
  booktitle = {Advances in Neural Information Processing Systems},
  volume    = {36},
  pages     = {3173--3188},
  year      = {2023},
  doi       = {10.52202/075280-0140}
}

@manual{sdmetrics,
  author = {{DataCebo, Inc.}},
  title  = {Synthetic Data Metrics},
  year   = {2023},
  note   = {Version 0.9.3},
  url    = {https://docs.sdv.dev/sdmetrics/}
}

@inproceedings{
    SDV,
    title={The Synthetic data vault},
    author={Patki, Neha and Wedge, Roy and Veeramachaneni, Kalyan},
    booktitle={IEEE International Conference on Data Science and Advanced Analytics (DSAA)},
    year={2016},
    pages={399-410},
    doi={10.1109/DSAA.2016.49},
    month={Oct}
}

@article{ward2025tables,
  title={When Tables Leak: Attacking String Memorization in LLM-Based Tabular Data Generation},
  author={Ward, Joshua and Gu, Bochao and Wang, Chi-Hua and Cheng, Guang},
  journal={arXiv preprint arXiv:2512.08875},
  year={2025}
}

@inproceedings{carlini2021extracting,
  title={Extracting training data from large language models},
  author={Carlini, Nicholas and Tramer, Florian and Wallace, Eric and Jagielski, Matthew and Herbert-Voss, Ariel and Lee, Katherine and Roberts, Adam and Brown, Tom and Song, Dawn and Erlingsson, Ulfar and others},
  booktitle={30th USENIX security symposium (USENIX Security 21)},
  pages={2633--2650},
  year={2021}
}

@article{rudin2019stop,
  title={Stop explaining black box machine learning models for high stakes decisions and use interpretable models instead},
  author={Rudin, Cynthia},
  journal={Nature Machine Intelligence},
  volume={1},
  number={5},
  pages={206--215},
  year={2019},
  publisher={Nature Publishing Group UK London}
}

@book{breiman1984classification,
  title     = {Classification and Regression Trees},
  author    = {Breiman, Leo and Friedman, Jerome H. and
               Olshen, Richard A. and Stone, Charles J.},
  publisher = {Wadsworth International Group},
  address   = {Belmont, CA},
  year      = {1984}
}

@inproceedings{fang2025understanding,
  title = 	 {Understanding and Mitigating Memorization in Diffusion Models for Tabular Data},
  author =       {Fang, Zhengyu and Jiang, Zhimeng and Chen, Huiyuan and Li, Xiao and Li, Jing},
  booktitle = 	 {Proceedings of the 42nd International Conference on Machine Learning},
  pages = 	 {15976--16005},
  year = 	 {2025},
  volume = 	 {267},
  series = 	 {Proceedings of Machine Learning Research},
  url = 	 {https://proceedings.mlr.press/v267/fang25f.html}
}

@inproceedings{madaan2023self,
  title={Self-Refine: Iterative Refinement with Self-Feedback},
  author={Madaan, Aman and Tandon, Niket and Gupta, Prakhar and Hallinan, Skyler and Gao, Luyu and Wiegreffe, Sarah and Alon, Uri and Dziri, Nouha and Prabhumoye, Shrimai and Yang, Yiming and Gupta, Shashank and Majumder, Bodhisattwa Prasad and Hermann, Katherine and Welleck, Sean and Yazdanbakhsh, Amir and Clark, Peter},
  booktitle={Advances in Neural Information Processing Systems},
  volume={36},
  pages={46534--46594},
  year={2023}
}

@inproceedings{shinn2023reflexion,
  title={Reflexion: Language Agents with Verbal Reinforcement Learning},
  author={Shinn, Noah and Cassano, Federico and Gopinath, Ashwin and Narasimhan, Karthik and Yao, Shunyu},
  booktitle={Advances in Neural Information Processing Systems},
  volume={36},
  pages={8634--8652},
  year={2023}
}

@inproceedings{yao2023tree,
  title={Tree of Thoughts: Deliberate Problem Solving with Large Language Models},
  author={Yao, Shunyu and Yu, Dian and Zhao, Jeffrey and Shafran, Izhak and Griffiths, Tom and Cao, Yuan and Narasimhan, Karthik},
  booktitle={Advances in Neural Information Processing Systems},
  volume={36},
  pages={11809--11822},
  year={2023}
}

@inproceedings{
    yao2022react,
    title={{ReAct}: Synergizing Reasoning and Acting in Language Models},
    author={Shunyu Yao and Jeffrey Zhao and Dian Yu and Nan Du and Izhak Shafran and Karthik R Narasimhan and Yuan Cao},
    booktitle={The Eleventh International Conference on Learning Representations },
    year={2023}
}

@inproceedings{
    lightman2024let,
    title={Let's Verify Step by Step},
    author={Hunter Lightman and Vineet Kosaraju and Yuri Burda and Harrison Edwards and Bowen Baker and Teddy Lee and Jan Leike and John Schulman and Ilya Sutskever and Karl Cobbe},
    booktitle={The Twelfth International Conference on Learning Representations},
    year={2024},
    url={https://openreview.net/forum?id=v8L0pN6EOi}
}

@inproceedings{lewis2020retrieval,
  title={Retrieval-Augmented Generation for Knowledge-Intensive {NLP} Tasks},
  author={Lewis, Patrick and Perez, Ethan and Piktus, Aleksandra and Petroni, Fabio and Karpukhin, Vladimir and Goyal, Naman and K{\"u}ttler, Heinrich and Lewis, Mike and Yih, Wen-tau and Rockt{\"a}schel, Tim and Riedel, Sebastian and Kiela, Douwe},
  booktitle={Advances in Neural Information Processing Systems},
  volume={33},
  pages={9459--9474},
  year={2020}
}

@inproceedings{
    velickovic2018graph,
    title={Graph Attention Networks},
    author={Petar Veličković and Guillem Cucurull and Arantxa Casanova and Adriana Romero and Pietro Liò and Yoshua Bengio},
    booktitle={International Conference on Learning Representations},
    year={2018},
    url={https://openreview.net/forum?id=rJXMpikCZ},
}

@misc{adult_2,
  author       = {Becker, Barry and Kohavi, Ronny},
  title        = {Adult},
  year         = {1996},
  howpublished = {UCI Machine Learning Repository},
  doi          = {10.24432/C5XW20}
}

@misc{thoracic_surgery_data_277,
  author       = {Lubicz, Marek and Pawelczyk, Konrad
                  and Rzechonek, Adam and Kolodziej, Jerzy},
  title        = {Thoracic Surgery Data},
  year         = {2014},
  howpublished = {UCI Machine Learning Repository},
  doi          = {10.24432/C5Z60N}
}

@inproceedings{goldsmith2007measuring,
  title={Measuring empirical computational complexity},
  author={Goldsmith, Simon F and Aiken, Alex S and Wilkerson, Daniel S},
  booktitle={Proceedings of the 6th Joint Meeting of the European Software Engineering Conference and the ACM SIGSOFT Symposium on The Foundations of Software Engineering},
  pages={395--404},
  year={2007}
}
\clearpage

\appendix
\resetlinenumber[1]

\clearpage

\begin{center}
{\LARGE\bfseries Supplement}
\end{center}

\vspace{1em}

\begingroup
\setcounter{tocdepth}{2}
\renewcommand{\contentsname}{Supplement Content}
\tableofcontents
\endgroup

\clearpage
\setcounter{figure}{0}
\renewcommand{\figurename}{Supplementary Fig.}
\renewcommand{\thefigure}{\arabic{figure}}
\renewcommand{\theHfigure}{supp.\arabic{figure}}

\setcounter{table}{0}
\renewcommand{\tablename}{Supplementary Table}
\renewcommand{\thetable}{\arabic{table}}
\renewcommand{\theHtable}{supp.\arabic{table}}

\section{Details of experimental setups}

\subsection{Statistics of the evaluated datasets}
\label{appendix:datasets}

We evaluate our method on 13 real-world tabular datasets spanning classification and regression tasks, small-sample settings, and high-dimensional feature spaces. These datasets contain between 4 and 450 features and between 174 and 53,940 samples. Each dataset is divided into training and test sets for downstream machine-learning evaluation. For the Adult dataset, we adopt the official train--test split. For the Thoracic dataset, we use a fixed split of 270 training samples and 200 test samples for the augmentation experiments. The remaining datasets are partitioned using either a 9:1 or 7:3 train--test ratio with a fixed random seed. To evaluate the applicability of our method to high-dimensional data, we additionally apply the greedy set-cover procedure in Algorithm~\ref{alg:framework} to the two highest-dimensional datasets, \textit{DNA} and \textit{DARWIN}, selecting a compact subset of trees that maximizes variable coverage while controlling prompt length.
This procedure retains 19 and 54 trees in total, respectively, including the label tree. Dataset statistics are summarized in Supplementary Table~\ref{tab:dataset_stat}.


\begin{table}[htbp]
\centering
\caption{Statistics of the evaluated datasets. Dataset names are hyperlinked to their public sources. Num and Cat denote the numbers of numerical and categorical predictor features, respectively. Imb. denotes the majority-to-minority class ratio in the training set.}
\label{tab:dataset_stat}
\scriptsize
\begin{tabular*}{\linewidth}{@{\extracolsep{\fill}}lllrrrrrll@{}}
\toprule
Dataset & Domain & Source & Samples & Num & Cat & Train & Test & Task & Imb. \\
\midrule
\href{https://archive.ics.uci.edu/dataset/2/adult}{Adult}
& Social Sci. & UCI & 48842 & 6 & 8 & 32561 & 16281 & Classification & 3.153 \\

\href{https://archive.ics.uci.edu/dataset/350/default+of+credit+card+clients}{Default}
& Finance & UCI & 30000 & 14 & 9 & 27000 & 3000 & Classification & 3.523 \\

\href{https://archive.ics.uci.edu/dataset/159/magic+gamma+telescope}{Magic}
& Physics & UCI & 19019 & 10 & 0 & 17117 & 1902 & Classification & 1.847 \\

\href{https://www.openml.org/d/42225}{Diamonds}
& Commerce & OpenML & 53940 & 8 & 3 & 48546 & 5394 & Regression & -- \\

\href{https://archive.ics.uci.edu/dataset/176/blood+transfusion+service+center}{Blood}
& Medicine & UCI & 748 & 4 & 0 & 523 & 225 & Classification & 3.023 \\

\href{https://archive.ics.uci.edu/dataset/19/car+evaluation}{Car}
& Decision & UCI & 1728 & 0 & 6 & 1209 & 519 & Classification & 19.409 \\

\href{https://archive.ics.uci.edu/dataset/145/statlog+heart}{Heart}
& Medicine & UCI & 270 & 5 & 8 & 189 & 81 & Classification & 1.305 \\

\href{https://www.openml.org/d/8}{Liver}
& Medicine & OpenML & 345 & 5 & 0 & 241 & 104 & Regression & -- \\

\href{https://www.openml.org/d/40536}{SpeedDating}
& Social Sci. & OpenML & 8378 & 53 & 61 & 7540 & 838 & Classification & 5.081 \\

\href{https://www.openml.org/d/41143}{Jasmine}
& Life Sci. & OpenML & 2984 & 8 & 136 & 2685 & 299 & Classification & 1.008 \\

\href{https://www.openml.org/d/46}{DNA}
& Biology & OpenML & 3186 & 0 & 180 & 2230 & 956 & Classification & 2.222 \\

\href{https://archive.ics.uci.edu/dataset/732/darwin}{DARWIN}
& Medicine & UCI & 174 & 450 & 0 & 121 & 53 & Classification & 1.051 \\

\href{https://archive.ics.uci.edu/dataset/277/thoracic%2Bsurgery%2Bdata}{Thoracic}
& Medical & UCI & 470 & 3 & 13 & 270 & 200 & Classification & 5.585 \\
\bottomrule
\end{tabular*}
\end{table}

\subsection{Implementation details}
\label{appendix:implementation_details}
For each dataset, we fit all synthesis methods exclusively on the real training split and use each method to generate a synthetic table with the same number of rows as the training split. All methods are evaluated under the same train--test partition and downstream evaluation protocol. Downstream utility is evaluated using the train-on-synthetic, test-on-real protocol with AdaBoost~\cite{freund1997decision}, Random Forest~\cite{breiman2001random}, and XGBoost~\cite{chen2016xgboost} as downstream predictors; the reported utility score is the average normalized performance across these models. The reported results are averaged over 20 randomly sampled synthetic data following TABSYN~\cite{tabsyn}. No information from the held-out test set is used for synthetic-data synthesis, candidate-strategy selection, or hyperparameter selection.

In the main experiments, we use Gemini-2.5-Pro~\cite{comanici2025gemini} as the LLM backbone. For each dataset, the LLM is queried ten times to generate candidate Python-based synthesis strategies. Sensitivity to the number of LLM queries is analysed in Supplementary Fig.~\ref{fig:query_n}. Each candidate strategy is required to follow a unified executable interface, taking the local training table and the target sample size as inputs and saving the generated synthetic table to disk. Candidates that fail to execute, violate the required interface, or produce invalid output tables are discarded before strategy selection. The final strategy is selected using statistics computed only from the real training split, including statistical fidelity, empirical privacy-related risk, and computational efficiency. The held-out test set is used only for final evaluation.

Our framework first extracts dependence information from the original training data using chained trees and then provides only the encoded dependence summaries to the LLM. The original records are used locally for dependence extraction, candidate-strategy execution, and evaluation, but are not transmitted to the LLM during strategy generation. This design separates strategy-level reasoning from record-level data exposure while allowing the generated executable strategy to operate locally on the training data.

The baseline methods include (1) Conditional Tabular Generative Adversarial Network (CTGAN)~\cite{xu2019modeling}, which uses conditional generation and mode-specific normalization to model multimodal variables, (2) TVAE~\cite{xu2019modeling}, which learns a probabilistic latent representation of mixed-type tabular data using a variational autoencoder, (3) Generation of Realistic Tabular Data (GReaT)~\cite{borisovlanguage}, which serializes table rows as text and generates synthetic records autoregressively using a fine-tuned LLM, (4) diffusion-based methods, which generate mixed-type tabular data by progressively corrupting real samples and learning to reverse the resulting diffusion process, including Tabular Denoising Diffusion Probabilistic Model (TabDDPM)~\cite{kotelnikov2023tabddpm}, TabDiff~\cite{shi2025tabdiff}, and TABSYN~\cite{tabsyn}, (5) Adversarial Random Forests (ARF)~\cite{watson2023adversarial}, a tree-based model that iteratively trains random forests to distinguish real from synthetic records and samples from the resulting leaf-wise distributions, and (6) NRGBoost~\cite{bravo2025nrgboost}, an energy-based generative model that learns a generative energy function through boosted regression trees and samples from the resulting distribution. In addition to the tabular generative models, we include the Synthetic Minority Oversampling Technique (SMOTE)~\cite{chawla2002smote} as a baseline due to its strong performance on synthetic data fidelity. However, SMOTE relies on interpolation between neighbouring records and duplicates specific local geometries. Real records may be reconstructed from synthetic samples \cite{ganev2026smote}, incurring severe privacy vulnerabilities that hinder SMOTE's applicability to secure data-sharing contexts. By benchmarking against SMOTE, we demonstrate that TabSSD achieves competitive synthetic data fidelity and predictive utility but provides a more viable solution for privacy-preserving data synthesis.
ARF, TabDiff, and NRGBoost are implemented using their official codebases with the recommended experimental configurations. TABSYN, TabDDPM, and GReaT are implemented using the code and configurations released by the TABSYN authors, whereas CTGAN and TVAE are implemented using the SDV library with its default configurations~\cite{SDV}.
Unless otherwise specified, we follow the default or recommended configurations of the corresponding implementations and use the same train--test splits and evaluation protocol for all methods. All experiments are conducted on a workstation equipped with two NVIDIA RTX 4090 GPUs and two Intel Xeon Gold 6430 CPUs. While TABSYN, TabDDPM, GReaT, TabDiff, CTGAN, and TVAE relied on GPU acceleration, our method, TabSSD, required no GPU, demonstrating substantially lower hardware requirements and greater accessibility.

\subsection{Evaluation metrics}
\label{appendix:Metrics}

\subsubsection{Distance to closest record (DCR)}

Following the DCR evaluation protocol adopted by
TABSYN~\cite{tabsyn}, we assess whether synthetic records exhibit
excessive proximity to real training records. For each synthetic record, we
compare its nearest-neighbour distances to the training and test sets and
compute the proportion for which the nearest real neighbour belongs to the
training set:
\begin{equation}
\mathrm{DCR}
=
\frac{1}{|\hat{\mathcal{D}}|}
\sum_{i=1}^{|\hat{\mathcal{D}}|}
\mathbb{I}
\left[
d_{\mathrm{train}}(\hat{\mathbf{x}}_i)
<
d_{\mathrm{test}}(\hat{\mathbf{x}}_i)
\right].
\end{equation}

Because this proportion is affected by unequal train--test split sizes, we
centre it relative to the split-induced null expectation. Under the null
hypothesis of no preferential proximity to the training data, exchangeability
implies that the nearest real neighbour belongs to the training set with
probability
\(
|\mathcal{D}_{\mathrm{train}}|/
\bigl(|\mathcal{D}_{\mathrm{train}}|
+|\mathcal{D}_{\mathrm{test}}|\bigr)
\).
We therefore define
\begin{equation}
\mathrm{DCR}_{\mathrm{adjusted}}
=
\max\left(
\mathrm{DCR}
-
\frac{|\mathcal{D}_{\mathrm{train}}|}
{|\mathcal{D}_{\mathrm{train}}|
+|\mathcal{D}_{\mathrm{test}}|},
\,0
\right).
\end{equation}
The positive-part operation ensures that only proximity exceeding the
split-induced expectation contributes to the risk score; values at or below
this expectation are assigned a score of zero. Throughout this paper,
adjusted DCR refers to this split-ratio-adjusted score. Lower values indicate
less excessive proximity to training records and therefore lower potential
memorization risk.
\subsubsection{\texorpdfstring{$\delta$-presence}{delta-presence}}
We further use $\delta$-presence to evaluate the re-identification risk of the
generated synthetic data. Unlike DCR, which measures whether synthetic
records are excessively close to real training records, $\delta$-presence
characterizes the worst-case imbalance between real and synthetic records
across cluster-based partitions. In our experiments, $\delta$-presence is
computed following the implementation in SynthCity version
0.2.12~\cite{qian2023synthcity}. Specifically, $k$-means clustering models are
fitted to the real training data using the full encoded feature space, and
synthetic records are then assigned to the learned clusters using the same
clustering models.

Let $c_K(\cdot)$ denote the clustering function learned from
$\mathcal{D}_{\mathrm{train}}$ using $K$ clusters. For the $q$-th cluster, the
numbers of real and synthetic records assigned to the same partition are
defined as
\begin{equation}
n^{\mathrm{real}}_{K,q}
=
\sum_{\mathbf{x}_i \in \mathcal{D}_{\mathrm{train}}}
\mathbb{I}
\left[
c_K(\mathbf{x}_i)=q
\right],
\end{equation}
and
\begin{equation}
n^{\mathrm{syn}}_{K,q}
=
\sum_{\hat{\mathbf{x}}_i \in \hat{\mathcal{D}}}
\mathbb{I}
\left[
c_K(\hat{\mathbf{x}}_i)=q
\right].
\end{equation}

For each cluster represented in both the real and synthetic datasets, the
cluster-wise $\delta$-presence score is computed as
\begin{equation}
\delta_{K,q}
=
\frac{
n^{\mathrm{real}}_{K,q}
}{
n^{\mathrm{syn}}_{K,q}+\epsilon
},
\end{equation}
where $\epsilon=10^{-8}$ is a small constant used for numerical stability.
Following the SynthCity implementation, we evaluate multiple partition
granularities with
$K \in \{2,5,10,15\}$ and skip a clustering setting when the average number
of real records per cluster is smaller than 10, i.e.,
$|\mathcal{D}_{\mathrm{train}}|/K < 10$.

The final $\delta$-presence score is defined as the maximum cluster-wise ratio
over all retained clustering granularities and clusters represented in both
datasets:
\begin{equation}
\delta\text{-presence}
=
\max_{\substack{K,q:\\ n^{\mathrm{syn}}_{K,q}>0}}
\delta_{K,q}.
\end{equation}

A larger $\delta$-presence value indicates a larger worst-case
real-to-synthetic cluster-count ratio, meaning that at least one partition
contains relatively many real records compared with synthetic records.
Accordingly, in our privacy evaluation, a lower $\delta$-presence value is
regarded as preferable.

\subsubsection{Downstream utility}
For downstream utility, we adopt the train-on-synthetic, test-on-real protocol. 
Predictive models are trained on synthetic data and evaluated on the real test set, 
and their results are compared with those of the same models trained on real data. 
We use AdaBoost~\cite{freund1997decision}, Random Forest~\cite{breiman2001random}, 
and XGBoost~\cite{chen2016xgboost} as downstream predictors and average their results. 
Specifically, the utility score is computed as
\[
\mathrm{Utility} =
\begin{cases}
\dfrac{\mathrm{AUROC}_{\mathrm{syn}}}{\mathrm{AUROC}_{\mathrm{real}}}, 
& \text{for classification tasks}, \\[1.2em]
\dfrac{\mathrm{RMSE}_{\mathrm{real}}}{\mathrm{RMSE}_{\mathrm{syn}}}, 
& \text{for regression tasks},
\end{cases}
\]
where \(\mathrm{AUROC}_{\mathrm{syn}}\) and \(\mathrm{RMSE}_{\mathrm{syn}}\) denote 
the averaged results obtained by training on synthetic data, while 
\(\mathrm{AUROC}_{\mathrm{real}}\) and \(\mathrm{RMSE}_{\mathrm{real}}\) denote 
the corresponding averaged results obtained by training on real data. 
Because higher AUROC and lower RMSE indicate better predictive performance, 
a utility value closer to \(1\) indicates that training on synthetic data achieves 
performance closer to training on real data.




\subsubsection{Marginal distribution error and pairwise correlation error}
Following TABSYN~\cite{tabsyn}, we evaluate the statistical fidelity of the generated synthetic data using marginal distribution error and pairwise correlation error.

Marginal distribution error assesses whether the marginal distribution of each individual variable is preserved in the synthetic data. For numerical variables, distributional similarity is evaluated using the Kolmogorov--Smirnov statistic, whereas for categorical variables, total variation distance is used to compare category-frequency distributions.

Pairwise correlation error assesses whether relationships between variable pairs are preserved. For numerical variable pairs, similarity is evaluated by comparing the Pearson correlation coefficients computed from the real and synthetic data. For categorical variable pairs, contingency similarity is used to compare their joint-frequency distributions. For mixed-type variable pairs, the numerical variable is first discretized into bins, after which contingency similarity is computed.

In our experiments, these metrics are computed using the single-table Quality Report in SDMetrics version 0.9.3~\cite{sdmetrics}. The marginal distribution error is defined as one minus the Column Shapes score, and the pairwise correlation error is defined as one minus the Column Pair Trends score. The Column Shapes score averages KSComplement scores for numerical variables and TVComplement scores for categorical variables, whereas the Column Pair Trends score averages CorrelationSimilarity scores for numerical--numerical pairs and ContingencySimilarity scores for categorical--categorical and mixed-type pairs.
\subsubsection{\texorpdfstring{$\alpha$-precision}{alpha-precision}}

We use \(\alpha\)-precision to evaluate the sample-level fidelity of the generated synthetic data. Unlike the marginal distribution estimation and pairwise correlation estimation metrics, which focus on marginal distributions and inter-feature relationships, \(\alpha\)-precision evaluates whether synthetic samples lie within real-data regions defined by quantile radii around the real-data centre. Following the implementation in SynthCity version 0.2.12~\cite{qian2023synthcity}, we compute the naive \(\alpha\)-precision score, where real and synthetic data are first normalized using a scaler fitted on the real data, and the centre of the normalized real data is used as the reference point. The metric compares the empirical \(\alpha\)-precision curve with the ideal curve across different quantile radii. A higher \(\alpha\)-precision value indicates closer agreement between the empirical and ideal curves and therefore better sample-level fidelity to the real data distribution. 

\section{Detailed results}
\subsection{Statistical fidelity results}
\label{appendix:fidelity}
\FloatBarrier
Dataset-level statistical fidelity results are reported for marginal distribution error (Supplementary Table~\ref{tab:marginal_density_error}), pairwise correlation error (Supplementary Table~\ref{tab:pairwise_correlation_error}), and $\alpha$-precision (Supplementary Table~\ref{tab:alpha_precision}), capturing complementary aspects of marginal, inter-variable, and distributional fidelity.

\begin{sidewaystable}[p]
\centering
\caption{\textbf{Dataset-level marginal distribution errors (\%).} Results are reported as mean $\pm$ s.d. across 20 runs. Lower values are better; dashes indicate unavailable results. The average row reports the mean of the dataset-level mean errors.}
\label{tab:marginal_density_error}
\scriptsize
\setlength{\tabcolsep}{2pt}
\renewcommand{\arraystretch}{1.12}
\begin{tabular*}{\linewidth}{
@{\extracolsep{\fill}}lcccccccccc
}
\toprule
Dataset & SMOTE & ARF & NRGBoost & TVAE & CTGAN & TABSYN & TabDDPM & TabDiff & GReaT & TabSSD \\
\midrule
\textit{Adult} & \(1.57{\scriptstyle \pm 0.030}\) & \(4.99{\scriptstyle \pm 0.000}\) & \(1.80{\scriptstyle \pm 0.001}\) & \(8.05{\scriptstyle \pm 0.000}\) & \(8.87{\scriptstyle \pm 0.040}\) & \(2.17{\scriptstyle \pm 0.056}\) & \(1.18{\scriptstyle \pm 0.044}\) & \(0.64{\scriptstyle \pm 0.047}\) & \(12.74{\scriptstyle \pm 0.079}\) & \(5.44{\scriptstyle \pm 0.001}\) \\
\textit{Blood} & \(5.57{\scriptstyle \pm 0.929}\) & \(6.37{\scriptstyle \pm 0.006}\) & \(8.04{\scriptstyle \pm 0.008}\) & \(12.28{\scriptstyle \pm 0.000}\) & \(19.27{\scriptstyle \pm 0.251}\) & \(5.52{\scriptstyle \pm 0.554}\) & \(5.34{\scriptstyle \pm 0.747}\) & \(5.83{\scriptstyle \pm 0.701}\) & \(10.67{\scriptstyle \pm 1.246}\) & \(3.32{\scriptstyle \pm 0.005}\) \\
\textit{Car} & \(1.57{\scriptstyle \pm 0.267}\) & \(1.58{\scriptstyle \pm 0.003}\) & \(2.00{\scriptstyle \pm 0.005}\) & \(3.85{\scriptstyle \pm 0.000}\) & \(15.84{\scriptstyle \pm 0.224}\) & \(1.99{\scriptstyle \pm 0.402}\) & \(1.70{\scriptstyle \pm 0.379}\) & \(1.73{\scriptstyle \pm 0.312}\) & \(3.15{\scriptstyle \pm 0.428}\) & \(1.68{\scriptstyle \pm 0.004}\) \\
\textit{Default} & \(1.56{\scriptstyle \pm 0.089}\) & \(4.80{\scriptstyle \pm 0.000}\) & \(3.92{\scriptstyle \pm 0.001}\) & \(9.29{\scriptstyle \pm 0.000}\) & \(7.48{\scriptstyle \pm 0.025}\) & \(1.08{\scriptstyle \pm 0.057}\) & \(1.50{\scriptstyle \pm 0.089}\) & \(1.23{\scriptstyle \pm 0.101}\) & \(20.17{\scriptstyle \pm 0.124}\) & \(1.00{\scriptstyle \pm 0.001}\) \\
\textit{Diamonds} & \(1.57{\scriptstyle \pm 0.058}\) & \(1.82{\scriptstyle \pm 0.001}\) & \(5.61{\scriptstyle \pm 0.001}\) & \(8.61{\scriptstyle \pm 0.000}\) & \(5.18{\scriptstyle \pm 0.033}\) & \(3.19{\scriptstyle \pm 0.097}\) & \(0.93{\scriptstyle \pm 0.086}\) & \(0.72{\scriptstyle \pm 0.096}\) & \(4.88{\scriptstyle \pm 0.082}\) & \(3.79{\scriptstyle \pm 0.000}\) \\
\textit{Heart} & \(5.40{\scriptstyle \pm 0.448}\) & \(5.78{\scriptstyle \pm 0.006}\) & \(5.87{\scriptstyle \pm 0.009}\) & \(11.90{\scriptstyle \pm 0.000}\) & \(20.23{\scriptstyle \pm 0.355}\) & \(6.65{\scriptstyle \pm 0.565}\) & \(5.03{\scriptstyle \pm 0.530}\) & \(5.70{\scriptstyle \pm 0.836}\) & \(11.42{\scriptstyle \pm 0.599}\) & \(6.08{\scriptstyle \pm 0.007}\) \\
\textit{Jasmine} & \(1.38{\scriptstyle \pm 0.065}\) & \(0.97{\scriptstyle \pm 0.001}\) & \(3.09{\scriptstyle \pm 0.002}\) & \(6.77{\scriptstyle \pm 0.000}\) & \(11.15{\scriptstyle \pm 0.027}\) & \(3.57{\scriptstyle \pm 0.052}\) & \(28.16{\scriptstyle \pm 0.270}\) & \(0.82{\scriptstyle \pm 0.079}\) & -- & \(0.48{\scriptstyle \pm 0.000}\) \\
\textit{Liver} & \(7.56{\scriptstyle \pm 0.764}\) & \(10.55{\scriptstyle \pm 0.009}\) & \(7.44{\scriptstyle \pm 0.013}\) & \(12.17{\scriptstyle \pm 0.000}\) & \(19.43{\scriptstyle \pm 0.293}\) & \(9.14{\scriptstyle \pm 0.715}\) & \(7.60{\scriptstyle \pm 0.727}\) & \(8.61{\scriptstyle \pm 0.655}\) & \(13.60{\scriptstyle \pm 0.910}\) & \(9.20{\scriptstyle \pm 0.000}\) \\
\textit{Magic} & \(0.93{\scriptstyle \pm 0.070}\) & \(2.67{\scriptstyle \pm 0.001}\) & \(2.37{\scriptstyle \pm 0.002}\) & \(8.71{\scriptstyle \pm 0.000}\) & \(8.14{\scriptstyle \pm 0.055}\) & \(0.90{\scriptstyle \pm 0.102}\) & \(0.99{\scriptstyle \pm 0.109}\) & \(0.83{\scriptstyle \pm 0.095}\) & \(15.07{\scriptstyle \pm 0.097}\) & \(3.09{\scriptstyle \pm 0.000}\) \\
\textit{SpeedDating} & \(1.82{\scriptstyle \pm 0.075}\) & \(6.09{\scriptstyle \pm 0.001}\) & \(12.59{\scriptstyle \pm 0.000}\) & \(12.84{\scriptstyle \pm 0.000}\) & \(10.04{\scriptstyle \pm 0.036}\) & \(12.73{\scriptstyle \pm 0.047}\) & \(53.09{\scriptstyle \pm 0.118}\) & \(5.86{\scriptstyle \pm 0.101}\) & -- & \(1.11{\scriptstyle \pm 0.000}\) \\
\textit{DNA} & \(1.41{\scriptstyle \pm 0.001}\) & \(0.73{\scriptstyle \pm 0.000}\) & \(0.86{\scriptstyle \pm 0.000}\) & \(13.15{\scriptstyle \pm 0.000}\) & \(11.58{\scriptstyle \pm 0.000}\) & \(8.09{\scriptstyle \pm 0.001}\) & \(0.87{\scriptstyle \pm 0.001}\) & \(2.12{\scriptstyle \pm 0.069}\) & -- & \(0.76{\scriptstyle \pm 0.000}\) \\
\textit{DARWIN} & \(11.43{\scriptstyle \pm 0.004}\) & \(14.75{\scriptstyle \pm 0.002}\) & \(14.60{\scriptstyle \pm 0.003}\) & \(21.06{\scriptstyle \pm 0.000}\) & \(46.19{\scriptstyle \pm 0.001}\) & \(19.42{\scriptstyle \pm 0.003}\) & \(52.53{\scriptstyle \pm 0.001}\) & \(10.85{\scriptstyle \pm 0.204}\) & -- & \(9.67{\scriptstyle \pm 0.002}\) \\
\midrule
\textbf{Average} & 3.48 & 5.09 & 5.68 & 10.72 & 15.28 & 6.20 & 13.24 & 3.74 & 11.46 & 3.80 \\
\bottomrule
\end{tabular*}
\end{sidewaystable}

\begin{sidewaystable}[p]
\centering
\caption{\textbf{Dataset-level pairwise correlation errors (\%).} Results are reported as mean $\pm$ s.d. across 20 runs. Lower values are better; dashes indicate unavailable results. The average row reports the mean of the dataset-level mean errors.}
\label{tab:pairwise_correlation_error}
\scriptsize
\setlength{\tabcolsep}{2pt}
\renewcommand{\arraystretch}{1.12}
\begin{tabular*}{\linewidth}{
@{\extracolsep{\fill}}lcccccccccc
}
\toprule
Dataset & SMOTE & ARF & NRGBoost & TVAE & CTGAN & TABSYN & TabDDPM & TabDiff & GReaT & TabSSD \\
\midrule
\textit{Adult} & \(2.78{\scriptstyle \pm 0.041}\) & \(7.79{\scriptstyle \pm 0.000}\) & \(2.92{\scriptstyle \pm 0.001}\) & \(10.01{\scriptstyle \pm 0.000}\) & \(13.88{\scriptstyle \pm 0.044}\) & \(3.83{\scriptstyle \pm 0.065}\) & \(2.35{\scriptstyle \pm 0.055}\) & \(1.47{\scriptstyle \pm 0.056}\) & \(18.00{\scriptstyle \pm 0.097}\) & \(3.53{\scriptstyle \pm 0.000}\) \\
\textit{Blood} & \(2.83{\scriptstyle \pm 0.546}\) & \(4.65{\scriptstyle \pm 0.006}\) & \(8.25{\scriptstyle \pm 0.006}\) & \(10.35{\scriptstyle \pm 0.000}\) & \(13.35{\scriptstyle \pm 0.272}\) & \(3.65{\scriptstyle \pm 0.368}\) & \(3.46{\scriptstyle \pm 0.541}\) & \(40.86{\scriptstyle \pm 0.322}\) & \(6.58{\scriptstyle \pm 0.624}\) & \(2.68{\scriptstyle \pm 0.004}\) \\
\textit{Car} & \(3.70{\scriptstyle \pm 0.313}\) & \(3.68{\scriptstyle \pm 0.003}\) & \(3.95{\scriptstyle \pm 0.005}\) & \(6.94{\scriptstyle \pm 0.000}\) & \(26.25{\scriptstyle \pm 0.242}\) & \(4.12{\scriptstyle \pm 0.401}\) & \(3.75{\scriptstyle \pm 0.427}\) & \(4.15{\scriptstyle \pm 0.315}\) & \(6.10{\scriptstyle \pm 0.503}\) & \(3.99{\scriptstyle \pm 0.004}\) \\
\textit{Default} & \(1.68{\scriptstyle \pm 0.079}\) & \(4.58{\scriptstyle \pm 0.001}\) & \(5.00{\scriptstyle \pm 0.001}\) & \(9.63{\scriptstyle \pm 0.000}\) & \(5.34{\scriptstyle \pm 0.031}\) & \(1.17{\scriptstyle \pm 0.046}\) & \(5.92{\scriptstyle \pm 0.184}\) & \(1.38{\scriptstyle \pm 0.101}\) & \(14.40{\scriptstyle \pm 0.244}\) & \(5.09{\scriptstyle \pm 0.001}\) \\
\textit{Diamonds} & \(0.77{\scriptstyle \pm 0.050}\) & \(1.09{\scriptstyle \pm 0.000}\) & \(5.09{\scriptstyle \pm 0.001}\) & \(9.25{\scriptstyle \pm 0.000}\) & \(7.70{\scriptstyle \pm 0.030}\) & \(2.89{\scriptstyle \pm 0.063}\) & \(2.93{\scriptstyle \pm 0.174}\) & \(1.04{\scriptstyle \pm 0.053}\) & \(6.15{\scriptstyle \pm 0.067}\) & \(2.11{\scriptstyle \pm 0.000}\) \\
\textit{Heart} & \(10.53{\scriptstyle \pm 0.660}\) & \(12.21{\scriptstyle \pm 0.006}\) & \(10.81{\scriptstyle \pm 0.010}\) & \(15.87{\scriptstyle \pm 0.000}\) & \(26.55{\scriptstyle \pm 0.375}\) & \(12.65{\scriptstyle \pm 0.710}\) & \(8.91{\scriptstyle \pm 0.638}\) & \(11.88{\scriptstyle \pm 0.845}\) & \(19.20{\scriptstyle \pm 0.626}\) & \(11.11{\scriptstyle \pm 0.006}\) \\
\textit{Jasmine} & \(2.05{\scriptstyle \pm 0.102}\) & \(1.99{\scriptstyle \pm 0.001}\) & \(5.06{\scriptstyle \pm 0.003}\) & \(11.44{\scriptstyle \pm 0.000}\) & \(16.89{\scriptstyle \pm 0.041}\) & \(5.83{\scriptstyle \pm 0.089}\) & \(44.64{\scriptstyle \pm 0.406}\) & \(1.49{\scriptstyle \pm 0.123}\) & -- & \(0.96{\scriptstyle \pm 0.000}\) \\
\textit{Liver} & \(3.25{\scriptstyle \pm 0.809}\) & \(3.98{\scriptstyle \pm 0.008}\) & \(4.21{\scriptstyle \pm 0.013}\) & \(5.77{\scriptstyle \pm 0.000}\) & \(8.27{\scriptstyle \pm 0.579}\) & \(3.57{\scriptstyle \pm 0.774}\) & \(2.76{\scriptstyle \pm 0.701}\) & \(2.98{\scriptstyle \pm 0.631}\) & \(7.54{\scriptstyle \pm 0.940}\) & \(3.49{\scriptstyle \pm 0.000}\) \\
\textit{Magic} & \(0.90{\scriptstyle \pm 0.095}\) & \(1.36{\scriptstyle \pm 0.001}\) & \(1.10{\scriptstyle \pm 0.001}\) & \(4.45{\scriptstyle \pm 0.000}\) & \(6.82{\scriptstyle \pm 0.059}\) & \(0.61{\scriptstyle \pm 0.088}\) & \(0.75{\scriptstyle \pm 0.082}\) & \(0.63{\scriptstyle \pm 0.085}\) & \(7.20{\scriptstyle \pm 0.116}\) & \(0.98{\scriptstyle \pm 0.000}\) \\
\textit{SpeedDating} & \(2.92{\scriptstyle \pm 0.084}\) & \(9.02{\scriptstyle \pm 0.000}\) & \(10.42{\scriptstyle \pm 0.000}\) & \(15.45{\scriptstyle \pm 0.000}\) & \(11.02{\scriptstyle \pm 0.030}\) & \(19.08{\scriptstyle \pm 0.057}\) & \(64.17{\scriptstyle \pm 0.053}\) & \(8.22{\scriptstyle \pm 0.119}\) & -- & \(5.07{\scriptstyle \pm 0.000}\) \\
\textit{DNA} & \(2.37{\scriptstyle \pm 0.001}\) & \(1.61{\scriptstyle \pm 0.000}\) & \(1.59{\scriptstyle \pm 0.001}\) & \(25.67{\scriptstyle \pm 0.000}\) & \(18.50{\scriptstyle \pm 0.000}\) & \(12.60{\scriptstyle \pm 0.001}\) & \(13.77{\scriptstyle \pm 0.001}\) & \(3.43{\scriptstyle \pm 0.099}\) & -- & \(1.75{\scriptstyle \pm 0.000}\) \\
\textit{DARWIN} & \(4.16{\scriptstyle \pm 0.003}\) & \(6.20{\scriptstyle \pm 0.001}\) & \(7.13{\scriptstyle \pm 0.000}\) & \(5.91{\scriptstyle \pm 0.000}\) & \(16.09{\scriptstyle \pm 0.001}\) & \(6.57{\scriptstyle \pm 0.001}\) & \(7.84{\scriptstyle \pm 0.000}\) & \(5.70{\scriptstyle \pm 0.115}\) & -- & \(5.91{\scriptstyle \pm 0.001}\) \\
\midrule
\textbf{Average} & 3.16 & 4.85 & 5.46 & 10.89 & 14.22 & 6.38 & 13.44 & 6.94 & 10.65 & 3.89 \\
\bottomrule
\end{tabular*}
\end{sidewaystable}

\begin{sidewaystable}[p]
\centering

\caption{\textbf{Dataset-level $\alpha$-precision scores.}
Results are reported as mean $\pm$ s.d. across 20 runs.
Higher values indicate better statistical fidelity; dashes indicate
unavailable results. The average row reports the mean of the
dataset-level mean scores.}

\label{tab:alpha_precision}

\scriptsize
\setlength{\tabcolsep}{2pt}
\renewcommand{\arraystretch}{1.12}

\begin{tabular*}{\linewidth}{
@{\extracolsep{\fill}}lcccccccccc
}
\toprule
Dataset & SMOTE & ARF & NRGBoost & TVAE & CTGAN & TABSYN & TabDDPM & TabDiff & GReaT & TabSSD \\
\midrule
\textit{Adult} & \(0.9307{\scriptstyle \pm 0.002}\) & \(0.9917{\scriptstyle \pm 0.002}\) & \(0.9514{\scriptstyle \pm 0.003}\) & \(0.9777{\scriptstyle \pm 0.000}\) & \(0.7666{\scriptstyle \pm 0.002}\) & \(0.9791{\scriptstyle \pm 0.002}\) & \(0.9529{\scriptstyle \pm 0.003}\) & \(0.9914{\scriptstyle \pm 0.002}\) & \(0.5443{\scriptstyle \pm 0.001}\) & \(0.9944{\scriptstyle \pm 0.002}\) \\
\textit{Blood} & \(0.9706{\scriptstyle \pm 0.009}\) & \(0.9683{\scriptstyle \pm 0.010}\) & \(0.9105{\scriptstyle \pm 0.020}\) & \(0.8857{\scriptstyle \pm 0.000}\) & \(0.7663{\scriptstyle \pm 0.005}\) & \(0.9638{\scriptstyle \pm 0.010}\) & \(0.9642{\scriptstyle \pm 0.012}\) & \(0.1718{\scriptstyle \pm 0.008}\) & \(0.9413{\scriptstyle \pm 0.015}\) & \(0.9627{\scriptstyle \pm 0.011}\) \\
\textit{Car} & \(0.9860{\scriptstyle \pm 0.006}\) & \(0.9808{\scriptstyle \pm 0.006}\) & \(0.9688{\scriptstyle \pm 0.019}\) & \(0.9751{\scriptstyle \pm 0.000}\) & \(0.9319{\scriptstyle \pm 0.005}\) & \(0.9825{\scriptstyle \pm 0.008}\) & \(0.9826{\scriptstyle \pm 0.006}\) & \(0.9825{\scriptstyle \pm 0.007}\) & \(0.9603{\scriptstyle \pm 0.014}\) & \(0.9848{\scriptstyle \pm 0.005}\) \\
\textit{Default} & \(0.9790{\scriptstyle \pm 0.003}\) & \(0.9427{\scriptstyle \pm 0.002}\) & \(0.9406{\scriptstyle \pm 0.003}\) & \(0.8576{\scriptstyle \pm 0.000}\) & \(0.9179{\scriptstyle \pm 0.001}\) & \(0.9885{\scriptstyle \pm 0.002}\) & \(0.9893{\scriptstyle \pm 0.002}\) & \(0.9872{\scriptstyle \pm 0.003}\) & \(0.8611{\scriptstyle \pm 0.002}\) & \(0.9790{\scriptstyle \pm 0.001}\) \\
\textit{Diamonds} & \(0.9954{\scriptstyle \pm 0.002}\) & \(0.9962{\scriptstyle \pm 0.002}\) & \(0.9553{\scriptstyle \pm 0.001}\) & \(0.9442{\scriptstyle \pm 0.000}\) & \(0.8667{\scriptstyle \pm 0.002}\) & \(0.9743{\scriptstyle \pm 0.003}\) & \(0.9956{\scriptstyle \pm 0.002}\) & \(0.9964{\scriptstyle \pm 0.001}\) & \(0.8522{\scriptstyle \pm 0.003}\) & \(0.9957{\scriptstyle \pm 0.001}\) \\
\textit{Heart} & \(0.9042{\scriptstyle \pm 0.036}\) & \(0.9527{\scriptstyle \pm 0.016}\) & \(0.9462{\scriptstyle \pm 0.020}\) & \(0.7023{\scriptstyle \pm 0.000}\) & \(0.8516{\scriptstyle \pm 0.018}\) & \(0.8934{\scriptstyle \pm 0.033}\) & \(0.9340{\scriptstyle \pm 0.029}\) & \(0.9504{\scriptstyle \pm 0.021}\) & \(0.6085{\scriptstyle \pm 0.032}\) & \(0.9442{\scriptstyle \pm 0.021}\) \\
\textit{Jasmine} & \(0.8616{\scriptstyle \pm 0.009}\) & \(0.9403{\scriptstyle \pm 0.007}\) & \(0.7595{\scriptstyle \pm 0.012}\) & \(0.4716{\scriptstyle \pm 0.000}\) & \(0.7703{\scriptstyle \pm 0.003}\) & \(0.7499{\scriptstyle \pm 0.010}\) & \(0.2855{\scriptstyle \pm 0.012}\) & \(0.9792{\scriptstyle \pm 0.007}\) & -- & \(0.9391{\scriptstyle \pm 0.007}\) \\
\textit{Liver} & \(0.8878{\scriptstyle \pm 0.032}\) & \(0.8791{\scriptstyle \pm 0.036}\) & \(0.9071{\scriptstyle \pm 0.031}\) & \(0.7834{\scriptstyle \pm 0.000}\) & \(0.7473{\scriptstyle \pm 0.012}\) & \(0.9404{\scriptstyle \pm 0.019}\) & \(0.9646{\scriptstyle \pm 0.011}\) & \(0.9386{\scriptstyle \pm 0.017}\) & \(0.6761{\scriptstyle \pm 0.028}\) & \(0.9659{\scriptstyle \pm 0.000}\) \\
\textit{Magic} & \(0.9814{\scriptstyle \pm 0.003}\) & \(0.9865{\scriptstyle \pm 0.002}\) & \(0.9916{\scriptstyle \pm 0.002}\) & \(0.8365{\scriptstyle \pm 0.000}\) & \(0.8547{\scriptstyle \pm 0.001}\) & \(0.9918{\scriptstyle \pm 0.003}\) & \(0.9827{\scriptstyle \pm 0.004}\) & \(0.9941{\scriptstyle \pm 0.002}\) & \(0.8610{\scriptstyle \pm 0.005}\) & \(0.9925{\scriptstyle \pm 0.000}\) \\
\textit{SpeedDating} & \(0.9391{\scriptstyle \pm 0.006}\) & \(0.9323{\scriptstyle \pm 0.003}\) & \(0.8807{\scriptstyle \pm 0.004}\) & \(0.7526{\scriptstyle \pm 0.000}\) & \(0.9683{\scriptstyle \pm 0.003}\) & \(0.6902{\scriptstyle \pm 0.006}\) & \(0.0832{\scriptstyle \pm 0.001}\) & \(0.7229{\scriptstyle \pm 0.007}\) & -- & \(0.8948{\scriptstyle \pm 0.002}\) \\
\textit{DNA} & \(0.8145{\scriptstyle \pm 0.014}\) & \(0.9669{\scriptstyle \pm 0.009}\) & \(0.8654{\scriptstyle \pm 0.007}\) & \(0.0771{\scriptstyle \pm 0.000}\) & \(0.6164{\scriptstyle \pm 0.004}\) & \(0.3043{\scriptstyle \pm 0.007}\) & \(0.9491{\scriptstyle \pm 0.011}\) & \(0.6674{\scriptstyle \pm 0.012}\) & -- & \(0.9550{\scriptstyle \pm 0.009}\) \\
\textit{DARWIN} & \(0.5071{\scriptstyle \pm 0.050}\) & \(0.6772{\scriptstyle \pm 0.018}\) & \(0.1892{\scriptstyle \pm 0.004}\) & \(0.0000{\scriptstyle \pm 0.000}\) & \(0.0000{\scriptstyle \pm 0.000}\) & \(0.0002{\scriptstyle \pm 0.000}\) & \(0.0000{\scriptstyle \pm 0.000}\) & \(0.2002{\scriptstyle \pm 0.016}\) & -- & \(0.6623{\scriptstyle \pm 0.019}\) \\
\midrule
\textbf{Average} & 0.8964 & 0.9346 & 0.8555 & 0.6887 & 0.7548 & 0.7882 & 0.7570 & 0.7985 & 0.7881 & 0.9392 \\
\bottomrule
\end{tabular*}

\end{sidewaystable}

\subsection{Downstream predictive results}
\label{appendix:utility}
Dataset-level downstream predictive performance is reported in Supplementary Table~\ref{tab:downstream_utility}, comparing models trained on synthetic data with the corresponding real-data reference performance.
Classification datasets report mean AUROC, whereas the regression datasets (\textit{Diamonds} and \textit{Liver}) report mean RMSE. Results are averaged across AdaBoost, Random Forest, and XGBoost. 

\begin{sidewaystable}[p]
\centering

\caption{\textbf{Dataset-level downstream predictive performance.} Real denotes the reference performance obtained by training on real data. Classification datasets report mean AUROC, whereas the regression datasets (\textit{Diamonds} and \textit{Liver}) report mean RMSE. Results are averaged across AdaBoost, Random Forest, and XGBoost. Synthetic-data results are reported as mean $\pm$ s.d. across 20 runs. Higher AUROC and lower RMSE indicate better performance; dashes indicate unavailable results.}

\label{tab:downstream_utility}

\scriptsize
\setlength{\tabcolsep}{2pt}
\renewcommand{\arraystretch}{1.12}

\begin{tabular*}{\linewidth}{
@{\extracolsep{\fill}}lccccccccccc
}
\toprule
Dataset & Real & SMOTE & ARF & NRGBoost & TVAE & CTGAN & TABSYN
& TabDDPM & TabDiff & GReaT & TabSSD \\
\midrule

\textit{Adult}
& 0.9209
& 0.9003{\tiny$\!\pm\!0.001$}
& 0.9065{\tiny$\!\pm\!0.001$}
& 0.9112{\tiny$\!\pm\!0.001$}
& 0.8889{\tiny$\!\pm\!0.000$}
& 0.8767{\tiny$\!\pm\!0.004$}
& 0.9063{\tiny$\!\pm\!0.001$}
& 0.9100{\tiny$\!\pm\!0.001$}
& 0.9107{\tiny$\!\pm\!0.001$}
& 0.9062{\tiny$\!\pm\!0.002$}
& 0.9001{\tiny$\!\pm\!0.001$} \\

\textit{Blood}
& 0.7353
& 0.7026{\tiny$\!\pm\!0.023$}
& 0.7169{\tiny$\!\pm\!0.045$}
& 0.5786{\tiny$\!\pm\!0.089$}
& 0.7935{\tiny$\!\pm\!0.000$}
& 0.7959{\tiny$\!\pm\!0.017$}
& 0.7229{\tiny$\!\pm\!0.046$}
& 0.6860{\tiny$\!\pm\!0.059$}
& 0.7235{\tiny$\!\pm\!0.032$}
& 0.6127{\tiny$\!\pm\!0.072$}
& 0.7036{\tiny$\!\pm\!0.026$} \\

\textit{Car}
& 0.9789
& 0.9772{\tiny$\!\pm\!0.006$}
& 0.9084{\tiny$\!\pm\!0.010$}
& 0.9790{\tiny$\!\pm\!0.006$}
& 0.9305{\tiny$\!\pm\!0.000$}
& 0.7924{\tiny$\!\pm\!0.040$}
& 0.9637{\tiny$\!\pm\!0.004$}
& 0.9780{\tiny$\!\pm\!0.007$}
& 0.8994{\tiny$\!\pm\!0.016$}
& 0.9521{\tiny$\!\pm\!0.006$}
& 0.9577{\tiny$\!\pm\!0.004$} \\

\textit{Default}
& 0.7717
& 0.7492{\tiny$\!\pm\!0.004$}
& 0.7514{\tiny$\!\pm\!0.002$}
& 0.7566{\tiny$\!\pm\!0.002$}
& 0.7319{\tiny$\!\pm\!0.000$}
& 0.7278{\tiny$\!\pm\!0.003$}
& 0.7619{\tiny$\!\pm\!0.002$}
& 0.7607{\tiny$\!\pm\!0.002$}
& 0.7617{\tiny$\!\pm\!0.004$}
& 0.7494{\tiny$\!\pm\!0.004$}
& 0.7520{\tiny$\!\pm\!0.002$} \\

\textit{Diamonds}
& 836.93
& 860.13{\tiny$\!\pm\!10.52$}
& 1027.78{\tiny$\!\pm\!12.78$}
& 956.81{\tiny$\!\pm\!17.24$}
& 1544.25{\tiny$\!\pm\!0.00$}
& 1521.09{\tiny$\!\pm\!90.47$}
& 927.47{\tiny$\!\pm\!21.21$}
& 949.41{\tiny$\!\pm\!16.25$}
& 891.76{\tiny$\!\pm\!39.39$}
& 984.00{\tiny$\!\pm\!26.59$}
& 934.09{\tiny$\!\pm\!15.11$} \\

\textit{Heart}
& 0.8498
& 0.8417{\tiny$\!\pm\!0.024$}
& 0.8022{\tiny$\!\pm\!0.031$}
& 0.8030{\tiny$\!\pm\!0.031$}
& 0.8467{\tiny$\!\pm\!0.000$}
& 0.7910{\tiny$\!\pm\!0.021$}
& 0.8348{\tiny$\!\pm\!0.025$}
& 0.8416{\tiny$\!\pm\!0.020$}
& 0.8409{\tiny$\!\pm\!0.023$}
& 0.7698{\tiny$\!\pm\!0.038$}
& 0.8233{\tiny$\!\pm\!0.028$} \\

\textit{Jasmine}
& 0.8860
& 0.8704{\tiny$\!\pm\!0.010$}
& 0.8518{\tiny$\!\pm\!0.009$}
& 0.8668{\tiny$\!\pm\!0.008$}
& 0.8556{\tiny$\!\pm\!0.000$}
& 0.6231{\tiny$\!\pm\!0.047$}
& 0.8456{\tiny$\!\pm\!0.013$}
& 0.5185{\tiny$\!\pm\!0.077$}
& 0.8549{\tiny$\!\pm\!0.009$}
& --
& 0.8622{\tiny$\!\pm\!0.009$} \\

\textit{Liver}
& 3.4243
& 3.4096{\tiny$\!\pm\!0.090$}
& 3.5508{\tiny$\!\pm\!0.102$}
& 3.5743{\tiny$\!\pm\!0.184$}
& 3.4199{\tiny$\!\pm\!0.000$}
& 3.8220{\tiny$\!\pm\!0.099$}
& 3.3874{\tiny$\!\pm\!0.147$}
& 3.5228{\tiny$\!\pm\!0.153$}
& 3.5010{\tiny$\!\pm\!0.186$}
& 4.1169{\tiny$\!\pm\!0.209$}
& 3.1812{\tiny$\!\pm\!0.000$} \\

\textit{Magic}
& 0.9362
& 0.9320{\tiny$\!\pm\!0.001$}
& 0.9137{\tiny$\!\pm\!0.002$}
& 0.9252{\tiny$\!\pm\!0.002$}
& 0.8934{\tiny$\!\pm\!0.000$}
& 0.8832{\tiny$\!\pm\!0.002$}
& 0.9293{\tiny$\!\pm\!0.002$}
& 0.9256{\tiny$\!\pm\!0.002$}
& 0.9281{\tiny$\!\pm\!0.002$}
& 0.9012{\tiny$\!\pm\!0.003$}
& 0.9206{\tiny$\!\pm\!0.000$} \\

\textit{SpeedDating}
& 0.8515
& 0.8357{\tiny$\!\pm\!0.006$}
& 0.7995{\tiny$\!\pm\!0.012$}
& 0.5051{\tiny$\!\pm\!0.025$}
& 0.7640{\tiny$\!\pm\!0.000$}
& 0.7755{\tiny$\!\pm\!0.011$}
& 0.8137{\tiny$\!\pm\!0.007$}
& 0.4952{\tiny$\!\pm\!0.044$}
& 0.8251{\tiny$\!\pm\!0.006$}
& --
& 0.8010{\tiny$\!\pm\!0.005$} \\

\textit{DNA}
& 0.9848
& 0.9847{\tiny$\!\pm\!0.002$}
& 0.9437{\tiny$\!\pm\!0.008$}
& 0.9833{\tiny$\!\pm\!0.002$}
& 0.9655{\tiny$\!\pm\!0.000$}
& 0.4791{\tiny$\!\pm\!0.020$}
& 0.9292{\tiny$\!\pm\!0.010$}
& 0.4965{\tiny$\!\pm\!0.031$}
& 0.9758{\tiny$\!\pm\!0.003$}
& --
& 0.9623{\tiny$\!\pm\!0.006$} \\

\textit{DARWIN}
& 0.9667
& 0.9081{\tiny$\!\pm\!0.035$}
& 0.8007{\tiny$\!\pm\!0.055$}
& 0.6241{\tiny$\!\pm\!0.111$}
& 0.8758{\tiny$\!\pm\!0.000$}
& 0.4977{\tiny$\!\pm\!0.081$}
& 0.8470{\tiny$\!\pm\!0.052$}
& 0.4580{\tiny$\!\pm\!0.071$}
& 0.8633{\tiny$\!\pm\!0.026$}
& --
& 0.8805{\tiny$\!\pm\!0.022$} \\

\bottomrule
\end{tabular*}

\end{sidewaystable}

\subsection{DCR results}
\label{appendix:dcr}
Dataset-level DCR results are reported in Supplementary Table~\ref{tab:raw_dcr}, together with the corresponding train-set proportions used as references for assessing empirical memorization risk.
\begin{sidewaystable}[p]
\centering
\caption{\textbf{Dataset-level DCR values.} DCR is the proportion of synthetic records closer to the training set than to the held-out test set. Results are reported as mean $\pm$ s.d. across 20 runs where available. Values closer to the train-set proportion indicate lower empirical memorization risk; dashes indicate unavailable results.}
\label{tab:raw_dcr}
\scriptsize
\setlength{\tabcolsep}{2pt}
\renewcommand{\arraystretch}{1.12}
\begin{tabular*}{\linewidth}{
@{\extracolsep{\fill}}lccccccccccc
}
\toprule
Dataset & Train prop. & SMOTE & ARF & NRGBoost & TVAE & CTGAN & TABSYN & TabDDPM & TabDiff & GReaT & TabSSD \\
\midrule
\textit{Adult} & 0.67 & \(0.9438{\scriptstyle \pm 0.001}\) & \(0.7231{\scriptstyle \pm 0.003}\) & \(0.7046{\scriptstyle \pm 0.003}\) & \(0.6685{\scriptstyle \pm 0.000}\) & \(0.6748{\scriptstyle \pm 0.002}\) & \(0.6687{\scriptstyle \pm 0.003}\) & \(0.6798{\scriptstyle \pm 0.002}\) & \(0.7057{\scriptstyle \pm 0.003}\) & \(0.6785{\scriptstyle \pm 0.004}\) & \(0.6715{\scriptstyle \pm 0.002}\) \\
\textit{Blood} & 0.70 & \(0.8732{\scriptstyle \pm 0.011}\) & \(0.7528{\scriptstyle \pm 0.021}\) & \(0.9444{\scriptstyle \pm 0.005}\) & \(0.7354{\scriptstyle \pm 0.000}\) & \(0.7258{\scriptstyle \pm 0.007}\) & \(0.7342{\scriptstyle \pm 0.022}\) & \(0.7676{\scriptstyle \pm 0.014}\) & \(0.7326{\scriptstyle \pm 0.016}\) & \(0.7569{\scriptstyle \pm 0.012}\) & \(0.7543{\scriptstyle \pm 0.017}\) \\
\textit{Car} & 0.70 & \(0.9972{\scriptstyle \pm 0.001}\) & \(0.8099{\scriptstyle \pm 0.012}\) & \(0.9973{\scriptstyle \pm 0.001}\) & \(0.7438{\scriptstyle \pm 0.000}\) & \(0.7419{\scriptstyle \pm 0.009}\) & \(0.7574{\scriptstyle \pm 0.012}\) & \(0.7644{\scriptstyle \pm 0.010}\) & \(0.7156{\scriptstyle \pm 0.015}\) & \(0.7186{\scriptstyle \pm 0.013}\) & \(0.7090{\scriptstyle \pm 0.009}\) \\
\textit{Default} & 0.90 & \(0.9856{\scriptstyle \pm 0.001}\) & \(0.9073{\scriptstyle \pm 0.002}\) & \(0.9022{\scriptstyle \pm 0.002}\) & \(0.8950{\scriptstyle \pm 0.000}\) & \(0.9009{\scriptstyle \pm 0.002}\) & \(0.9031{\scriptstyle \pm 0.002}\) & \(0.9039{\scriptstyle \pm 0.002}\) & \(0.9158{\scriptstyle \pm 0.002}\) & \(0.9004{\scriptstyle \pm 0.002}\) & \(0.9027{\scriptstyle \pm 0.002}\) \\
\textit{Diamonds} & 0.90 & \(0.9813{\scriptstyle \pm 0.001}\) & \(0.9176{\scriptstyle \pm 0.001}\) & \(0.9047{\scriptstyle \pm 0.002}\) & \(0.8984{\scriptstyle \pm 0.000}\) & \(0.9040{\scriptstyle \pm 0.001}\) & \(0.9013{\scriptstyle \pm 0.002}\) & \(0.9009{\scriptstyle \pm 0.001}\) & \(0.9119{\scriptstyle \pm 0.001}\) & \(0.9040{\scriptstyle \pm 0.001}\) & \(0.9020{\scriptstyle \pm 0.001}\) \\
\textit{Heart} & 0.70 & \(0.9958{\scriptstyle \pm 0.004}\) & \(0.7616{\scriptstyle \pm 0.031}\) & \(0.9995{\scriptstyle \pm 0.002}\) & \(0.8254{\scriptstyle \pm 0.000}\) & \(0.7561{\scriptstyle \pm 0.027}\) & \(0.9328{\scriptstyle \pm 0.017}\) & \(0.9873{\scriptstyle \pm 0.007}\) & \(0.8651{\scriptstyle \pm 0.020}\) & \(0.7543{\scriptstyle \pm 0.021}\) & \(0.7616{\scriptstyle \pm 0.035}\) \\
\textit{Jasmine} & 0.90 & \(1.0000{\scriptstyle \pm 0.000}\) & \(0.9291{\scriptstyle \pm 0.004}\) & \(0.9200{\scriptstyle \pm 0.005}\) & \(0.9363{\scriptstyle \pm 0.000}\) & \(0.9142{\scriptstyle \pm 0.003}\) & \(0.9125{\scriptstyle \pm 0.006}\) & \(0.9421{\scriptstyle \pm 0.004}\) & \(0.9129{\scriptstyle \pm 0.005}\) & -- & \(0.9199{\scriptstyle \pm 0.006}\) \\
\textit{Liver} & 0.70 & \(0.9749{\scriptstyle \pm 0.011}\) & \(0.7562{\scriptstyle \pm 0.025}\) & \(1.0000{\scriptstyle \pm 0.000}\) & \(0.7427{\scriptstyle \pm 0.000}\) & \(0.7114{\scriptstyle \pm 0.016}\) & \(0.7618{\scriptstyle \pm 0.028}\) & \(1.0000{\scriptstyle \pm 0.000}\) & \(0.7205{\scriptstyle \pm 0.026}\) & \(0.7326{\scriptstyle \pm 0.036}\) & \(0.7178{\scriptstyle \pm 0.000}\) \\
\textit{Magic} & 0.90 & \(0.9945{\scriptstyle \pm 0.001}\) & \(0.9105{\scriptstyle \pm 0.002}\) & \(0.9116{\scriptstyle \pm 0.002}\) & \(0.8964{\scriptstyle \pm 0.000}\) & \(0.8999{\scriptstyle \pm 0.002}\) & \(0.9044{\scriptstyle \pm 0.002}\) & \(0.9052{\scriptstyle \pm 0.002}\) & \(0.9054{\scriptstyle \pm 0.002}\) & \(0.9051{\scriptstyle \pm 0.003}\) & \(0.8993{\scriptstyle \pm 0.000}\) \\
\textit{SpeedDating} & 0.90 & \(1.0000{\scriptstyle \pm 0.000}\) & \(0.9419{\scriptstyle \pm 0.003}\) & \(0.9056{\scriptstyle \pm 0.002}\) & \(0.9188{\scriptstyle \pm 0.000}\) & \(0.9111{\scriptstyle \pm 0.003}\) & \(0.9082{\scriptstyle \pm 0.004}\) & \(0.7022{\scriptstyle \pm 0.004}\) & \(0.9716{\scriptstyle \pm 0.002}\) & -- & \(0.9090{\scriptstyle \pm 0.003}\) \\
\textit{DNA} & 0.70 & \(0.9830{\scriptstyle \pm 0.001}\) & \(0.8537{\scriptstyle \pm 0.006}\) & \(0.7999{\scriptstyle \pm 0.010}\) & \(0.9350{\scriptstyle \pm 0.000}\) & \(0.7985{\scriptstyle \pm 0.006}\) & \(0.8773{\scriptstyle \pm 0.007}\) & \(0.7541{\scriptstyle \pm 0.007}\) & \(0.9533{\scriptstyle \pm 0.004}\) & -- & \(0.7687{\scriptstyle \pm 0.006}\) \\
\textit{DARWIN} & 0.70 & \(1.0000{\scriptstyle \pm 0.000}\) & \(0.6583{\scriptstyle \pm 0.044}\) & \(0.6215{\scriptstyle \pm 0.046}\) & \(0.5455{\scriptstyle \pm 0.000}\) & \(0.6810{\scriptstyle \pm 0.024}\) & \(0.9256{\scriptstyle \pm 0.024}\) & \(0.9146{\scriptstyle \pm 0.024}\) & \(0.5983{\scriptstyle \pm 0.045}\) & -- & \(0.7287{\scriptstyle \pm 0.050}\) \\
\bottomrule
\end{tabular*}
\end{sidewaystable}

\subsection{\texorpdfstring{$\delta$}{delta}-presence results}
\label{appendix:delta}
Dataset-level $\delta$-presence results are reported in Supplementary Table~\ref{tab:delta_presence}, providing an empirical assessment of re-identification risk across datasets and synthesis methods.
\begin{sidewaystable}[p]
\centering
\caption{\textbf{Dataset-level $\delta$-presence scores.} Results are reported as mean $\pm$ s.d. across 20 runs. Lower values indicate lower empirical re-identification risk; dashes indicate unavailable results. The average row reports the mean of the dataset-level mean scores.}
\label{tab:delta_presence}
\scriptsize
\setlength{\tabcolsep}{2pt}
\renewcommand{\arraystretch}{1.12}
\begin{tabular*}{\linewidth}{
@{\extracolsep{\fill}}lcccccccccc
}
\toprule
Dataset & SMOTE & ARF & NRGBoost & TVAE & CTGAN & TABSYN & TabDDPM & TabDiff & GReaT & TabSSD \\
\midrule
\textit{Adult} & \(5.5076{\scriptstyle \pm 5.042}\) & \(1.4935{\scriptstyle \pm 0.473}\) & \(1.8159{\scriptstyle \pm 0.309}\) & \(25.0000{\scriptstyle \pm 0.000}\) & \(468.2476{\scriptstyle \pm 153.135}\) & \(5.9231{\scriptstyle \pm 1.395}\) & \(1.5995{\scriptstyle \pm 0.277}\) & \(1.1333{\scriptstyle \pm 0.103}\) & \(2.8423{\scriptstyle \pm 0.341}\) & \(1.3100{\scriptstyle \pm 0.297}\) \\
\textit{Blood} & \(2.9354{\scriptstyle \pm 1.013}\) & \(2.7067{\scriptstyle \pm 0.884}\) & \(2.4535{\scriptstyle \pm 1.023}\) & \(7.0000{\scriptstyle \pm 0.000}\) & \(3.4894{\scriptstyle \pm 0.195}\) & \(10.0354{\scriptstyle \pm 5.799}\) & \(3.3739{\scriptstyle \pm 1.408}\) & \(8.0617{\scriptstyle \pm 4.713}\) & \(6.3392{\scriptstyle \pm 2.833}\) & \(2.5710{\scriptstyle \pm 0.968}\) \\
\textit{Car} & \(1.2454{\scriptstyle \pm 0.091}\) & \(1.2465{\scriptstyle \pm 0.111}\) & \(1.2899{\scriptstyle \pm 0.136}\) & \(1.6444{\scriptstyle \pm 0.000}\) & \(11.1778{\scriptstyle \pm 6.092}\) & \(1.2876{\scriptstyle \pm 0.080}\) & \(1.2679{\scriptstyle \pm 0.090}\) & \(1.2646{\scriptstyle \pm 0.102}\) & \(1.4477{\scriptstyle \pm 0.159}\) & \(1.2093{\scriptstyle \pm 0.069}\) \\
\textit{Default} & \(2.7467{\scriptstyle \pm 1.176}\) & \(1.9903{\scriptstyle \pm 1.094}\) & \(1.5135{\scriptstyle \pm 0.117}\) & \(2.2105{\scriptstyle \pm 0.000}\) & \(2.1417{\scriptstyle \pm 0.116}\) & \(1.3138{\scriptstyle \pm 0.176}\) & \(1.2053{\scriptstyle \pm 0.040}\) & \(1.5399{\scriptstyle \pm 0.615}\) & \(5.3066{\scriptstyle \pm 0.654}\) & \(1.6712{\scriptstyle \pm 0.209}\) \\
\textit{Diamonds} & \(1.1531{\scriptstyle \pm 0.059}\) & \(1.0982{\scriptstyle \pm 0.033}\) & \(2.0167{\scriptstyle \pm 0.101}\) & \(1.8487{\scriptstyle \pm 0.000}\) & \(1.3069{\scriptstyle \pm 0.009}\) & \(1.1529{\scriptstyle \pm 0.013}\) & \(1.0989{\scriptstyle \pm 0.020}\) & \(1.0721{\scriptstyle \pm 0.019}\) & \(1.0446{\scriptstyle \pm 0.016}\) & \(4.1604{\scriptstyle \pm 0.275}\) \\
\textit{Heart} & \(4.1689{\scriptstyle \pm 2.114}\) & \(2.2363{\scriptstyle \pm 0.770}\) & \(2.8924{\scriptstyle \pm 1.355}\) & \(6.0000{\scriptstyle \pm 0.000}\) & \(26.0000{\scriptstyle \pm 0.000}\) & \(6.1792{\scriptstyle \pm 3.822}\) & \(2.5880{\scriptstyle \pm 1.086}\) & \(3.1417{\scriptstyle \pm 1.073}\) & \(11.1667{\scriptstyle \pm 4.065}\) & \(2.9773{\scriptstyle \pm 1.528}\) \\
\textit{Jasmine} & \(3.0117{\scriptstyle \pm 0.670}\) & \(1.4277{\scriptstyle \pm 0.125}\) & \(2.2063{\scriptstyle \pm 0.160}\) & \(5.1667{\scriptstyle \pm 0.000}\) & \(179.2250{\scriptstyle \pm 74.877}\) & \(4.4896{\scriptstyle \pm 4.162}\) & \(336.2175{\scriptstyle \pm 204.257}\) & \(1.3864{\scriptstyle \pm 0.137}\) & -- & \(1.5088{\scriptstyle \pm 0.172}\) \\
\textit{Liver} & \(3.4452{\scriptstyle \pm 1.116}\) & \(2.5318{\scriptstyle \pm 0.914}\) & \(2.6779{\scriptstyle \pm 0.849}\) & \(4.6000{\scriptstyle \pm 0.000}\) & \(14.3500{\scriptstyle \pm 5.776}\) & \(3.8750{\scriptstyle \pm 1.907}\) & \(2.7193{\scriptstyle \pm 0.976}\) & \(3.2900{\scriptstyle \pm 1.557}\) & \(12.4000{\scriptstyle \pm 4.659}\) & \(2.3333{\scriptstyle \pm 0.000}\) \\
\textit{Magic} & \(1.2153{\scriptstyle \pm 0.133}\) & \(1.1471{\scriptstyle \pm 0.038}\) & \(1.0934{\scriptstyle \pm 0.025}\) & \(9.0645{\scriptstyle \pm 0.000}\) & \(2.0099{\scriptstyle \pm 0.234}\) & \(1.1348{\scriptstyle \pm 0.086}\) & \(1.4130{\scriptstyle \pm 0.186}\) & \(1.2422{\scriptstyle \pm 0.179}\) & \(112.3600{\scriptstyle \pm 71.578}\) & \(1.1601{\scriptstyle \pm 0.000}\) \\
\textit{SpeedDating} & \(2.3761{\scriptstyle \pm 0.324}\) & \(1.8952{\scriptstyle \pm 0.229}\) & \(2.5269{\scriptstyle \pm 0.294}\) & \(47.7500{\scriptstyle \pm 0.000}\) & \(2.4987{\scriptstyle \pm 0.291}\) & \(5.6823{\scriptstyle \pm 0.781}\) & \(654.5083{\scriptstyle \pm 381.861}\) & \(2.1643{\scriptstyle \pm 0.200}\) & -- & \(1.7783{\scriptstyle \pm 0.129}\) \\
\textit{DNA} & \(1.3362{\scriptstyle \pm 0.090}\) & \(1.5581{\scriptstyle \pm 0.298}\) & \(2.7525{\scriptstyle \pm 4.189}\) & \(3.7500{\scriptstyle \pm 0.000}\) & \(38.5008{\scriptstyle \pm 24.136}\) & \(4.5089{\scriptstyle \pm 1.957}\) & \(3.6837{\scriptstyle \pm 0.539}\) & \(2.7289{\scriptstyle \pm 1.154}\) & -- & \(1.9511{\scriptstyle \pm 0.312}\) \\
\textit{DARWIN} & \(3.9152{\scriptstyle \pm 1.944}\) & \(2.8598{\scriptstyle \pm 1.908}\) & \(5.1759{\scriptstyle \pm 2.720}\) & \(7.0000{\scriptstyle \pm 0.000}\) & \(13.0917{\scriptstyle \pm 5.863}\) & \(3.6742{\scriptstyle \pm 1.005}\) & \(24.1524{\scriptstyle \pm 9.551}\) & \(2.5042{\scriptstyle \pm 0.878}\) & -- & \(5.5708{\scriptstyle \pm 2.564}\) \\
\midrule
\textbf{Average} & 2.7547 & 1.8493 & 2.3679 & 10.0862 & 63.5033 & 4.1047 & 86.1523 & 2.4608 & 19.1134 & 2.3501 \\
\bottomrule
\end{tabular*}
\end{sidewaystable}
\FloatBarrier

\section{Visualization of tree-derived variable dependence}
\label{appendix:cart}

\begin{figure*}[htbp]
\internallinenumbers
\centering
\includegraphics[width=\textwidth]{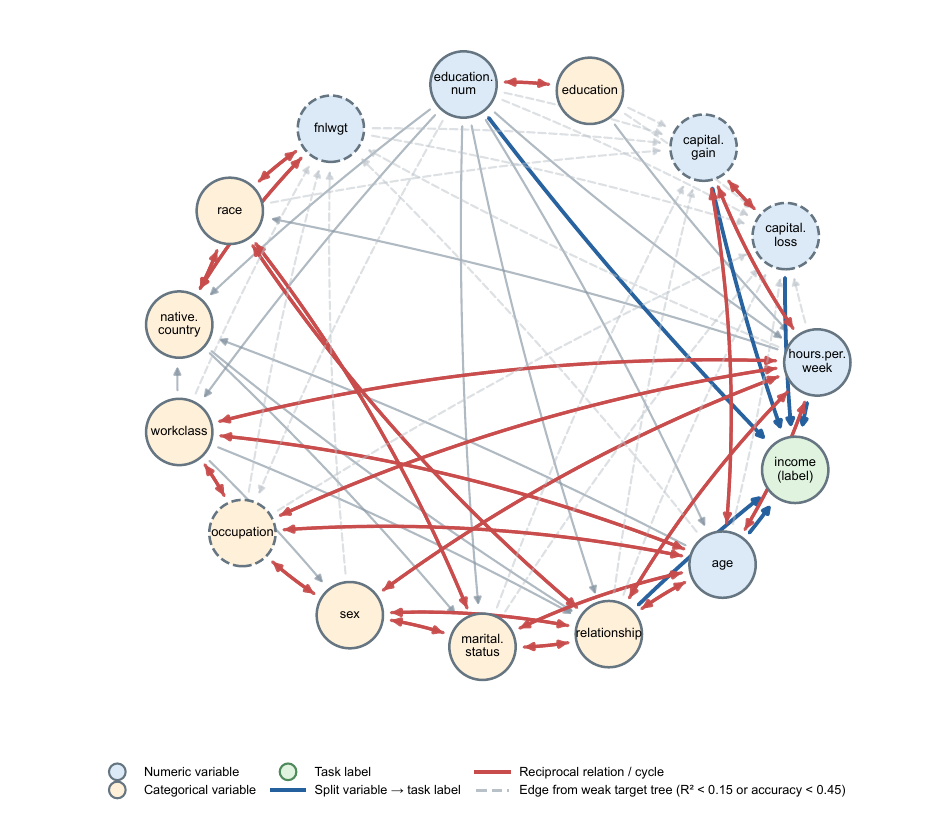}
\caption{\textbf{Raw variable-dependence signals extracted by the chained
trees for the \textit{Adult} dataset.}
Nodes represent the input variables and task label. A directed edge
\(X_j \rightarrow X_i\) indicates that \(X_j\) was selected as a splitting
variable by the CART fitted to predict \(X_i\). Because variables can appear
in one another's prediction trees, the resulting graph may contain reciprocal,
redundant, or cyclic dependence signals. Grey dashed edges denote signals
extracted from trees with weak predictive performance, defined as regression
trees with \(R^2 < 0.15\) or classification trees with accuracy \(< 0.45\).
These edges represent predictive associations rather than causal relationships.}
\label{fig:cart}
\end{figure*}

\FloatBarrier
Supplementary Fig.~\ref{fig:cart} visualizes the raw variable-dependence
signals extracted by the chained trees for the \textit{Adult} dataset.
The graph contains overlapping,
reciprocal, and cyclic associations and should not be interpreted as a coherent
causal structure.

\FloatBarrier
\FloatBarrier
\section{Prompt and generated strategy for the \textit{Adult} dataset}
\label{appendix:adult_case}

This section presents the original prompt and the corresponding LLM-generated 
synthesis strategy for the Adult dataset. Both are reproduced verbatim to 
preserve the actual strategy-generation process and may differ from the terminology used in the manuscript.

\subsection{Original prompt provided to the LLM}
\label{appendix:adult_original_prompt}

According to the prompt construction described in our framework, the prompt is 
conceptually composed of:
\begin{equation}
p =
p_{\mathrm{task}}
\oplus
p_{\mathrm{metric}}
\oplus
p_{\mathrm{requirement}}
\oplus
p_{\mathrm{dep}}
\oplus
p_{\mathrm{output}}.
\label{eq:adult_prompt_composition}
\end{equation}
Here, \(p_{\mathrm{dep}}\) contains both the feature-wise structural summaries 
and the complete label tree structure. The component symbols are provided only 
for explanation and are not inserted into the original prompt text.

\begin{promptbox}{Prompt Part I: Task Specification and Evaluation Criteria ($p_{\mathrm{task}} \oplus p_{\mathrm{metric}}$)}
Your task is to generate a tabular data synthesis strategy by inferring the 
original data's intrinsic characteristics from the provided pre-trained 
decision trees and feature information. The synthesized data will replace the 
original data for model training, with the core goal of making the performance 
of models trained on synthetic data as close as possible to those trained on 
the original data.

# Metrics
When evaluating synthetic data, typically considering the following metrics:

1. ML Efficiency:
   A core metric reflecting synthetic data's support for downstream ML tasks, 
   evaluated via the TSTR (Train on Synthetic, Test on Real) protocol.

2. Fidelity:
   Measures preservation of raw data statistical properties from two aspects:
   - Pairwise: Uses DPCM (numerical features), DCSM (categorical features), 
     and contingency similarity matrices (mixed features) to assess 
     feature-pair dependencies.
   - Column-wise: Compares individual features; KST is commonly used for 
     numerical features.

3. Privacy:
   Privacy is evaluated using split-ratio-adjusted DCR risk, which measures 
   whether synthetic records exhibit excessive proximity to training records 
   beyond the expected preference induced by the train--test split ratio. 
   A lower adjusted DCR value indicates lower potential memorization risk.

The synthetic data you generate should perform well on these metrics.
\end{promptbox}

\begin{promptbox}{Prompt Part II: Core Requirements for Strategy Generation ($p_{\mathrm{requirement}}$)}
# Core Requirements for Your Strategy

Multiple decision tree models are provided. The final decision tree takes the 
original label of the raw data as the prediction target. All preceding decision 
trees are trained with a single data feature as the target and all other 
features, excluding the original label, as inputs. Their outputs are restricted 
to structured summaries, for example:

{feature_0}:{type:continuous,min:0.0,max:1.0,
associated_features:[feature_1,feature_3,feature_89],
performance:{R2:0.992,RMSE:0.0379}}

The features listed in the associated_features field participate in node 
splitting in the decision tree trained to predict the corresponding target 
feature. Their order follows the top-down splitting order of the original tree.

You need to analyse the underlying feature dependency patterns and distribution 
characteristics based on the provided decision tree information. You should 
then construct a routing function to assign each sample to an appropriate 
subspace. The routing function should reflect your interpretation of the 
data structure rather than merely reproducing tree rules.

Within each routed subspace, use suitable model fitting procedures to learn the 
distribution characteristics of real data and support the generation of 
high-quality synthetic samples.

Attention:
Decision tree rules are not intended for direct record sampling. They should be 
analysed to infer feature dependencies and data distribution characteristics. 
Your output must incorporate this analysis when designing the synthesis 
procedure.
\end{promptbox}

\begin{promptbox}{Prompt Part III: Feature inter-dependency for \textit{Adult} ($ p_{\mathrm{dep}}$)}

{feature_0}:{type:continuous,min:17.0,max:90.0,
associated_features:[feature_5,feature_12,feature_7,feature_1,feature_6,feature_4,feature_10],
performance:{R2:0.4227,RMSE:10.3665}}

{feature_1}:{type:categorical,ordinal_encoding_range:[0,8],class_num:9,
associated_features:[feature_6,feature_0,feature_4,feature_12],
performance:{accuracy:0.7553}}

{feature_2}:{type:continuous,min:12285.0,max:1484705.0,
associated_features:[feature_13,feature_8,feature_0,feature_9,feature_1,feature_6,feature_12],
performance:{R2:0.0549,RMSE:102680.6073}}

{feature_3}:{type:categorical,ordinal_encoding_range:[0,15],class_num:16,
associated_features:[feature_4],
performance:{accuracy:1.0}}

{feature_4}:{type:continuous,min:1.0,max:16.0,
associated_features:[feature_3],
performance:{R2:1.0,RMSE:0.0}}

{feature_5}:{type:categorical,ordinal_encoding_range:[0,6],class_num:7,
associated_features:[feature_7,feature_0,feature_9,feature_4,feature_8,feature_13],
performance:{accuracy:0.8419}}

{feature_6}:{type:categorical,ordinal_encoding_range:[0,14],class_num:15,
associated_features:[feature_1,feature_4,feature_9,feature_0,feature_12],
performance:{accuracy:0.3717}}

{feature_7}:{type:categorical,ordinal_encoding_range:[0,5],class_num:6,
associated_features:[feature_5,feature_9,feature_0,feature_12,feature_4,feature_1,feature_13,feature_8],
performance:{accuracy:0.7891}}

{feature_8}:{type:categorical,ordinal_encoding_range:[0,4],class_num:5,
associated_features:[feature_13,feature_7,feature_12,feature_2,feature_5],
performance:{accuracy:0.8752}}

{feature_9}:{type:categorical,ordinal_encoding_range:[0,1],class_num:2,
associated_features:[feature_7,feature_6,feature_1,feature_5,feature_12],
performance:{accuracy:0.8336}}

{feature_10}:{type:continuous,min:0.0,max:99999.0,
associated_features:[feature_4,feature_0,feature_12,feature_11,feature_2,feature_3,feature_8,feature_7,feature_5],
performance:{R2:0.1142,RMSE:6927.3273}}

{feature_11}:{type:continuous,min:0.0,max:4356.0,
associated_features:[feature_4,feature_7,feature_10,feature_5,feature_2,feature_0,feature_12,feature_3,feature_6],
performance:{R2:0.0354,RMSE:394.4608}}

{feature_12}:{type:continuous,min:1.0,max:99.0,
associated_features:[feature_0,feature_3,feature_4,feature_9,feature_10,feature_1,feature_7,feature_6],
performance:{R2:0.2214,RMSE:10.9254}}

{feature_13}:{type:categorical,ordinal_encoding_range:[0,41],class_num:42,
associated_features:[feature_8,feature_2,feature_4,feature_0,feature_1],
performance:{accuracy:0.9031}}
\end{promptbox}

\begin{promptbox}{Prompt Part IV: Complete Label-Tree Structure ($ p_{\mathrm{dep}}$)}
feature_14 (label) (categorical) ordinal encoding range is integers in [0, 1]:
accuracy: 0.8545

--- feature_7 <= 0.5
|   --- feature_4 <= 12.5
|   |   --- feature_10 <= 5095.5
|   |   |   --- feature_11 <= 1782.5
|   |   |   |   --- feature_4 <= 8.5
|   |   |   |   |   --- class: 0 [leaf_id:1]
|   |   |   |   |       Sample Count 1247, Label Ratio {'0': 0.901, '1': 0.099}
|   |   |   |   --- feature_4 > 8.5
|   |   |   |   |   --- feature_0 <= 35.5
|   |   |   |   |   |   --- class: 0 [leaf_id:2]
|   |   |   |   |   |       Sample Count 2042, Label Ratio {'0': 0.800, '1': 0.200}
|   |   |   |   |   --- feature_0 > 35.5
|   |   |   |   |   |   --- feature_12 <= 34.5
|   |   |   |   |   |   |   --- class: 0 [leaf_id:3]
|   |   |   |   |   |   |       Sample Count 376, Label Ratio {'0': 0.899, '1': 0.101}
|   |   |   |   |   |   --- feature_12 > 34.5
|   |   |   |   |   |   |   --- feature_4 <= 9.5
|   |   |   |   |   |   |   |   --- class: 0 [leaf_id:4]
|   |   |   |   |   |   |   |       Sample Count 2136, Label Ratio {'0': 0.659, '1': 0.341}
|   |   |   |   |   |   |   --- feature_4 > 9.5
|   |   |   |   |   |   |   |   --- class: 0 [leaf_id:5]
|   |   |   |   |   |   |   |       Sample Count 1716, Label Ratio {'0': 0.538, '1': 0.462}
|   |   |   --- feature_11 > 1782.5
|   |   |   |   --- feature_11 <= 1989.5
|   |   |   |   |   --- class: 1 [leaf_id:6]
|   |   |   |   |       Sample Count 209, Label Ratio {'0': 0.053, '1': 0.947}
|   |   |   |   --- feature_11 > 1989.5
|   |   |   |   |   --- class: 0 [leaf_id:7]
|   |   |   |   |       Sample Count 82, Label Ratio {'0': 0.829, '1': 0.171}
|   |   --- feature_10 > 5095.5
|   |   |   --- class: 1 [leaf_id:8]
|   |   |       Sample Count 403, Label Ratio {'0': 0.020, '1': 0.980}
|   --- feature_4 > 12.5
|   |   --- feature_10 <= 5095.5
|   |   |   --- feature_11 <= 1782.5
|   |   |   |   --- feature_12 <= 31.0
|   |   |   |   |   --- class: 0 [leaf_id:9]
|   |   |   |   |       Sample Count 207, Label Ratio {'0': 0.691, '1': 0.309}
|   |   |   |   --- feature_12 > 31.0
|   |   |   |   |   --- feature_0 <= 28.5
|   |   |   |   |   |   --- class: 0 [leaf_id:10]
|   |   |   |   |   |       Sample Count 149, Label Ratio {'0': 0.624, '1': 0.376}
|   |   |   |   |   --- feature_0 > 28.5
|   |   |   |   |   |   --- class: 1 [leaf_id:11]
|   |   |   |   |   |       Sample Count 2321, Label Ratio {'0': 0.318, '1': 0.682}
|   |   |   --- feature_11 > 1782.5
|   |   |   |   --- class: 1 [leaf_id:12]
|   |   |   |       Sample Count 309, Label Ratio {'0': 0.036, '1': 0.964}
|   |   --- feature_10 > 5095.5
|   |   |   --- class: 1 [leaf_id:13]
|   |   |       Sample Count 533, Label Ratio {'0': 0.004, '1': 0.996}
--- feature_7 > 0.5
|   --- feature_10 <= 7073.5
|   |   --- feature_7 <= 4.5
|   |   |   --- feature_4 <= 13.5
|   |   |   |   --- feature_12 <= 42.5
|   |   |   |   |   --- class: 0 [leaf_id:14]
|   |   |   |   |       Sample Count 11927, Label Ratio {'0': 0.979, '1': 0.021}
|   |   |   |   --- feature_12 > 42.5
|   |   |   |   |   --- class: 0 [leaf_id:15]
|   |   |   |   |       Sample Count 2793, Label Ratio {'0': 0.890, '1': 0.110}
|   |   |   --- feature_4 > 13.5
|   |   |   |   --- feature_12 <= 43.5
|   |   |   |   |   --- class: 0 [leaf_id:16]
|   |   |   |   |       Sample Count 494, Label Ratio {'0': 0.836, '1': 0.164}
|   |   |   |   --- feature_12 > 43.5
|   |   |   |   |   --- class: 0 [leaf_id:17]
|   |   |   |   |       Sample Count 318, Label Ratio {'0': 0.594, '1': 0.406}
|   |   --- feature_7 > 4.5
|   |   |   --- feature_4 <= 10.5
|   |   |   |   --- class: 0 [leaf_id:18]
|   |   |   |       Sample Count 784, Label Ratio {'0': 0.684, '1': 0.316}
|   |   |   --- feature_4 > 10.5
|   |   |   |   --- class: 1 [leaf_id:19]
|   |   |   |       Sample Count 503, Label Ratio {'0': 0.364, '1': 0.636}
|   --- feature_10 > 7073.5
|   |   --- class: 1 [leaf_id:20]
|   |       Sample Count 395, Label Ratio {'0': 0.033, '1': 0.967}
\end{promptbox}

\begin{promptbox}{Prompt Part V: Output and Implementation Constraints}
# Output Requirements

Please carefully examine the decision tree rules and adhere to the core 
requirements.

You must write a complete and executable function to generate samples covering 
all features. The resulting synthetic data will replace the original dataset.

Additional requirements:
1. The generated synthetic data must be saved in CSV format.
2. Only output the function generate_synthesis_data.
3. All library imports and auxiliary functions must be contained within 
   generate_synthesis_data.
4. Categorical features must use ordinal encoding and be represented directly 
   as integers rather than strings.
5. If probabilities are used, they must sum to 1.
6. Generated values must not exceed the predefined feature ranges.
7. You may select different distribution-fitting models for different feature 
   types or subspaces, provided that feature dependencies are appropriately 
   considered.

# Output Format

def generate_synthesis_data(save_path: str, n_sample: int, 
                            X_train: np.ndarray, y_train: np.ndarray) -> None:
    import pandas as pd
    import csv
    ...

    fieldnames = [f"feature_{i}" for i in range(14)] + ["label"]
    with open(save_path, "w", newline="") as csv_file:
        writer = csv.DictWriter(csv_file, fieldnames=fieldnames)
        writer.writeheader()
        writer.writerows(synthesis_samples)
\end{promptbox}

\begin{promptbox}{Prompt Part VI: Reference Analysis Provided to the LLM ($ p_{\mathrm{output}}$)}
# Reference Analysis

An analysis of the provided decision tree rules reveals a complex web of 
inter-feature dependencies. The goal is to construct a generative process that 
honors these relationships to produce high-fidelity synthetic data. The strategy 
is founded on the principle of divide and conquer: partitioning the data space 
into smaller, more homogeneous subspaces and then modelling the distribution 
within each subspace.

1. Feature Dependency Analysis
- Hub Features: feature_4, feature_7, and feature_0 appear frequently as 
  splitting variables, indicating that they are central to the data structure.
- Strong Co-dependency: feature_3 and feature_4 exhibit a near-deterministic 
  relationship. Once feature_4 is generated, feature_3 can be generated through 
  an explicit mapping.
- Cyclic Dependencies: dependencies such as feature_5 relying on feature_7 and 
  feature_7 relying on feature_5 imply that the generation strategy must adopt 
  a practical ordering to break cycles.

2. Subspace-based Synthesis Strategy
- Partition the data using major task-relevant attributes identified from the 
  label tree, especially feature_7 and feature_4.
- Fit conditional generative models within each subspace rather than learning 
  one global distribution.
- Use probabilistic classification models for categorical attributes.
- Use regression models with residual sampling for continuous attributes.
- Explicitly preserve the deterministic mapping from feature_4 to feature_3.

3. Data Generation and Validation
- Allocate generated samples according to the empirical sizes of routed 
  subspaces.
- Aggregate all generated records into a complete synthetic dataset.
- Enforce feature type, range, and encoding constraints before saving the CSV.
\end{promptbox}

\subsection{LLM-generated analysis and synthesis strategy}
\label{appendix:adult_generated_strategy}


\begin{lstlisting}[
    style=textlisting,
    numbers=none,
    caption={Original LLM output containing the analysis and generated synthesis strategy for the \textit{Adult} dataset.},
    label={lst:adult_generated_strategy}
]
To synthesize the data effectively, I will follow a "Divide and Conquer" approach based on the provided decision tree rules.

### Analysis of Decision Tree Rules and Feature Dependencies

1.  **Hub Features and Partitioning:**
    *   **Feature 7:** This is the root of the label decision tree. It also appears as a primary split for features 5 and 9. This is the most natural candidate for the primary routing split.
    *   **Feature 4 and Feature 3:** These features have a perfect deterministic relationship ($R^2=1$, Accuracy=1). They are essentially two representations of the same information. In the label tree, feature 4 is a critical secondary splitter.
    *   **Feature 10 and Feature 11:** These features show high sparsity (min=0) and appear as important split features in the label tree.

2.  **Dependency Chain:**
    *   **Root Features:** feature 0, feature 9, and feature 13 tend to be independent or act as root nodes.
    *   **Related Features:** feature 4 influences feature 1, feature 6, and feature 12.
    *   **Target:** Label is heavily conditioned on feature 7, feature 4, features 10 and feature 11.

### Synthesis Strategy

1.  **Routing Function:** I will partition the data into 4 major subspaces based on the primary splits of the label tree: feature 7 (<= 0.5 vs > 0.5) and feature 4 (<= 12.5 vs > 12.5). This ensures that the most important decision boundaries for ML Efficiency are preserved.
2.  **Generative Chain within Subspaces:**
    *   **Step 1:** Fit and sample "Root Features" (feature 7, feature 4, feature 0) using their empirical joint distribution in the subspace.
    *   **Step 2:** Generate feature 3 via a direct mapping from feature 4.
    *   **Step 3:** Generate remaining categorical features (feature 1, feature 5, feature 6, feature 8, feature 9, feature 13) using DecisionTreeClassifier models conditioned on the root features. To maintain fidelity, I will sample from the predicted probability distributions rather than taking the argmax.
    *   **Step 4:** Generate continuous features (feature 2, feature 10, feature 11, feature 12) using DecisionTreeRegressor models. To preserve variance and privacy, I will add sampled residuals from the training data to the model predictions.
    *   **Step 5:** Generate the label using a DecisionTreeClassifier conditioned on all significant predictors identified in the rules.

3.  **Refining Constraints:** I will strictly enforce the min and max values provided in the rules for continuous features and use ordinal_encoding_range for categorical features.

'''python
import numpy as np
import pandas as pd
from sklearn.tree import DecisionTreeClassifier, DecisionTreeRegressor
import csv

def generate_synthesis_data(save_path: str, n_sample: int, X_train: np.ndarray, y_train: np.ndarray) -> None:
    # Combine X and y for training modeling
    full_data = np.column_stack((X_train, y_train))
    feat_cols = [f"feature_{i}" for i in range(14)]
    all_cols = feat_cols + ["label"]
    df = pd.DataFrame(full_data, columns=all_cols)

    # Feature metadata derived from rules
    continuous_features = {
        'feature_0': (17.0, 90.0),
        'feature_2': (12285.0, 1484705.0),
        'feature_4': (1.0, 16.0),
        'feature_10': (0.0, 99999.0),
        'feature_11': (0.0, 4356.0),
        'feature_12': (1.0, 99.0)
    }
    categorical_features = {
        'feature_1': 9, 'feature_3': 16, 'feature_5': 7, 'feature_6': 15,
        'feature_7': 6, 'feature_8': 5, 'feature_9': 2, 'feature_13': 42,
        'label': 2
    }

    # Define subspaces based on the label decision tree splits
    df['subspace'] = 0
    df.loc[(df['feature_7'] <= 0.5) & (df['feature_4'] <= 12.5), 'subspace'] = 0
    df.loc[(df['feature_7'] <= 0.5) & (df['feature_4'] > 12.5), 'subspace'] = 1
    df.loc[(df['feature_7'] > 0.5) & (df['feature_4'] <= 12.5), 'subspace'] = 2
    df.loc[(df['feature_7'] > 0.5) & (df['feature_4'] > 12.5), 'subspace'] = 3

    subspace_counts = df['subspace'].value_counts(normalize=True)
    
    synthetic_data = []

    for sub_id in range(4):
        sub_df = df[df['subspace'] == sub_id]
        if len(sub_df) < 5: continue # Safety for empty subspaces
        
        n_sub_samples = int(n_sample * subspace_counts.get(sub_id, 0))
        if n_sub_samples <= 0: continue

        # Local Models for this subspace
        # Order: 7, 4, 0 -> 3 -> 9, 5 -> 1, 6, 12 -> 10, 11 -> 2, 8, 13 -> label
        
        # 1. Roots: 7, 4, 0 (Sampled directly to maintain core structure)
        indices = np.random.choice(sub_df.index, size=n_sub_samples, replace=True)
        roots = sub_df.loc[indices, ['feature_7', 'feature_4', 'feature_0']].values
        
        sub_gen = pd.DataFrame(roots, columns=['feature_7', 'feature_4', 'feature_0'])
        
        # 2. Map 3 from 4 (Deterministic rule)
        # We learn the most frequent mapping in this subspace
        map_3_4 = sub_df.groupby('feature_4')['feature_3'].agg(lambda x: x.value_counts().index[0]).to_dict()
        sub_gen['feature_3'] = sub_gen['feature_4'].map(map_3_4).fillna(sub_df['feature_3'].mode()[0])

        # 3. Conditional Generators
        def fit_sample_cat(target, predictors, current_gen_df):
            clf = DecisionTreeClassifier(max_depth=5)
            clf.fit(sub_df[predictors], sub_df[target])
            probs = clf.predict_proba(current_gen_df[predictors])
            # Sample from probabilities
            return [np.random.choice(clf.classes_, p=p) for p in probs]

        def fit_sample_cont(target, predictors, current_gen_df):
            reg = DecisionTreeRegressor(max_depth=5)
            reg.fit(sub_df[predictors], sub_df[target])
            preds = reg.predict(current_gen_df[predictors])
            # Add residuals
            residuals = sub_df[target] - reg.predict(sub_df[predictors])
            sampled_res = np.random.choice(residuals, size=len(preds), replace=True)
            res = preds + sampled_res
            # Clip to range
            low, high = continuous_features[target]
            return np.clip(res, low, high)

        # Sequence of generation
        sub_gen['feature_9'] = fit_sample_cat('feature_9', ['feature_7', 'feature_0'], sub_gen)
        sub_gen['feature_5'] = fit_sample_cat('feature_5', ['feature_7', 'feature_0', 'feature_9'], sub_gen)
        sub_gen['feature_1'] = fit_sample_cat('feature_1', ['feature_4', 'feature_0', 'feature_5'], sub_gen)
        sub_gen['feature_6'] = fit_sample_cat('feature_6', ['feature_1', 'feature_4', 'feature_9'], sub_gen)
        sub_gen['feature_12'] = fit_sample_cont('feature_12', ['feature_0', 'feature_4', 'feature_6'], sub_gen)
        sub_gen['feature_10'] = fit_sample_cont('feature_10', ['feature_4', 'feature_0', 'feature_12'], sub_gen)
        sub_gen['feature_11'] = fit_sample_cont('feature_11', ['feature_4', 'feature_7', 'feature_10'], sub_gen)
        sub_gen['feature_2'] = fit_sample_cont('feature_2', ['feature_0', 'feature_7', 'feature_12'], sub_gen)
        sub_gen['feature_8'] = fit_sample_cat('feature_8', ['feature_7', 'feature_9', 'feature_2'], sub_gen)
        sub_gen['feature_13'] = fit_sample_cat('feature_13', ['feature_8', 'feature_2', 'feature_0'], sub_gen)
        sub_gen['label'] = fit_sample_cat('label', ['feature_7', 'feature_4', 'feature_10', 'feature_11', 'feature_0', 'feature_12'], sub_gen)

        synthetic_data.append(sub_gen)

    # Combine and shuffle
    final_df = pd.concat(synthetic_data).sample(frac=1).reset_index(drop=True)
    
    # Final data type correction
    for col in categorical_features:
        final_df[col] = final_df[col].astype(int)

    # Reorder columns to match fieldnames
    final_df = final_df[feat_cols + ["label"]]

    # Save to CSV
    final_df.to_csv(save_path, index=False)
'''
\end{lstlisting}
\FloatBarrier

\clearpage

\section{Ablation results}
\label{sec:ablation_results}

The ablation experiments are conducted on eight datasets for which all evaluated variants can be accommodated within the LLM context window. In particular, the Full Feature Trees variant retains the complete feature-tree structures and therefore produces substantially longer prompts; for the remaining high-dimensional datasets, these prompts exceed the context-window limit. We therefore restrict the ablation analysis to datasets on which all variants can be evaluated under the same experimental protocol, ensuring a consistent and fair comparison. Across the following ablations, we focus on overall performance trends across datasets rather than requiring each design choice to improve every individual metric on every dataset.

\subsection{Contribution of feature-dependence extraction}
\label{subsec:contribution_of_feature_dependence_extraction}

This section reports detailed ablation results on feature-dependence extraction. We compare the proposed tree-based extraction mechanism with two alternative mechanisms based on linear models and mutual information. To isolate the effect of feature-wise dependence extraction, the complete label tree and all other prompt components are kept unchanged across all variants. Across the evaluated datasets, the chained-tree mechanism provides a stronger overall balance of marginal fidelity, pairwise-dependence preservation, and downstream task performance than the alternative extraction mechanisms, although the relative advantage varies across individual datasets and metrics. Detailed per-dataset results are reported in Supplementary Table~\ref{tab:feature_dependence_extraction_ablation}.

\begin{table*}[htbp]
\centering
\caption{Ablation results for feature-dependence extraction mechanisms. Values are reported as mean $\pm$ s.d. Bold values indicate the best result for each dataset.}
\label{tab:feature_dependence_extraction_ablation}
\scriptsize
\setlength{\tabcolsep}{2pt}
\renewcommand{\arraystretch}{1.08}

\begin{adjustbox}{max width=\linewidth}
\begin{tabular}{l|ccc|ccc|ccc}
\toprule
\multirow{2}{*}{Dataset}
& \multicolumn{3}{c|}{Linear Models}
& \multicolumn{3}{c|}{Mutual Information}
& \multicolumn{3}{c}{Chained Trees (Ours)} \\
\cmidrule(lr){2-4}
\cmidrule(lr){5-7}
\cmidrule(lr){8-10}
& \shortstack{Marginal distribution\\error ($\downarrow$)}
& \shortstack{Pairwise correlation\\error ($\downarrow$)}
& Task$^{a}$
& \shortstack{Marginal distribution\\error ($\downarrow$)}
& \shortstack{Pairwise correlation\\error ($\downarrow$)}
& Task$^{a}$
& \shortstack{Marginal distribution\\error ($\downarrow$)}
& \shortstack{Pairwise correlation\\error ($\downarrow$)}
& Task$^{a}$ \\
\midrule

\textit{Adult}
& $12.31\%_{\pm 0.000}$ & $9.78\%_{\pm 0.000}$ & $\boldsymbol{0.903_{\pm 0.001}}$
& $9.93\%_{\pm 0.031}$ & $11.90\%_{\pm 0.000}$ & $0.892_{\pm 0.002}$
& $\boldsymbol{5.44\%_{\pm 0.001}}$ & $\boldsymbol{3.53\%_{\pm 0.000}}$ & $0.900_{\pm 0.001}$ \\

\textit{Blood}
& $8.64\%_{\pm 0.000}$ & $2.91\%_{\pm 0.000}$ & $\boldsymbol{0.712_{\pm 0.000}}$
& $8.38\%_{\pm 0.004}$ & $3.21\%_{\pm 0.004}$ & $0.699_{\pm 0.055}$
& $\boldsymbol{3.32\%_{\pm 0.005}}$ & $\boldsymbol{2.68\%_{\pm 0.004}}$ & $0.704_{\pm 0.026}$ \\

\textit{Car}
& $8.67\%_{\pm 0.003}$ & $17.70\%_{\pm 0.003}$ & $\boldsymbol{0.969_{\pm 0.004}}$
& $\boldsymbol{1.64\%_{\pm 0.003}}$ & $\boldsymbol{3.84\%_{\pm 0.003}}$ & $0.961_{\pm 0.006}$
& $1.68\%_{\pm 0.004}$ & $3.99\%_{\pm 0.004}$ & $0.958_{\pm 0.006}$ \\

\textit{Default}
& $10.57\%_{\pm 0.001}$ & $\boldsymbol{4.74\%_{\pm 0.001}}$ & $0.751_{\pm 0.004}$
& $8.48\%_{\pm 0.000}$ & $5.72\%_{\pm 0.000}$ & $0.735_{\pm 0.005}$
& $\boldsymbol{1.00\%_{\pm 0.001}}$ & $5.09\%_{\pm 0.001}$ & $\boldsymbol{0.752_{\pm 0.002}}$ \\

\textit{Diamonds}
& $\boldsymbol{3.50\%_{\pm 0.001}}$ & $3.18\%_{\pm 0.000}$ & $1287.506_{\pm 125.968}$
& $4.29\%_{\pm 0.000}$ & $3.58\%_{\pm 0.000}$ & $1262.624_{\pm 13.715}$
& $3.79\%_{\pm 0.000}$ & $\boldsymbol{2.11\%_{\pm 0.000}}$ & $\boldsymbol{934.090_{\pm 15.111}}$ \\

\textit{Heart}
& $\boldsymbol{5.94\%_{\pm 0.004}}$ & $11.17\%_{\pm 0.004}$ & $0.783_{\pm 0.029}$
& $8.42\%_{\pm 0.007}$ & $13.96\%_{\pm 0.008}$ & $\boldsymbol{0.852_{\pm 0.019}}$
& $6.08\%_{\pm 0.007}$ & $\boldsymbol{11.11\%_{\pm 0.006}}$ & $0.823_{\pm 0.028}$ \\

\textit{Liver}
& $14.94\%_{\pm 0.000}$ & $5.21\%_{\pm 0.000}$ & $3.291_{\pm 0.000}$
& $11.62\%_{\pm 0.007}$ & $\boldsymbol{3.15\%_{\pm 0.006}}$ & $3.421_{\pm 0.088}$
& $\boldsymbol{9.20\%_{\pm 0.000}}$ & $3.49\%_{\pm 0.000}$ & $\boldsymbol{3.181_{\pm 0.000}}$ \\

\textit{Magic}
& $\boldsymbol{3.02\%_{\pm 0.000}}$ & $\boldsymbol{0.98\%_{\pm 0.000}}$ & $\boldsymbol{0.920_{\pm 0.000}}$
& $3.57\%_{\pm 0.000}$ & $1.08\%_{\pm 0.000}$ & $0.918_{\pm 0.001}$
& $3.09\%_{\pm 0.000}$ & $\boldsymbol{0.98\%_{\pm 0.000}}$ & $\boldsymbol{0.920_{\pm 0.000}}$ \\
\bottomrule
\end{tabular}
\end{adjustbox}

\vspace{3pt}
\parbox{\linewidth}{\footnotesize
Task$^{a}$ is AUROC ($\uparrow$) for classification and RMSE ($\downarrow$) for regression. The complete label tree and all other prompt components are retained across all variants.
}
\end{table*}

\subsection{Effect of dependence information encoding}
\label{subsec:effect_of_dependence_information_encoding}

This section examines how the representation and ordering of dependence information affect synthesis performance. We compare the proposed encoding with three variants: retaining the full feature-tree structures, replacing the complete label tree with a compact summary, and randomly shuffling the order of associated features. Overall, the proposed representation provides a favourable balance across the evaluated metrics and datasets, while individual variants can perform better on particular datasets or metrics. These results indicate that compact representation and structured ordering are useful at the aggregate level rather than uniformly improving every individual evaluation measure. The corresponding results are reported in Supplementary Table~\ref{tab:dependence_information_encoding_ablation}.

\begin{table*}[htbp]
\centering
\caption{Ablation results for dependence-information representation and ordering. Values are reported as mean $\pm$ s.d. Bold values indicate the best result for each dataset.}
\label{tab:dependence_information_encoding_ablation}
\scriptsize
\setlength{\tabcolsep}{2pt}
\renewcommand{\arraystretch}{1.08}

\begin{adjustbox}{max width=\linewidth}
\begin{tabular}{l|ccc|ccc|ccc|ccc}
\toprule
\multirow{2}{*}{Dataset}
& \multicolumn{3}{c|}{Full Feature Trees}
& \multicolumn{3}{c|}{Condensed Label Tree}
& \multicolumn{3}{c|}{Shuffled Dependence Order}
& \multicolumn{3}{c}{Proposed Representation (Ours)} \\
\cmidrule(lr){2-4}
\cmidrule(lr){5-7}
\cmidrule(lr){8-10}
\cmidrule(lr){11-13}
& \shortstack{Marginal distribution\\error ($\downarrow$)}
& \shortstack{Pairwise correlation\\error ($\downarrow$)}
& Task$^{a}$
& \shortstack{Marginal distribution\\error ($\downarrow$)}
& \shortstack{Pairwise correlation\\error ($\downarrow$)}
& Task$^{a}$
& \shortstack{Marginal distribution\\error ($\downarrow$)}
& \shortstack{Pairwise correlation\\error ($\downarrow$)}
& Task$^{a}$
& \shortstack{Marginal distribution\\error ($\downarrow$)}
& \shortstack{Pairwise correlation\\error ($\downarrow$)}
& Task$^{a}$ \\
\midrule

\textit{Adult}
& $6.21\%_{\pm 0.000}$ & $4.65\%_{\pm 0.000}$ & $0.897_{\pm 0.002}$
& $\boldsymbol{3.14\%_{\pm 0.000}}$ & $5.99\%_{\pm 0.000}$ & $0.887_{\pm 0.003}$
& $9.09\%_{\pm 0.000}$ & $5.53\%_{\pm 0.000}$ & $\boldsymbol{0.907_{\pm 0.001}}$
& $5.44\%_{\pm 0.001}$ & $\boldsymbol{3.53\%_{\pm 0.000}}$ & $0.900_{\pm 0.001}$ \\

\textit{Blood}
& $3.77\%_{\pm 0.006}$ & $4.02\%_{\pm 0.007}$ & $0.696_{\pm 0.035}$
& $5.92\%_{\pm 0.006}$ & $3.53\%_{\pm 0.005}$ & $0.636_{\pm 0.050}$
& $7.99\%_{\pm 0.007}$ & $3.39\%_{\pm 0.003}$ & $\boldsymbol{0.713_{\pm 0.033}}$
& $\boldsymbol{3.32\%_{\pm 0.005}}$ & $\boldsymbol{2.68\%_{\pm 0.004}}$ & $0.704_{\pm 0.026}$ \\

\textit{Car}
& $4.21\%_{\pm 0.000}$ & $7.37\%_{\pm 0.000}$ & $0.952_{\pm 0.000}$
& $1.68\%_{\pm 0.003}$ & $4.47\%_{\pm 0.003}$ & $0.943_{\pm 0.013}$
& $\boldsymbol{1.29\%_{\pm 0.002}}$ & $\boldsymbol{3.44\%_{\pm 0.002}}$ & $\boldsymbol{0.964_{\pm 0.005}}$
& $1.68\%_{\pm 0.004}$ & $3.99\%_{\pm 0.004}$ & $0.958_{\pm 0.006}$ \\

\textit{Default}
& $9.25\%_{\pm 0.001}$ & $5.48\%_{\pm 0.001}$ & $\boldsymbol{0.756_{\pm 0.003}}$
& $11.78\%_{\pm 0.000}$ & $9.54\%_{\pm 0.000}$ & $0.730_{\pm 0.000}$
& $9.99\%_{\pm 0.000}$ & $7.37\%_{\pm 0.000}$ & $0.750_{\pm 0.000}$
& $\boldsymbol{1.00\%_{\pm 0.001}}$ & $\boldsymbol{5.09\%_{\pm 0.001}}$ & $0.752_{\pm 0.002}$ \\

\textit{Diamonds}
& $\boldsymbol{1.40\%_{\pm 0.000}}$ & $2.18\%_{\pm 0.000}$ & $1419.174_{\pm 5.703}$
& $3.92\%_{\pm 0.000}$ & $2.31\%_{\pm 0.000}$ & $1477.964_{\pm 13.909}$
& $2.36\%_{\pm 0.000}$ & $2.46\%_{\pm 0.000}$ & $1406.246_{\pm 5.114}$
& $3.79\%_{\pm 0.000}$ & $\boldsymbol{2.11\%_{\pm 0.000}}$ & $\boldsymbol{934.090_{\pm 15.111}}$ \\

\textit{Heart}
& $5.46\%_{\pm 0.004}$ & $10.76\%_{\pm 0.004}$ & $0.845_{\pm 0.016}$
& $5.67\%_{\pm 0.005}$ & $10.77\%_{\pm 0.004}$ & $\boldsymbol{0.853_{\pm 0.013}}$
& $\boldsymbol{5.42\%_{\pm 0.004}}$ & $\boldsymbol{10.08\%_{\pm 0.004}}$ & $0.837_{\pm 0.033}$
& $6.08\%_{\pm 0.007}$ & $11.11\%_{\pm 0.006}$ & $0.823_{\pm 0.028}$ \\

\textit{Liver}
& $9.26\%_{\pm 0.010}$ & $3.52\%_{\pm 0.008}$ & $3.550_{\pm 0.119}$
& $\boldsymbol{8.83\%_{\pm 0.006}}$ & $3.77\%_{\pm 0.007}$ & $3.448_{\pm 0.185}$
& $10.30\%_{\pm 0.000}$ & $\boldsymbol{2.16\%_{\pm 0.000}}$ & $3.378_{\pm 0.000}$
& $9.20\%_{\pm 0.000}$ & $3.49\%_{\pm 0.000}$ & $\boldsymbol{3.181_{\pm 0.000}}$ \\

\textit{Magic}
& $2.98\%_{\pm 0.000}$ & $\boldsymbol{0.96\%_{\pm 0.000}}$ & $0.918_{\pm 0.000}$
& $3.41\%_{\pm 0.000}$ & $\boldsymbol{0.96\%_{\pm 0.000}}$ & $0.917_{\pm 0.001}$
& $\boldsymbol{2.95\%_{\pm 0.001}}$ & $2.12\%_{\pm 0.001}$ & $0.914_{\pm 0.002}$
& $3.09\%_{\pm 0.000}$ & $0.98\%_{\pm 0.000}$ & $\boldsymbol{0.920_{\pm 0.000}}$ \\
\bottomrule
\end{tabular}
\end{adjustbox}

\vspace{3pt}
\parbox{\linewidth}{\footnotesize
Task$^{a}$ is AUROC ($\uparrow$) for classification and RMSE ($\downarrow$) for regression. The complete label tree is retained in all variants except the Condensed Label Tree variant.
}
\end{table*}

\subsection{Contribution of prompt guidance}
\label{subsec:contribution_of_prompt_guidance}

We further examine whether the prompt components used to guide LLM reasoning contribute to synthesis quality. In this experiment, the task definition and the feature-dependence information extracted from the chained trees are retained in all variants, while metric guidance, routing guidance, or the reference analysis is removed individually. Across datasets, the full prompt provides a more balanced overall synthesis performance than the ablated variants, although removing an individual component can occasionally improve a particular metric on a particular dataset. The ablation therefore supports the overall contribution of the prompt-guidance components without implying that each component improves every metric individually. Detailed results are reported in Supplementary Table~\ref{tab:prompt_guidance_ablation}.

\begin{table*}[htbp]
\centering
\caption{Ablation results for prompt-guidance components. Values are reported as mean $\pm$ s.d. Bold values indicate the best result for each dataset.}
\label{tab:prompt_guidance_ablation}
\scriptsize
\setlength{\tabcolsep}{2pt}
\renewcommand{\arraystretch}{1.08}

\begin{adjustbox}{max width=\linewidth}
\begin{tabular}{l|ccc|ccc|ccc|ccc}
\toprule
\multirow{2}{*}{Dataset}
& \multicolumn{3}{c|}{w/o Routing}
& \multicolumn{3}{c|}{w/o Metric}
& \multicolumn{3}{c|}{w/o Reference}
& \multicolumn{3}{c}{Full Prompt (Ours)} \\
\cmidrule(lr){2-4}
\cmidrule(lr){5-7}
\cmidrule(lr){8-10}
\cmidrule(lr){11-13}
& \shortstack{Marginal distribution\\error ($\downarrow$)}
& \shortstack{Pairwise correlation\\error ($\downarrow$)}
& Task$^{a}$
& \shortstack{Marginal distribution\\error ($\downarrow$)}
& \shortstack{Pairwise correlation\\error ($\downarrow$)}
& Task$^{a}$
& \shortstack{Marginal distribution\\error ($\downarrow$)}
& \shortstack{Pairwise correlation\\error ($\downarrow$)}
& Task$^{a}$
& \shortstack{Marginal distribution\\error ($\downarrow$)}
& \shortstack{Pairwise correlation\\error ($\downarrow$)}
& Task$^{a}$ \\
\midrule

\textit{Adult}
& $5.85\%_{\pm 0.054}$ & $5.07\%_{\pm 0.038}$ & $0.904_{\pm 0.001}$
& $8.29\%_{\pm 0.064}$ & $4.70\%_{\pm 0.064}$ & $\boldsymbol{0.905_{\pm 0.001}}$
& $11.92\%_{\pm 0.059}$ & $10.82\%_{\pm 0.052}$ & $0.888_{\pm 0.002}$
& $\boldsymbol{5.44\%_{\pm 0.001}}$ & $\boldsymbol{3.53\%_{\pm 0.000}}$ & $0.900_{\pm 0.001}$ \\

\textit{Blood}
& $6.97\%_{\pm 1.180}$ & $3.94\%_{\pm 0.493}$ & $0.702_{\pm 0.034}$
& $8.14\%_{\pm 0.146}$ & $3.50\%_{\pm 0.232}$ & $\boldsymbol{0.713_{\pm 0.054}}$
& $8.83\%_{\pm 0.404}$ & $3.78\%_{\pm 0.597}$ & $0.700_{\pm 0.075}$
& $\boldsymbol{3.32\%_{\pm 0.005}}$ & $\boldsymbol{2.68\%_{\pm 0.004}}$ & $0.704_{\pm 0.026}$ \\

\textit{Car}
& $1.70\%_{\pm 0.380}$ & $3.92\%_{\pm 0.403}$ & $0.915_{\pm 0.017}$
& $\boldsymbol{1.37\%_{\pm 0.224}}$ & $\boldsymbol{3.40\%_{\pm 0.292}}$ & $\boldsymbol{0.974_{\pm 0.005}}$
& $1.68\%_{\pm 0.382}$ & $3.92\%_{\pm 0.320}$ & $0.963_{\pm 0.004}$
& $1.68\%_{\pm 0.004}$ & $3.99\%_{\pm 0.004}$ & $0.958_{\pm 0.006}$ \\

\textit{Default}
& $9.58\%_{\pm 0.091}$ & $\boldsymbol{3.54\%_{\pm 0.046}}$ & $\boldsymbol{0.756_{\pm 0.003}}$
& $11.67\%_{\pm 0.078}$ & $5.89\%_{\pm 0.079}$ & $0.751_{\pm 0.003}$
& $13.89\%_{\pm 0.041}$ & $10.42\%_{\pm 0.028}$ & $0.723_{\pm 0.004}$
& $\boldsymbol{1.00\%_{\pm 0.001}}$ & $5.09\%_{\pm 0.001}$ & $0.752_{\pm 0.002}$ \\

\textit{Diamonds}
& $4.91\%_{\pm 0.000}$ & $2.97\%_{\pm 0.000}$ & $\boldsymbol{889.692_{\pm 6.476}}$
& $4.64\%_{\pm 0.000}$ & $2.51\%_{\pm 0.000}$ & $1195.031_{\pm 0.000}$
& $6.17\%_{\pm 0.000}$ & $5.06\%_{\pm 0.000}$ & $1011.838_{\pm 0.000}$
& $\boldsymbol{3.79\%_{\pm 0.000}}$ & $\boldsymbol{2.11\%_{\pm 0.000}}$ & $934.090_{\pm 15.111}$ \\

\textit{Heart}
& $7.25\%_{\pm 1.046}$ & $12.57\%_{\pm 0.864}$ & $0.817_{\pm 0.031}$
& $6.23\%_{\pm 0.873}$ & $11.21\%_{\pm 0.808}$ & $0.817_{\pm 0.029}$
& $\boldsymbol{5.70\%_{\pm 0.462}}$ & $\boldsymbol{11.08\%_{\pm 0.609}}$ & $\boldsymbol{0.849_{\pm 0.022}}$
& $6.08\%_{\pm 0.007}$ & $11.11\%_{\pm 0.006}$ & $0.823_{\pm 0.028}$ \\

\textit{Liver}
& $14.19\%_{\pm 0.729}$ & $4.84\%_{\pm 1.205}$ & $3.413_{\pm 0.085}$
& $9.67\%_{\pm 1.019}$ & $2.59\%_{\pm 0.357}$ & $3.484_{\pm 0.086}$
& $13.34\%_{\pm 0.946}$ & $\boldsymbol{2.55\%_{\pm 0.482}}$ & $3.546_{\pm 0.103}$
& $\boldsymbol{9.20\%_{\pm 0.000}}$ & $3.49\%_{\pm 0.000}$ & $\boldsymbol{3.181_{\pm 0.000}}$ \\

\textit{Magic}
& $3.29\%_{\pm 0.026}$ & $1.19\%_{\pm 0.033}$ & $0.915_{\pm 0.002}$
& $3.39\%_{\pm 0.000}$ & $0.98\%_{\pm 0.000}$ & $\boldsymbol{0.921_{\pm 0.000}}$
& $\boldsymbol{2.75\%_{\pm 0.011}}$ & $\boldsymbol{0.84\%_{\pm 0.007}}$ & $0.898_{\pm 0.002}$
& $3.09\%_{\pm 0.000}$ & $0.98\%_{\pm 0.000}$ & $0.920_{\pm 0.000}$ \\
\bottomrule
\end{tabular}
\end{adjustbox}

\vspace{3pt}
\parbox{\linewidth}{\footnotesize
Task$^{a}$ is AUROC ($\uparrow$) for classification and RMSE ($\downarrow$) for regression. The task definition and tree-derived dependence information are retained in all variants.
}
\end{table*}

\subsection{Sensitivity to different LLM backbones}
\label{subsec:sensitivity_to_different_llm_backbones}

We evaluate the sensitivity of the proposed framework to the choice of LLM backbone. We replace Gemini-2.5-Pro~\cite{comanici2025gemini} with GLM-5~\cite{zeng2026glm} and Qwen3-Max~\cite{yang2025qwen3}, while retaining the same prompt, number of candidate queries, and strategy-selection protocol. The results show that the overall TabSSD pipeline remains applicable across all three evaluated LLM backbones, indicating that the framework is not restricted to a single specific model. However, synthesis performance varies across backbones and datasets, suggesting that the choice of backbone can still affect the quality of the generated strategies. Detailed results are reported in Supplementary Table~\ref{tab:llm_backbone_sensitivity}.

\begin{table*}[htbp]
\centering
\caption{Sensitivity of TabSSD to different LLM backbones. Values are reported as mean $\pm$ s.d. Bold values indicate the best result for each dataset.}
\label{tab:llm_backbone_sensitivity}
\scriptsize
\setlength{\tabcolsep}{2pt}
\renewcommand{\arraystretch}{1.08}

\begin{adjustbox}{max width=\linewidth}
\begin{tabular}{l|ccc|ccc|ccc}
\toprule
\multirow{2}{*}{Dataset}
& \multicolumn{3}{c|}{GLM-5}
& \multicolumn{3}{c|}{Qwen3-Max}
& \multicolumn{3}{c}{Gemini-2.5-Pro} \\
\cmidrule(lr){2-4}
\cmidrule(lr){5-7}
\cmidrule(lr){8-10}
& \shortstack{Marginal distribution\\error ($\downarrow$)}
& \shortstack{Pairwise correlation\\error ($\downarrow$)}
& Task$^{a}$
& \shortstack{Marginal distribution\\error ($\downarrow$)}
& \shortstack{Pairwise correlation\\error ($\downarrow$)}
& Task$^{a}$
& \shortstack{Marginal distribution\\error ($\downarrow$)}
& \shortstack{Pairwise correlation\\error ($\downarrow$)}
& Task$^{a}$ \\
\midrule

\textit{Adult}
& $11.66\%_{\pm 0.000}$ & $11.89\%_{\pm 0.000}$ & $0.898_{\pm 0.000}$
& $6.33\%_{\pm 0.000}$ & $4.35\%_{\pm 0.000}$ & $0.894_{\pm 0.000}$
& $\boldsymbol{5.44\%_{\pm 0.001}}$ & $\boldsymbol{3.53\%_{\pm 0.000}}$ & $\boldsymbol{0.900_{\pm 0.001}}$ \\

\textit{Blood}
& $9.84\%_{\pm 0.008}$ & $5.01\%_{\pm 0.005}$ & $\boldsymbol{0.722_{\pm 0.033}}$
& $5.54\%_{\pm 0.000}$ & $3.11\%_{\pm 0.000}$ & $0.703_{\pm 0.000}$
& $\boldsymbol{3.32\%_{\pm 0.005}}$ & $\boldsymbol{2.68\%_{\pm 0.004}}$ & $0.704_{\pm 0.026}$ \\

\textit{Car}
& $1.48\%_{\pm 0.002}$ & $3.70\%_{\pm 0.003}$ & $\boldsymbol{0.963_{\pm 0.005}}$
& $\boldsymbol{1.30\%_{\pm 0.003}}$ & $\boldsymbol{3.49\%_{\pm 0.003}}$ & $0.961_{\pm 0.006}$
& $1.68\%_{\pm 0.004}$ & $3.99\%_{\pm 0.004}$ & $0.958_{\pm 0.006}$ \\

\textit{Default}
& $12.42\%_{\pm 0.000}$ & $7.03\%_{\pm 0.000}$ & $\boldsymbol{0.752_{\pm 0.000}}$
& $12.54\%_{\pm 0.000}$ & $6.64\%_{\pm 0.000}$ & $0.724_{\pm 0.000}$
& $\boldsymbol{1.00\%_{\pm 0.001}}$ & $\boldsymbol{5.09\%_{\pm 0.001}}$ & $\boldsymbol{0.752_{\pm 0.002}}$ \\

\textit{Diamonds}
& $4.62\%_{\pm 0.000}$ & $2.54\%_{\pm 0.000}$ & $\boldsymbol{881.093_{\pm 18.327}}$
& $\boldsymbol{2.97\%_{\pm 0.000}}$ & $2.54\%_{\pm 0.000}$ & $903.082_{\pm 0.000}$
& $3.79\%_{\pm 0.000}$ & $\boldsymbol{2.11\%_{\pm 0.000}}$ & $934.090_{\pm 15.111}$ \\

\textit{Heart}
& $\boldsymbol{5.23\%_{\pm 0.005}}$ & $\boldsymbol{10.83\%_{\pm 0.005}}$ & $0.852_{\pm 0.017}$
& $5.93\%_{\pm 0.000}$ & $11.36\%_{\pm 0.000}$ & $\boldsymbol{0.873_{\pm 0.000}}$
& $6.08\%_{\pm 0.007}$ & $11.11\%_{\pm 0.006}$ & $0.823_{\pm 0.028}$ \\

\textit{Liver}
& $10.79\%_{\pm 0.000}$ & $\boldsymbol{2.51\%_{\pm 0.000}}$ & $3.537_{\pm 0.000}$
& $15.01\%_{\pm 0.000}$ & $4.41\%_{\pm 0.000}$ & $3.362_{\pm 0.000}$
& $\boldsymbol{9.20\%_{\pm 0.000}}$ & $3.49\%_{\pm 0.000}$ & $\boldsymbol{3.181_{\pm 0.000}}$ \\

\textit{Magic}
& $3.94\%_{\pm 0.000}$ & $1.23\%_{\pm 0.000}$ & $0.902_{\pm 0.000}$
& $\boldsymbol{2.60\%_{\pm 0.000}}$ & $1.69\%_{\pm 0.000}$ & $0.910_{\pm 0.000}$
& $3.09\%_{\pm 0.000}$ & $\boldsymbol{0.98\%_{\pm 0.000}}$ & $\boldsymbol{0.920_{\pm 0.000}}$ \\
\bottomrule
\end{tabular}
\end{adjustbox}

\vspace{3pt}
\parbox{\linewidth}{\footnotesize
Task$^{a}$ is AUROC ($\uparrow$) for classification and RMSE ($\downarrow$) for regression. All three LLM backbones use the same prompt, number of candidate queries, and strategy-selection protocol.
}
\end{table*}
\subsection{Sensitivity to LLM query budget}
\label{appendix:query_budget}


Because LLM-generated strategies are stochastic, we further examine whether the final performance depends strongly on a single LLM output. We conduct this query-budget analysis on three datasets, \textit{Adult}, \textit{Default}, and \textit{Blood}. For each query budget \(k\), we select the best-performing strategy among the first \(k\) candidate strategies according to the average statistical error, defined as the mean of marginal distribution error and pairwise correlation error. As shown in Supplementary Fig.~\ref{fig:query_n}, the best-so-far error generally decreases and then stabilizes as the number of candidate strategies increases. This suggests that querying the LLM multiple times improves strategy-selection robustness and reduces reliance on a single stochastic LLM output, while avoiding the need for extensive search.

\begin{figure}[!htbp]
\internallinenumbers
\centering
\includegraphics[width=0.75\linewidth]{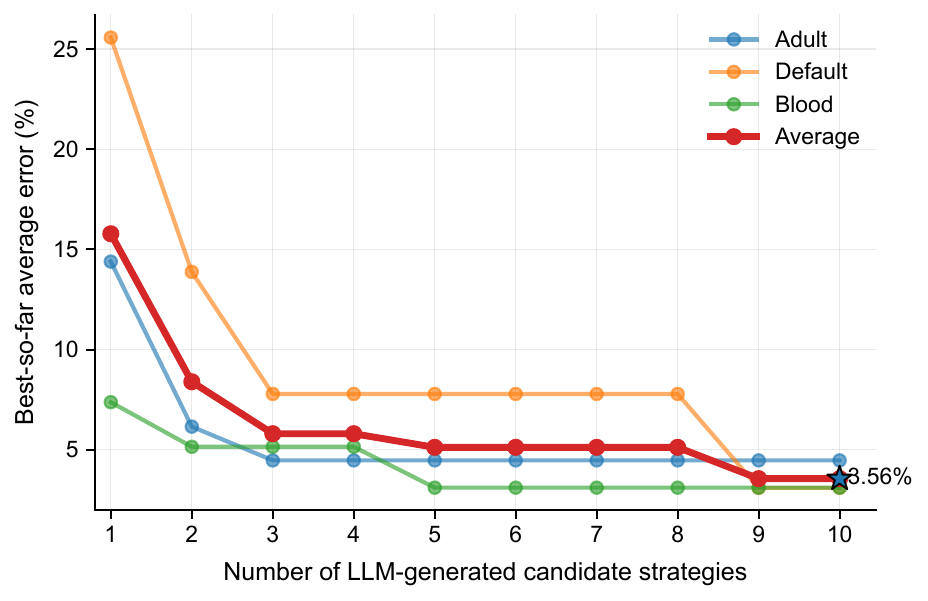}
\caption{\textbf{Effect of LLM query budget on strategy selection.}
Best-so-far average statistical error, defined as the mean of marginal distribution error and pairwise correlation error, is shown as a function of the number of LLM-generated candidate strategies. For each query budget \(k\), the selected strategy is the one with the lowest average statistical error among the first \(k\) candidates. Thin curves show results for \textit{Adult}, \textit{Default}, and \textit{Blood}, and the bold curve shows their mean.
}
\label{fig:query_n}
\end{figure}

\clearpage

\section{Detailed runtime of different algorithms on various datasets}
\label{appendix:Detailed}
Dataset-level runtime results are reported in Supplementary Table~\ref{tab:run_time}, including the total runtime of each baseline and the decomposition of TabSSD runtime into feature-CART extraction, LLM-based synthesis-strategy design, and execution of the generated strategy.
\FloatBarrier
\begin{table*}[!htbp]
\centering
\caption{Runtime of different algorithms across datasets, including single-run and 10-run parallel TabSSD runtimes, in seconds.}
\label{tab:run_time}
\scriptsize
\setlength{\tabcolsep}{1.5pt}
\renewcommand{\arraystretch}{0.95}

\begin{adjustbox}{width=\textwidth,center}
\begin{tabular}{lrrrrrrrr|rrrrrrr}
\toprule
Dataset
& ARF & NRGB & CTGAN & TVAE & GReaT & TDDPM & TabDiff & TSYN
& Tree
& Query
& Synth.
& Total
& \shortstack{Parallel\\query}
& \shortstack{Parallel\\synth.}
& \shortstack{Parallel\\total} \\
\midrule
Adult
& 586.29 & 1145.87 & 3950.74 & 1361.28 & 12492.07 & 2794.10
& 4449.82 & 2306
& 11.47 & 30.06 & 2.50 & 44.03
& 77.83 & 35.90 & 125.20 \\
Default
& 881.77 & 1191.42 & 4172.98 & 1954.85 & 23840.83 & 2414.62
& 4849.87 & 3110
& 29.39 & 82.38 & 73.65 & 185.42
& 168.70 & 93.16 & 291.25 \\
Magic
& 342.90 & 587.95 & 1846.34 & 883.19 & 9752.93 & 1504.44
& 3276.78 & 5518
& 7.91 & 48.62 & 2.57 & 59.10
& 85.09 & 13.79 & 106.79 \\
Diamonds
& 911.43 & 764.76 & 4823.80 & 1952.56 & 14456.91 & 1697.19
& 5217.43 & 2997
& 5.48 & 34.71 & 12.12 & 52.31
& 90.55 & 43.04 & 139.07 \\
Jasmine
& 233.54 & 1987.75 & 954.65 & 365.96 & -- & 7830.38
& 4488.51 & 2651
& 13.92 & 97.88 & 28.19 & 139.99
& 169.99 & 1.29 & 185.20 \\
SpeedDating
& 668.07 & 12392.89 & 2241.91 & 962.28 & -- & 5865.19
& 4415.97 & 2309
& 36.65 & 73.06 & 5.91 & 115.62
& 378.45 & 2.45 & 417.55 \\
DNA
& 322.20 & 1251.77 & 551.40 & 210.51 & -- & 7833.71
& 4728.01 & 3872
& 88.61 & 78.93 & 5.12 & 172.66
& 120.66 & 4.95 & 214.22 \\
DARWIN
& 14.54 & 8530.16 & 437.17 & 380.86 & -- & 1177.65
& 2941.14 & 1226
& 129.78 & 289.65 & 10.02 & 429.45
& 289.65 & 3.56 & 422.99 \\
Blood
& 7.05 & 327.37 & 58.93 & 44.44 & 190.21 & 1622.75
& 2980.23 & 1304
& 0.14 & 74.08 & 0.29 & 74.51
& 53.81 & 5.93 & 59.88 \\
Car
& 9.25 & 415.17 & 88.48 & 42.11 & 384.97 & 1733.42
& 1663.72 & 1619
& 0.20 & 86.41 & 0.12 & 86.73
& 197.71 & 13.14 & 211.05 \\
Heart
& 6.54 & 334.39 & 73.97 & 30.23 & 116.48 & 2018.31
& 2621.01 & 1257
& 0.37 & 96.54 & 5.10 & 102.01
& 100.06 & 5.37 & 105.80 \\
Liver
& 7.54 & 388.90 & 50.88 & 27.09 & 94.72 & 1390.71
& 1662.97 & 1489
& 0.08 & 56.62 & 0.04 & 56.74
& 112.35 & 3.68 & 116.11 \\
\bottomrule
\end{tabular}
\end{adjustbox}

\vspace{2pt}
\parbox{\textwidth}{\footnotesize
``--'' indicates that the corresponding method could not be successfully
executed on the dataset. NRGB, TDDPM, and TSYN denote NRGBoost, TabDDPM,
and TABSYN, respectively.

For TabSSD, Tree, Query, and Synth. denote the runtimes required for
feature-CART extraction, generation of the selected synthesis strategy,
and execution of the selected strategy, respectively. Total is the sum
of Tree, Query, and Synth.

The parallel columns correspond to 10 independent candidate runs.
Parallel query and Parallel synth. denote the maximum single-query time and single-synthesis time, respectively, across these runs. Parallel total is the sum of Tree, Parallel query, and
Parallel synth. The runtime analyses in Fig.~\ref{fig:scalability} use
Parallel total as the TabSSD end-to-end runtime.
}
\end{table*}
\FloatBarrier

\section{Case study of failure modes}
\label{appendix:failure}
Not all LLM-generated synthesis strategies are reliable. Supplementary Table~\ref{tab:failure_modes} presents representative failure cases in terms of fidelity, privacy, and efficiency. Unlike black-box generators, our framework produces inspectable executable strategies, allowing failure sources to be traced to specific code-level design choices and unreliable strategies to be excluded through multi-objective evaluation before final selection and deployment.
\FloatBarrier
\begin{table}[htbp]
\centering
\caption{Representative failure modes in LLM-generated synthesis strategies.
The displayed code fragments highlight the operations directly responsible for each failure.}
\label{tab:failure_modes}
\footnotesize
\renewcommand{\arraystretch}{1.22}
\setlength{\tabcolsep}{4pt}

\begin{tabular}{
@{}
>{\centering\arraybackslash\bfseries}m{0.105\linewidth}
>{\raggedright\arraybackslash}m{0.185\linewidth}
>{\raggedright\arraybackslash}m{0.625\linewidth}
@{}
}
\toprule
\rowcolor{headergray}
\textbf{Dataset} &
\textbf{Observed failure} &
\textbf{Failure-inducing design and key code fragment} \\
\midrule
Blood &
\textbf{Low fidelity}

\smallskip
Marginal distribution error:
\textbf{28.80\%}

Pairwise correlation error:
\textbf{23.82\%}
&
Subspace-specific Gaussian mixture models are used to generate the strongly dependent continuous variables
\texttt{feature\_1} and \texttt{feature\_2}. Repeated single-sample generation with a fixed
random state concentrates the outputs on only a few recurring value pairs, causing
distribution collapse. The regenerated routing attributes are also not constrained to
remain in their assigned subspaces.

\smallskip
\setlength{\fboxsep}{4pt}
\colorbox{codegray}{
\begin{minipage}[t]{0.95\linewidth}
\scriptsize\ttfamily
gmm = GaussianMixture(..., random\_state=42)\par
gmm.fit(subspace\_data[['feature\_1', 'feature\_2']])\par
f1\_f2\_sample, \_ = subspace\_model['gmm'].sample(1)\par
f0 = f0\_pred + f0\_residual
\end{minipage}}
\\[5pt]
\midrule
Car &
\textbf{Privacy leakage}

\smallskip
DCR:
\textbf{0.994}

Train-set proportion: 0.700
&
The strategy stores complete feature--label records within each subspace and directly
resamples an observed row with probability 0.98. Consequently, most synthetic records
are copied from the training support rather than newly generated. Laplace smoothing
changes only sampling probabilities and cannot prevent exact matches.

\smallskip
\setlength{\fboxsep}{4pt}
\colorbox{codegray}{
\begin{minipage}[t]{0.95\linewidth}
\scriptsize\ttfamily
self.unique\_vectors = unique\_rows\par
if np.random.rand() > diversity\_rate: \# diversity\_rate = 0.02\par
\hspace*{1em}idx = np.random.choice(len(self.unique\_vectors), ...)\par
\hspace*{1em}return self.unique\_vectors[idx]
\end{minipage}}
\\[5pt]
\midrule
Jasmine &
\textbf{Excessive cost}

\smallskip
Runtime:
\textbf{\textit{1,888.74 s}}
&
The strategy partitions a 144-feature dataset into multiple subspaces and trains a
separate conditional boosting model for nearly every feature in each subspace. During
synthesis, it further generates each record feature by feature, producing substantial
nested training and sampling overhead.

\smallskip
\setlength{\fboxsep}{4pt}
\colorbox{codegray}{
\begin{minipage}[t]{0.95\linewidth}
\scriptsize\ttfamily
for space\_id in sorted(df['subspace\_id'].unique()):\par
\hspace*{1em}for target\_feature in GENERATION\_ORDER:\par
\hspace*{2em}model.fit(X\_fit, y\_fit)\par
for \_ in range(num\_samples\_for\_subspace):\par
\hspace*{1em}for target\_feature in GENERATION\_ORDER:
\end{minipage}}
\\

\bottomrule
\end{tabular}
\end{table}

\section{Detailed predictive results for clinical data augmentation}
\label{appendix:augmentation_results}

Detailed predictive results under the \(2\times\) augmentation
setting are reported in Supplementary Table~\ref{tab:thoracic_predictive_utility}.
Balanced accuracy and minority-class accuracy are evaluated on the real test set. 
\begin{table}[htbp]
\centering
\caption{\textbf{Predictive utility under the \(2\times\) augmentation setting.}}
\label{tab:thoracic_predictive_utility}
\small
\setlength{\tabcolsep}{6pt}
\renewcommand{\arraystretch}{1.1}
\begin{tabular}{lcc}
\toprule
Method & Balanced accuracy & Minority-class accuracy \\
\midrule
TabSSD   & \(0.5785 \pm 0.0162\) & \(0.5874 \pm 0.0331\) \\
SMOTE    & \(0.5638 \pm 0.0158\) & \(0.2557 \pm 0.0303\) \\
ARF      & \(0.5499 \pm 0.0251\) & \(0.3259 \pm 0.0495\) \\
CTGAN    & \(0.5106 \pm 0.0130\) & \(0.1052 \pm 0.0254\) \\
TVAE     & \(0.5197 \pm 0.0051\) & \(0.2586 \pm 0.0102\) \\
GReaT    & \(0.5229 \pm 0.0199\) & \(0.2236 \pm 0.0460\) \\
TabDDPM  & \(0.5550 \pm 0.0141\) & \(0.2224 \pm 0.0290\) \\
TabDiff  & \(0.5634 \pm 0.0203\) & \(0.3236 \pm 0.0459\) \\
TABSYN   & \(0.5360 \pm 0.0168\) & \(0.2460 \pm 0.0306\) \\
NRGBoost & \(0.5525 \pm 0.0108\) & \(0.1707 \pm 0.0202\) \\
\bottomrule
\end{tabular}
\end{table}
\section{Tailored data synthesis strategies}

\label{appendix:tailored}

This section presents the concrete Python synthesis strategies generated by TabSSD for the selected datasets. For clarity, only the executable strategy code is shown, without reproducing the corresponding LLM analyses. Owing to its substantially greater length, the strategy for the \textit{DARWIN} dataset is presented in an abridged form, with part of the embedded feature-dependence information omitted for brevity.
\begin{lstlisting}[
    style=textlisting,
    numbers=none,
    breaklines=true,
    breakatwhitespace=false,
    caption={Generated synthesis strategy for the Blood dataset.},
    label={lst:blood}
]
def generate_synthesis_data(save_path: str, n_sample: int, X_train: np.ndarray, y_train: np.ndarray) -> None:
    """
    Generates synthetic tabular data by inferring characteristics from pre-trained decision trees.

    The strategy involves:
    1.  Analyzing feature dependencies from provided tree performance metrics.
    2.  Using the main label-prediction decision tree to define subspaces (leaves).
    3.  Fitting a Gaussian Copula model to the continuous features within each subspace to capture local correlations.
    4.  Modeling the categorical label based on its empirical distribution in each subspace.
    5.  Generating new data by sampling from these fitted models, ensuring generated values respect subspace boundaries.
    """
    # All necessary imports are contained within the function.
    import numpy as np
    import pandas as pd
    from sklearn.preprocessing import QuantileTransformer
    import csv
    import warnings

    # Suppress warnings from QuantileTransformer for small n_quantiles
    warnings.filterwarnings("ignore", category=UserWarning)

    feature_names = [f"feature_{i}" for i in range(4)]
    all_feature_names = feature_names + ["label"]
    
    # Define global feature ranges for final clamping
    feature_ranges = {
        0: (0.0, 74.0),
        1: (1.0, 50.0),
        2: (250.0, 12500.0),
        3: (2.0, 98.0)
    }

    # Helper function to get leaf boundaries for clamping. This is critical for fidelity.
    def get_leaf_bounds():
        bounds = {
            1: {0: (-np.inf, 6.5), 2: (-np.inf, 1125.0), 3: (-np.inf, 3.0)},
            2: {0: (-np.inf, 5.0), 2: (-np.inf, 1125.0), 3: (3.0, 12.0)},
            3: {0: (5.0, 6.5), 2: (-np.inf, 1125.0), 3: (3.0, 12.0)},
            4: {0: (-np.inf, 6.5), 2: (-np.inf, 625.0), 3: (12.0, np.inf)},
            5: {0: (-np.inf, 6.5), 2: (625.0, 1125.0), 3: (12.0, 24.5)},
            6: {0: (-np.inf, 6.5), 2: (625.0, 1125.0), 3: (24.5, np.inf)},
            7: {0: (-np.inf, 6.5), 2: (1125.0, np.inf), 3: (-np.inf, 22.5)},
            8: {0: (-np.inf, 6.5), 2: (1125.0, np.inf), 3: (22.5, 31.0)},
            9: {0: (-np.inf, 6.5), 2: (1125.0, np.inf), 3: (31.0, 33.5)},
            10: {0: (-np.inf, 6.5), 1: (-np.inf, 5.5), 2: (1125.0, np.inf), 3: (33.5, 50.0)},
            11: {0: (-np.inf, 6.5), 1: (5.5, np.inf), 2: (1125.0, np.inf), 3: (33.5, 50.0)},
            12: {0: (-np.inf, 6.5), 1: (-np.inf, 11.5), 2: (1125.0, np.inf), 3: (50.0, np.inf)},
            13: {0: (-np.inf, 6.5), 1: (-np.inf, 15.0), 2: (1125.0, 4500.0), 3: (50.0, 78.5)},
            14: {0: (-np.inf, 3.5), 1: (15.0, np.inf), 2: (1125.0, 4500.0), 3: (50.0, 78.5)},
            15: {0: (3.5, 5.0), 1: (15.0, np.inf), 2: (1125.0, 4500.0), 3: (50.0, 78.5)},
            16: {0: (5.0, 6.5), 1: (15.0, np.inf), 2: (1125.0, 4500.0), 3: (50.0, 78.5)},
            17: {0: (-np.inf, 6.5), 2: (1125.0, 4500.0), 3: (78.5, np.inf)},
            18: {0: (-np.inf, 6.5), 2: (4500.0, np.inf), 3: (50.0, np.inf)},
            19: {0: (6.5, np.inf), 3: (-np.inf, 58.5)},
            20: {0: (6.5, np.inf), 3: (58.5, np.inf)}
        }
        return bounds
    
    LEAF_BOUNDS = get_leaf_bounds()

    # Step 1: Define the Routing Function based on the decision tree for the label.
    def get_leaf_id(sample):
        f0, f1, f2, f3 = sample['feature_0'], sample['feature_1'], sample['feature_2'], sample['feature_3']
        if f0 <= 6.5:
            if f2 <= 1125.0:
                if f3 <= 3.0: return 1
                else:
                    if f3 <= 12.0:
                        if f0 <= 5.0: return 2
                        else: return 3
                    else:
                        if f2 <= 625.0: return 4
                        else:
                            if f3 <= 24.5: return 5
                            else: return 6
            else:
                if f3 <= 50.0:
                    if f3 <= 22.5: return 7
                    else:
                        if f3 <= 31.0: return 8
                        else:
                            if f3 <= 33.5: return 9
                            else:
                                if f1 <= 5.5: return 10
                                else: return 11
                else:
                    if f1 <= 11.5: return 12
                    else:
                        if f2 <= 4500.0:
                            if f3 <= 78.5:
                                if f1 <= 15.0: return 13
                                else:
                                    if f0 <= 3.5: return 14
                                    else:
                                        if f0 <= 5.0: return 15
                                        else: return 16
                            else: return 17
                        else: return 18
        else:
            if f3 <= 58.5: return 19
            else: return 20
        return -1 # Should not be reached with valid data

    # Step 2: Partition the original data into subspaces using the routing function.
    X_df = pd.DataFrame(X_train, columns=feature_names)
    data_df = X_df.copy()
    data_df['label'] = y_train

    subspaces = {i: [] for i in range(1, 21)}
    for _, row in data_df.iterrows():
        leaf_id = get_leaf_id(row)
        if leaf_id != -1:
            subspaces[leaf_id].append(row.values)

    for leaf_id in subspaces:
        subspaces[leaf_id] = pd.DataFrame(subspaces[leaf_id], columns=all_feature_names)

    # Step 3: Learn a generative model for each subspace.
    models = {}
    for leaf_id, subspace_df in subspaces.items():
        n_points = len(subspace_df)
        if n_points < 4:  # Need a few points to fit a model reliably.
            models[leaf_id] = None
            continue

        continuous_data = subspace_df[feature_names]
        labels = subspace_df['label']

        qt = QuantileTransformer(output_distribution='normal', n_quantiles=max(2, min(n_points // 2, 100)), subsample=min(n_points, int(1e5)))
        transformed_data = qt.fit_transform(continuous_data)
        transformed_data = np.nan_to_num(transformed_data) # Handle constant columns
        
        copula_corr = np.corrcoef(transformed_data, rowvar=False)
        copula_corr = np.nan_to_num(copula_corr, nan=0.0) # Handle NaN in correlation matrix
        np.fill_diagonal(copula_corr, 1.0)
        
        label_dist = labels.value_counts(normalize=True).to_dict()
        if 0 not in label_dist: label_dist[0] = 0.0
        if 1 not in label_dist: label_dist[1] = 0.0

        models[leaf_id] = {
            "qt": qt,
            "copula_corr": copula_corr,
            "label_dist": label_dist,
            "n_points": n_points
        }

    # Step 4: Generate new data by sampling from the fitted models.
    total_original_points = len(data_df) if len(data_df) > 0 else 1
    synthesis_samples_list = []

    for leaf_id, model in models.items():
        if model is None:
            continue

        n_samples_leaf = int(np.round(n_sample * model['n_points'] / total_original_points))
        if n_samples_leaf == 0:
            continue

        mean = np.zeros(len(feature_names))
        try:
            # Add a small ridge to ensure matrix is positive semi-definite
            cov_matrix = model['copula_corr'] + np.eye(len(feature_names)) * 1e-6
            mvn_samples = np.random.multivariate_normal(mean, cov_matrix, n_samples_leaf)
        except np.linalg.LinAlgError:
            continue # Skip subspace if generation fails

        synthetic_continuous = model['qt'].inverse_transform(mvn_samples)

        leaf_bounds = LEAF_BOUNDS.get(leaf_id, {})
        for i in range(len(feature_names)):
            global_min, global_max = feature_ranges[i]
            leaf_min, leaf_max = leaf_bounds.get(i, (-np.inf, np.inf))
            final_min = max(global_min, leaf_min)
            final_max = min(global_max, leaf_max)
            synthetic_continuous[:, i] = np.clip(synthetic_continuous[:, i], final_min, final_max)

        labels_to_gen = list(model['label_dist'].keys())
        label_probs = list(model['label_dist'].values())
        synthetic_labels = np.random.choice(labels_to_gen, size=n_samples_leaf, p=label_probs)

        for j in range(n_samples_leaf):
            record = {fname: val for fname, val in zip(feature_names, synthetic_continuous[j])}
            record['label'] = int(synthetic_labels[j])
            synthesis_samples_list.append(record)
    
    # Adjust sample count to exactly match n_sample
    if len(synthesis_samples_list) > n_sample:
        synthesis_samples_list = synthesis_samples_list[:n_sample]
    elif len(synthesis_samples_list) < n_sample and synthesis_samples_list:
        diff = n_sample - len(synthesis_samples_list)
        extras = np.random.choice(synthesis_samples_list, size=diff)
        synthesis_samples_list.extend(extras)

    # Step 5: Save the generated data to a CSV file.
    with open(save_path, "w", newline="") as csv_file:
        writer = csv.DictWriter(csv_file, fieldnames=all_feature_names)
        writer.writeheader()
        if synthesis_samples_list:
            writer.writerows(synthesis_samples_list)
\end{lstlisting}

\begin{lstlisting}[
    style=textlisting,
    numbers=none,
    breaklines=true,
    breakatwhitespace=false,
    caption={Generated synthesis strategy for the Car dataset.},
    label={lst:car}
]
def generate_synthesis_data(save_path: str, n_sample: int, X_train: np.ndarray, y_train: np.ndarray) -> None:
    import pandas as pd
    import numpy as np
    import csv

    # 1. Data Preparation
    # Map X_train to features 0-5 and y_train to label
    feat_cols = [f"feature_{i}" for i in range(6)]
    df = pd.DataFrame(X_train, columns=feat_cols)
    df['label'] = y_train

    # 2. Routing Function: Partitioning into Subspaces based on Decision Tree logic
    def get_subspace(data):
        f3 = data['feature_3']
        f5 = data['feature_5']
        # Condition 1: feature_3 <= 0.5 (Significant split for Class 2)
        # Condition 2: feature_3 > 0.5 and feature_5 <= 0.5
        # Condition 3: feature_3 > 0.5 and 0.5 < feature_5 <= 1.5
        # Condition 4: feature_3 > 0.5 and feature_5 > 1.5
        conditions = [
            (f3 <= 0.5),
            (f3 > 0.5) & (f5 <= 0.5),
            (f3 > 0.5) & (f5 > 0.5) & (f5 <= 1.5),
            (f3 > 0.5) & (f5 > 1.5)
        ]
        choices = [1, 2, 3, 4]
        # Use select to assign each row to a subspace ID
        return np.select(conditions, choices, default=4)

    df['subspace'] = get_subspace(df)

    # 3. Fit: Learning distribution characteristics within each subspace
    subspace_stats = {}
    subspace_weights_series = df['subspace'].value_counts(normalize=True)

    for s_id in df['subspace'].unique():
        s_df = df[df['subspace'] == s_id]

        # Model P(Label | Subspace)
        l_counts = s_df['label'].value_counts(normalize=True).to_dict()
        l_vals = list(l_counts.keys())
        l_weights = np.array(list(l_counts.values()), dtype='float64')
        l_weights /= l_weights.sum() # Normalize for safety

        # Model P(Feature_i | Label, Subspace)
        f_probs_given_l = {}
        for l_val in s_df['label'].unique():
            sl_df = s_df[s_df['label'] == l_val]
            f_probs_given_l[l_val] = {}
            for col in feat_cols:
                counts = sl_df[col].value_counts(normalize=True).to_dict()
                f_v = list(counts.keys())
                f_w = np.array(list(counts.values()), dtype='float64')
                f_w /= f_w.sum()
                f_probs_given_l[l_val][col] = (f_v, f_w)

        subspace_stats[s_id] = {
            'label_dist': (l_vals, l_weights),
            'feature_dists': f_probs_given_l
        }

    # 4. Synthesize Data
    synthesis_samples = []
    s_ids = list(subspace_stats.keys())
    s_probs = [subspace_weights_series.get(sid, 0) for sid in s_ids]
    s_probs = np.array(s_probs) / sum(s_probs)

    for _ in range(n_sample):
        # Sample Subspace
        sid = np.random.choice(s_ids, p=s_probs)
        stats = subspace_stats[sid]

        # Sample Label within Subspace
        l_vals, l_weights = stats['label_dist']
        l_choice = np.random.choice(l_vals, p=l_weights)

        # Sample Features conditioned on Label and Subspace
        row = {'label': int(l_choice)}
        for col in feat_cols:
            f_vals, f_weights = stats['feature_dists'][l_choice][col]
            f_choice = np.random.choice(f_vals, p=f_weights)
            row[col] = int(f_choice)

        synthesis_samples.append(row)

    # 5. Output to CSV
    fieldnames = feat_cols + ["label"]
    with open(save_path, "w", newline="") as csv_file:
        writer = csv.DictWriter(csv_file, fieldnames=fieldnames)
        writer.writeheader()
        writer.writerows(synthesis_samples)
\end{lstlisting}

\begin{lstlisting}[
    style=textlisting,
    numbers=none,
    breaklines=true,
    breakatwhitespace=false,
    caption={Abridged excerpt from the synthesis strategy generated for the \textit{DARWIN} dataset.},
    label={lst:darwin}
]
def generate_synthesis_data(save_path: str, n_sample: int, X_train:
                            np.ndarray, y_train: np.ndarray) -> None:
    """
    Generates synthetic tabular data based on inferred characteristics from pre-trained decision trees.

    The strategy involves:
    1.  Parsing decision tree summaries to build a feature dependency graph.
    2.  Performing a topological sort to establish a valid, sequential generation order.
    3.  Using the final label-prediction tree to define a "routing function" that partitions
        the data space into subspaces with homogeneous labels.
    4.  Fitting models within each subspace to capture local data distributions:
        - Linear Regression + Residual Sampling for features with strong predictors.
        - Empirical Distribution Sampling for root/un-modeled features.
    5.  Sequentially generating new samples according to the determined order and fitted models,
        ensuring values are within known bounds.
    6.  Saving the generated data to a CSV file.
    """
    import pandas as pd
    import numpy as np
    import csv
    import re
    from collections import defaultdict, Counter
    from sklearn.linear_model import LinearRegression
    import warnings

    warnings.filterwarnings("ignore", category=UserWarning)

    # 1. Parse feature information and define constants
    # The decision tree rule strings are embedded for self-containment.
    feature_info_str = """
    {feature_49}:{type:continuous,min:0.0,max:14.0,associated_features:[feature_36,feature_38,feature_334,feature_105],performance:{R2:0.9911,RMSE:0.2114}}
    {feature_94}:{type:continuous,min:736.0,max:9830.0,associated_features:[feature_251,feature_95,feature_201,feature_100],performance:{R2:0.9752,RMSE:199.1262}}

    ... additional feature summaries omitted for brevity ...

    {feature_41}:{type:continuous,min:0.0,max:4456.0,associated_features:[feature_53,feature_49,feature_48,feature_227],performance:{R2:0.9838,RMSE:87.5136}}
    {feature_88}:{type:continuous,min:0.0,max:340361.9688,associated_features:[feature_87,feature_122,feature_248,feature_222],performance:{R2:0.9922,RMSE:4425.3103}}
    """
    
    def parse_feature_info(info_str):
        feature_models = {}
        feature_metas = {}
        pattern = re.compile(
            r"\{feature_(\d+)\}:\{type:([^,]+),min:([^,]+),max:([^,]+),associated_features:\[([^\]]*)],performance:\{R2:([^,]+),RMSE:([^\}]+)\}\}"
        )
        for line in info_str.strip().split('\n'):
            match = pattern.match(line.strip())
            if match:
                target_id = int(match.group(1))
                assoc_str = match.group(5)
                predictors = []
                if assoc_str:
                    predictors = [int(f.replace('feature_', '')) for f in assoc_str.split(',')]
                
                feature_models[target_id] = predictors
                feature_metas[target_id] = {
                    'type': match.group(2),
                    'min': float(match.group(3)),
                    'max': float(match.group(4)),
                    'r2': float(match.group(6)),
                    'rmse': float(match.group(7))
                }
        return feature_models, feature_metas

    feature_models, feature_metas = parse_feature_info(feature_info_str)
    
    NUM_FEATURES = 450
    all_feature_indices = list(range(NUM_FEATURES))
    feature_names = [f"feature_{i}" for i in all_feature_indices]
    
    CONSTANT_FEATURES = {k for k, v in feature_metas.items() if v['min'] == v['max']}

    # 2. Define the Subspace Routing Logic from the label decision tree
    def get_subspace_id(row):
        if row['feature_413'] <= 8690.0:
            if row['feature_446'] <= 54097.5:
                if row['feature_204'] <= 1.61:
                    return 0  # Leaf 1 -> Label 0
                else:
                    return 1  # Leaf 2 -> Label 1
            else:  # feature_446 > 54097.5
                return 2  # Leaf 3 -> Label 1
        else:  # feature_413 > 8690.0
            if row['feature_305'] <= 28607.5:
                if row['feature_202'] <= 763.0:
                    return 3  # Leaf 4 -> Label 0
                else:  # feature_202 > 763.0
                    if row['feature_171'] <= 0.03:
                        return 4  # Leaf 5 -> Label 0
                    else:  # feature_171 > 0.03
                        return 5  # Leaf 6 -> Label 1
            else:  # feature_305 > 28607.5
                if row['feature_329'] <= 11045.5:
                    return 6  # Leaf 7 -> Label 1
                else:  # feature_329 > 11045.5
                    return 7  # Leaf 8 -> Label 0
    
    subspace_labels = {0: 0, 1: 1, 2: 1, 3: 0, 4: 0, 5: 1, 6: 1, 7: 0}

    # 3. Determine Generation Order via Topological Sort
    dependencies = {k: v for k, v in feature_models.items()}
    all_features_set = set(all_feature_indices)
    
    generation_order = []
    generated_features = set()

    # Start with features that are not targets of any model (root nodes)
    unmodeled_targets = all_features_set - set(dependencies.keys())
    generation_order.extend(list(unmodeled_targets))
    generated_features.update(unmodeled_targets)

    # Iteratively add features whose dependencies are met
    while len(generated_features) < len(all_features_set):
        features_to_add = set()
        for f in (all_features_set - generated_features):
            if f in dependencies:
                preds = dependencies.get(f, [])
                if all(p in generated_features for p in preds):
                    features_to_add.add(f)
        
        if not features_to_add:
            # Cycle detected. Break it by finding the feature with the lowest R2 in the cycle.
            cyclic_features = all_features_set - generated_features
            feature_to_break = min(cyclic_features, key=lambda f: feature_metas.get(f, {'r2': 999})['r2'])
            features_to_add.add(feature_to_break)
        
        # Sort to make order deterministic, then add
        for f in sorted(list(features_to_add)):
            generation_order.append(f)
            generated_features.add(f)
    
    # 4. Synthesizer Class
    class SubspaceSynthesizer:
        def __init__(self, order, metas, models, constants, num_features):
            self.order = order
            self.metas = metas
            self.models = models
            self.constants = constants
            self.num_features = num_features
            
            self.subspace_models = defaultdict(dict)
            self.subspace_residuals = defaultdict(dict)
            self.subspace_marginals = defaultdict(dict)
            self.subspace_proportions = {}
            self.global_marginals = {} # Fallback

        def fit(self, df):
            df['subspace_id'] = df.apply(get_subspace_id, axis=1)
            self.subspace_proportions = df['subspace_id'].value_counts(normalize=True).to_dict()

            # Store global marginals as a fallback
            for feature_id in self.order:
                 self.global_marginals[feature_id] = df[f'feature_{feature_id}'].values

            grouped = df.groupby('subspace_id')
            for subspace_id, subspace_df in grouped:
                if subspace_df.empty: continue
                
                for feature_id in self.order:
                    fname = f'feature_{feature_id}'
                    
                    if feature_id in self.constants: continue
                    
                    # Decide if this feature is modeled conditionally or marginally
                    is_conditionally_modeled = feature_id in self.models and all(
                        p in self.order[:self.order.index(feature_id)] for p in self.models[feature_id]
                    )

                    if is_conditionally_modeled:
                        predictors = self.models[feature_id]
                        pred_names = [f'feature_{p}' for p in predictors]
                        
                        # Ensure all predictor columns exist in the subspace df
                        if not all(p in subspace_df.columns for p in pred_names):
                            is_conditionally_modeled = False
                        else:
                            X_sub = subspace_df[pred_names]
                            y_sub = subspace_df[fname]
                            
                            if X_sub.empty or y_sub.empty:
                                is_conditionally_modeled = False
                            else:
                                try:
                                    model = LinearRegression()
                                    model.fit(X_sub, y_sub)
                                    predictions = model.predict(X_sub)
                                    residuals = y_sub - predictions
                                    
                                    self.subspace_models[subspace_id][feature_id] = model
                                    self.subspace_residuals[subspace_id][feature_id] = residuals.values
                                except ValueError: # E.g., if X_sub has no variation
                                    is_conditionally_modeled = False

                    if not is_conditionally_modeled:
                        self.subspace_marginals[subspace_id][feature_id] = subspace_df[fname].values

        def sample(self, n_samples):
            all_samples = []
            
            subspace_ids = list(self.subspace_proportions.keys())
            subspace_probs = list(self.subspace_proportions.values())
            
            if not subspace_ids: # Handle case where training data is empty
                return []
                
            # Assign samples to subspaces based on proportions
            assignments = np.random.choice(subspace_ids, size=n_samples, p=subspace_probs)
            
            for i in range(n_samples):
                subspace_id = assignments[i]
                new_sample = {}
                
                for feature_id in self.order:
                    fname = f'feature_{feature_id}'
                    fmeta = self.metas.get(feature_id, {})
                    fmin = fmeta.get('min', -np.inf)
                    fmax = fmeta.get('max', np.inf)

                    val = 0.0 # Default value
                    if feature_id in self.constants:
                        val = 0.0
                    elif feature_id in self.subspace_models.get(subspace_id, {}):
                        model = self.subspace_models[subspace_id][feature_id]
                        residuals = self.subspace_residuals[subspace_id][feature_id]
                        predictors = self.models[feature_id]
                        pred_names = [f'feature_{p}' for p in predictors]

                        pred_vals_list = [new_sample[p_name] for p_name in pred_names]
                        pred_df = pd.DataFrame([pred_vals_list], columns=pred_names)
                        
                        base_val = model.predict(pred_df)[0]
                        residual = np.random.choice(residuals) if len(residuals) > 0 else 0
                        val = base_val + residual
                    else:
                        # Use subspace marginal if available, else global marginal as fallback
                        marginal_vals = self.subspace_marginals.get(subspace_id, {}).get(feature_id)
                        if marginal_vals is None or len(marginal_vals) == 0:
                            marginal_vals = self.global_marginals.get(feature_id)

                        if marginal_vals is not None and len(marginal_vals) > 0:
                            val = np.random.choice(marginal_vals)
                        else: # Ultimate fallback for features with no data
                            val = np.random.uniform(fmin if np.isfinite(fmin) else 0, fmax if np.isfinite(fmax) else 1)
                    
                    # Clip to ensure validity
                    new_sample[fname] = np.clip(val, fmin, fmax)
                
                new_sample['label'] = subspace_labels[subspace_id]
                all_samples.append(new_sample)
                
            return all_samples

    # 5. Main execution logic
    df_train = pd.DataFrame(X_train, columns=feature_names)
    df_train['label'] = y_train

    # Instantiate, fit, and sample
    synthesizer = SubspaceSynthesizer(generation_order, feature_metas, feature_models, CONSTANT_FEATURES, NUM_FEATURES)
    synthesizer.fit(df_train)
    synthetic_samples = synthesizer.sample(n_sample)

    # 6. Save to CSV
    if not synthetic_samples:
        print("Warning: No synthetic samples were generated.")
        return

    output_fieldnames = [f"feature_{i}" for i in all_feature_indices] + ["label"]
    with open(save_path, "w", newline="") as csv_file:
        writer = csv.DictWriter(csv_file, fieldnames=output_fieldnames)
        writer.writeheader()
        writer.writerows(synthetic_samples)

    print(f"Successfully generated {len(synthetic_samples)} synthetic samples and saved to {save_path}")
\end{lstlisting}

\end{document}